\documentclass[runningheads]{llncs}
\usepackage[T1]{fontenc}
\usepackage{microtype}
\usepackage{graphicx}
\usepackage{subfigure}
\usepackage{booktabs} 

\usepackage{amsfonts, amsmath, amsthm}
\usepackage{makecell}
\usepackage{graphicx}

\usepackage{xcolor}
\definecolor{outerspace}{rgb}{0.1, 0.34, 0.45}

\usepackage[misc]{ifsym}

\usepackage{hyperref}
\hypersetup{
	pdftitle={Rethinking Exponential Averaging of the Fisher},
	pdfauthor={},
	plainpages=false,
	colorlinks=true,
	linkcolor=outerspace,
	citecolor=outerspace,
	urlcolor=outerspace
}

\begin{document}
	
\newtheorem*{Proposition_3_1}{Proposition 3.1: Analytic Solution of \textsc{w}o\textsc{qm} step}

\newtheorem*{Proposition_4_1}{Proposition 4.1: Analytic Solution of \textsc{so-kld-wrm} step}

\newtheorem*{Proposition_4_2}{Proposition 4.2: Analytic Solution of \textsc{q-kld-wrm} step}

\newcommand{\norm}[1]{\left\lVert#1\right\rVert}
\title{Rethinking Exponential Averaging of the Fisher}

\author{Constantin Octavian Puiu\orcidID{0000-0002-1724-4533} \Letter}

\authorrunning{C. O. Puiu}
%
\institute{University of Oxford, Mathematical Institute,\\
	\email{constantin.puiu@maths.ox.ac.uk}\\
}

\toctitle{Rethinking Exponential Averaging of the Fisher}
\tocauthor{Constantin~Octavian~Puiu}

\maketitle              
\begin{abstract}
In optimization for Machine learning (ML), it is typical that curvature-matrix (CM) estimates rely on an exponential average (EA) of local estimates (giving \textsc{ea-cm} algorithms). This approach has little principled justification, but is very often used in practice. In this paper, we draw a connection between \textsc{ea-cm} algorithms and what we call a \textit{``Wake of Quadratic models''}. The outlined connection allows us to understand what \textsc{ea-cm} algorithms are doing from an optimization perspective. Generalizing from the established connection, we propose a new family of algorithms, \textit{KL-Divergence Wake-Regularized Models} (\textsc{kld-wrm}). We give three different practical instantiations of \textsc{kld-wrm}, and show numerically that these outperform \textsc{k-fac} on MNIST.


\end{abstract}
\section{Introduction}

Recent research in optimization for ML has focused on finding tractable approximations for curvature matrices. In addition to employing ingenious approximating structures (\cite{KFAC,SENG,ADAM}), curvature matrices are typically estimated as an exponential average (EA) of the previously encountered local estimates (\cite{KFAC,ADAM,efficientbackprop,nomorepeskyLR,AdaptiveNGforVarious_stochastic_models}). This is in particular true for Natural Gradient (NG) algorithms, which we focus on here. These exponential averages emerge rather heuristically, and have so far only been given incomplete motivations. The main such motivation is \textit{``allowing curvature information to depend on much more data, with exponentially less dependence on older data''} \cite{KFAC}. However, it remains unclear what such an EA algorithm is actually doing from an optimization perspective. In this paper, we show that such EA algorithms can be seen as solving a sequence of local quadratic models whose construction obeys a recursive relationship - which we refer to as a \textit{``Wake of Quadratic Models''}. Inspired by this recursion, we consider a similar, more principled and general recursion for our local models - which we refer to as a \textit{''KL-Divergence Wake-Regularized Models''} (\textsc{kld-wrm}). We show that under \textit{suitable approximations}, \textsc{kld-wrm} gives very similar optimization steps to \textsc{ea-ng}. This equivalence raises the hope that using better approximations in our proposed class of algorithms (\textsc{kld-wrm}) might lead to algorithms which outperform \textsc{ea-ng}. We propose three practical instantiations of \textsc{kld-wrm} (of increasing approx.\ accuracy) and compare them with \textsc{k-fac}, the most widely used practical implementation of NG for DNNs. Numerically, \textsc{kld-wrm} outperforms \textsc{k-fac} on MNIST: higher test accuracy, lower test loss, lower variance.
\newpage

\section{Preliminaries}
\subsection{Neural Networks, Supervised Learning and Notation}
We very briefly look at \textit{fully-connected (FC) nets} (we omit CNN for simplicity). We have $\bar a_0 := [x,1]^T$ ($x$ is the input to the net). The \textit{pre-activation} at layer $l$: $z_l = W_l \bar a_{l-1}$ for $l \in \{1,2,...,n_L\}$. The \textit{post-activation} at layer $l$: $a_l = \phi_l(z_l)$ for $l \in \{1,2,...,n_L\}$.  The \textit{augmented post-activation} at layer $l$: $\bar a_l = [a_l, 1]^T$ for $l \in \{1,2,..., n_L-1\}$ (we augment the post-activations to incorporate the bias into the weight matrix $W_{l+1}$ w.l.o.g.; this is standard practice for \textsc{k-fac} \cite{KFAC}).

In the above, we consider $n_L$ layers, and $\phi_l(\cdot)$ are nonlinear activation functions. We collect all parameters, namely $\{W_{l}\}_{l=1}^{n_L}$, in a parameter vector $\theta\in\mathbb R^d$. Let us consider $N_o\in\mathbb Z^+$ neurons in the output layer. The output of the net is $h_\theta(x) := a_{n_L}\in\mathbb R^{N_o}$. The predictive (model) distribution of $y|x$ is $p_\theta(y|x) = p_\theta(y| h_\theta(x))$, that is $p(y|x)$ depends on $x$ only through $h_\theta(x)$.

In supervised learning, where we have labeled pairs of datasets $\mathcal D = \{(x_i, y_i)\}_i$. Our objective to minimize is typically some regularized modification of 
\begin{equation}
f(\theta) = -\log p(\mathcal D|\theta) = \sum_{(x_i,y_i)\in\mathcal D}\big(- \log p (y_i|h_\theta(x_i))\big). 
\end{equation}
Thus, we have our loss function $L(y_i,x_i;\theta) := - \log p (y_i|h_\theta(x_i))$, and $f(\theta) = \sum_{(x_i,y_i)\in\mathcal D} L(y_i,x_i;\theta)$. This is what we will focus on here, for simplicity of exposition, but our ideas directly apply to different ML paradigms, such as Variational Inference (VI) in Bayesian Neural Networks (BNNs), and to RL as well. 

Let us consider the iterates $\{\theta_k\}_{k=0,1,...}$, with $\theta_0$ initialized in standard manner (perhaps on the edge of Chaos)\footnote{Note that the initialization issue is orthogonal to our purpose.}. Let us denote the optimization steps taken at $\theta_k$ as $s_k$, that is $\theta_{k+1} = \theta_k + s_k$. Let $g(\theta)$ and $H(\theta)$ be the gradient and hessian of our objective $f(\theta)$. We will use the notation $g_k:=g(\theta_k)$ and $H_k:=H(\theta_k)$.

\subsection{KL-Divergence and Fisher Information Matrix}
The \textit{symmetric}\footnote{For convenience, we will refer to the \textit{symmetric KL-Divergence} as \textit{SKL-Divergence}.} KL-Divergence is a distance measure between two distributions:
\begin{equation}
\mathbb D_{KL}(p,q):= \frac{1}{2}\biggr[\mathbb D_{KL}(q\,||\,p) + \mathbb D_{KL}(p\,||\,q)\biggl] =   \frac{1}{2}\mathbb E_{x\sim p}\biggl[\log \frac{p(x)}{q(x)}\biggr] +  \frac{1}{2}\mathbb E_{x\sim q}\biggl[\log \frac{q(x)}{p(x)}\biggr].
\end{equation}
In our case, we are interested in the symmetric KL-divergence (SKL) between the data joint distribution with one parameter value and the data joint distribution with a different parameter value. We have
\begin{equation}
\mathbb D_{KL}(\theta_1 , \theta_2) := \mathbb D_{KL}\big(p_{\theta_1}(x,y),p_{\theta_2}(x,y)\big).
\end{equation}
\noindent Since we only model the conditional distribution, we let $p_{\theta_i}(x,y) = \hat q(x) p(y|h_{\theta_i}(x))$, where $\hat q(x)$ is the marginal (empirical) data distribution of $x$. This gives

\begin{equation}
\mathbb D_{KL}(\theta_1 , \theta_2) = \frac{1}{N}\sum_{x_i\in\mathcal D} \mathbb D_{KL}\big(p_{\theta_1}(y| x_i) , p_{\theta_2}(y| x_i)\big).
\label{final_D_KL_we_care_about_2}
\end{equation}
The Fisher has multiple definitions depending on the situation, but here we have
\begin{equation}
F_k := F(\theta_k) := \mathbb E_{\substack{ x\sim \hat q(x) \\ y\sim p_\theta(y|x)}}\biggl[\nabla_\theta \log p_\theta(y|x) \nabla_\theta \log p_\theta(y|x)^T\biggr].
\end{equation}
This is the Fisher of the joint distribution $p_{\theta}(x,y) = \hat q(x) p(y|h_{\theta}(x))$. We let $F(\theta_k,x) := \mathbb E_{y \sim p_\theta(y|x)}[\nabla_\theta\log p_\theta(y|x) \nabla_\theta\log p_\theta(y|x) ^T]$ be the Fisher of the conditional distribution $p_{\theta}(y|x)$. Then, we have $F_k = \mathbb E_{x\sim \hat q}[F(\theta_k,x)]$. Using the Fisher, we have the following approximation (see \cite{KFAC,AMARI}) for the SKL-Divergence
\begin{equation}
\mathbb D_{KL}\big(p_{\theta_1}(y| x), p_{\theta_2}(y| x)\big) \approx \frac{1}{2}(\theta_2-\theta_1)^T F(\theta_1,x)(\theta_2-\theta_1),
\label{KL_Div_eqn_to_ref_final_approx}
\end{equation}
which is exact as $\theta_2 \to \theta_1$. By plugging (\ref{KL_Div_eqn_to_ref_final_approx}) into (\ref{final_D_KL_we_care_about_2}) and using linearity we get 
\begin{equation}
\mathbb D_{KL}(\theta_1 , \theta_2)  \approx \frac{1}{2}(\theta_2-\theta_1)^T F_1(\theta_2-\theta_1).
\end{equation}

\subsection{Natural Gradient and K-FAC}
The natural gradient (NG) is defined as \cite{AMARI}
\begin{equation}
\nabla_{NG} f(\theta_k) = F_k^{-1}g_k.
\end{equation}
The NG descent (NGD) step is then taken to be $s^{(NGD)}_k = - \alpha_k \nabla_{NG} f(\theta_k)$, for some stepsize $\alpha_k$. NGD has favorable properties, the most notable of which is re-parametrization invariance (when the step-size is infinitesimally small) \cite{New_insights_and_perspectives}. 
The NGD step can be expressed as the solution to the quadratic problem
(see \cite{KFAC,New_insights_and_perspectives})
\begin{equation}
s^{(NGD)}_k : = \arg\min_s s^Tg_k +\frac{1}{2\alpha_k}s^TF_ks.
\label{NGD_as_quadratic_model}
\end{equation}
\subsubsection{K-FAC}
Storing and inverting the Fisher is prohibitively expensive. \textsc{k-fac} is a practical implementation of NG which bypasses this problem by approximating the Fisher as a block-diagonal\footnote{Block tri-diagonal approximation is also possible - but this lies outside our scope.} matrix, with each diagonal block further approximated as the Kronecker product of two smaller matrices \cite{KFAC}. That is, we have
\begin{equation}
F_k^{(KFAC)} := \text{blockdiag}\big(\{A^{(l)}_k \otimes \Gamma^{(l)}_k \}_{l=1,...,n_L}\big),
\label{KFAC_matrix_defn}
\end{equation}
where each block corresponds to a layer and $\otimes$ denotes the Kronecker product \cite{KFAC}. For example, for FC nets, the Kronecker factors are given by
\begin{equation}
A_k^{(l)} := \mathbb E_{x,y \sim p} [\bar a_{l-1} \bar a_{l-1}^T],\,\,\, \Gamma_k^{(l)} := \mathbb E_{x,y \sim p} [\nabla_{z_l} L \nabla_{z_l} L^T].
\label{eqn_kroenecker_factors}
\end{equation}
Note that the Kronecker factors depend on $\theta_k$ which influences both the forward and backward pass. Also, note they can be efficiently worked with since $(A^{(l)}_k \otimes \Gamma^{(l)}_k)^{-1} = [A^{(l)}_k]^{-1} \otimes [\Gamma^{(l)}_k]^{-1}$, and $\big([A^{(l)}_k]^{-1} \otimes [\Gamma^{(l)}_k]^{-1}\big) v = \text{vec}\big([\Gamma^{(l)}_k]^{-1} V [A^{(l)}_k]^{-1}\big)$, where $v$ maps to $V$ in the same way $\text{vec}(W_l)$ maps to $W_l$ \cite{KFAC}.

\subsection{Curvature in Practice: Exponential Averaging}
In practice, many algorithms (including \textsc{adam} and \textsc{k-fac}) do not use the curvature matrix estimate as computed. Instead, they maintain an exponential average (EA) of it (eg.\ \cite{ADAM,efficientbackprop,nomorepeskyLR,AdaptiveNGforVarious_stochastic_models}). In the case of NG, this EA is
\begin{equation}
\bar F_k := \rho^k F_0 + (1-\rho)\sum_{i=1}^k\rho^{k-i}F_i,
\label{EAFM_eqn}
\end{equation}
where $\rho\in[0,1)$ is the exponential decay parameter. Let us refer to NG algorithms which replace the Fisher, $F_k$, with its exponential average, $\bar F_k$, as \textsc{ea-ng}. 

In a similar spirit, \textsc{k-fac} maintains an EA of the Kronecker factors
\begin{equation}
\bar A_k^{(l)} := \rho^k A_0^{(l)} + (1-\rho)\sum_{i=1}^k\rho^{k-i}A_i^{(l)},\,\,\,\, \bar \Gamma_k^{(l)} := \rho^k \Gamma_0^{(l)} + (1-\rho)\sum_{i=1}^k\rho^{k-i}\Gamma_i^{(l)},
\label{EA_KF_1}
\end{equation}
and in practice, $ A_k^{(l)} $  and $\Gamma_k^{(l)}$ in (\ref{KFAC_matrix_defn}) are replaced with $\bar A_k^{(l)}$ and $\bar \Gamma_k^{(l)}$ respectively. We will refer to the practical implementation of \textsc{k-fac} which uses EA for the \textsc{k-fac} matrices as \textsc{ea-kfac}, to emphasize the presence of the EA aspect. However, this is the norm in practice rather than an exception, and virtually any algorithm referred to as \textsc{k-fac} (or as using \textsc{k-fac}) is in fact an \textsc{ea-kfac} algorithm.

\section{A Wake of Quadratic Models (WoQM)}
The idea behind \textsc{w}o\textsc{qm} is simple. Instead of taking a step $s_k$, at $\theta_k$ which relies only on the local quadratic model, $s_k = \arg\min_s g^Ts +(1/2)s^TB_ks$, we take a step which relies on an EA of all previous local models. Formally, let us define
\begin{equation}
M^{(Q)}_k(s) := g_k^Ts +\frac{\lambda_k}{2}s^TB_ks,
\label{eqn_quadratic_model_defn}
\end{equation}
for an arbitrary \textit{symmetric-positive definite} curvature matrix $B_k$ (typically an approximation of $H_k$ or $F_k$). Note that $1/\lambda_k$ is a step-size (or learning rate) parameter.
\noindent Our \textsc{w}o\textsc{qm} step, $s_k$, is then defined as the solution to
\begin{equation}
\min_s \sum_{i=0}^k\rho^{k-i}M^{(Q)}_i\biggl(s + \sum_{j=i}^{k-1}s_j\biggr), \,\,\,\text{with}
\label{eqn_WoQM_in_terms_of_quadratic_model_defn}
\end{equation}
\begin{equation}
\lambda_0 = \lambda,\,\,\text{and}\,\,\lambda_k = (1-\rho)\lambda,\,\,\,\forall k\in\mathbb Z^+,
\label{WoQM_defn_2_lambda}
\end{equation}
where we set $\sum_{j=k}^{k-1}s_j=0$ by convention, $\rho\in[0,1)$ is an exponential decay parameter, and $\lambda > 0$ is a hyperparameter. 
While (\ref{eqn_WoQM_in_terms_of_quadratic_model_defn}) does not appear to be a proper exponential averaging, missing a $1-\rho$ factor in terms where $i\geq 1$, it can be easily rearranged as such by slightly modifying the definition of $M^{(Q)}_0$ (to disobey (\ref{eqn_quadratic_model_defn})). To see this, multiply (\ref{eqn_WoQM_in_terms_of_quadratic_model_defn}) by $(1-\rho)$ and then absorb the $1-\rho$ factor in the definition of $M^{(Q)}_0$. Our stated definition makes the exposition more compact while preserving intuition.

For our choice of model (\ref{eqn_quadratic_model_defn}), the \textsc{w}o\textsc{qm} step $s_k$ (at $\theta_k$) is the solution of 
\begin{equation}
\begin{split}
\min_s s^T\biggl[\sum_{i=0}^k\rho^{k-i}\biggl(g_i + \lambda \kappa(i) B_i\sum_{j=i}^{k-1}s_j\biggr)\biggr] 
+\frac{\lambda}{2}s^T\biggl[\sum_{i=0}^k \kappa(i) \rho^{k-i}B_i\biggr]s,
\end{split}
\label{eqn_WoQM_for_particualr_model}
\end{equation}
where we dropped all the constant terms, and have $\kappa(i) := \exp(\mathbb I_{\{i>0\}}\log(1-\rho))$, where $\mathbb I_{\mathcal E}$ is the indicatior function of event $\mathcal E$. We use the $\kappa(i)$ term for notational compactness. Note that definition (\ref{eqn_WoQM_in_terms_of_quadratic_model_defn}) can be used for general models $M_i$ (rather than quadratic $M^{(Q)}_i$), leading to a larger family of algorithms for which \textsc{w}o\textsc{qm} represents a particular instantiation: the \textit{Wake of Models (WoM)} family.

\subsection{Connection with Exponential-Averaging in Curvature Matrices}
We now look at how the \textsc{w}o\textsc{qm} step relates to Exponential-Averaging Curvature matrices (\textsc{ea-cm}). \textsc{ea-cm} is standard practice in stochastic optimization, and in particular in training DNNs (see for example \cite{KFAC,ADAM,efficientbackprop,nomorepeskyLR,AdaptiveNGforVarious_stochastic_models}). Formally, using \textsc{ea-cm} boils down to taking a step based on (\ref{eqn_quadratic_model_defn}), but with $B_k$ replaced by
\vspace{-1.5ex}
\begin{equation}
\bar B_k:= \rho^k B_0 + (1-\rho)\sum_{i=1}^k\rho^{k-i}B_i = \sum_{i=0}^k \kappa(i) \rho^{k-i}B_i.
\end{equation}
\vspace{-2.5ex}

Note that we used the same exponential decay parameter for convenience, but this is not required.

It is obvious from (\ref{eqn_WoQM_for_particualr_model}) that we can get an analytic solution for \textsc{w}o\textsc{qm} step $s_k$, as a function of $\{s_j\}_{j=0}^{k-1}$, $\{(g_j, B_j)\}_{j=0}^k$, $\rho$ and $\lambda$. Thus, by using the relationship recursively we can get an analytic solution for $s_k$ as a function of $\{(g_j, B_j)\}_{j=0}^k$, $\rho$ and $\lambda$. When doing this, the connection between \textsc{w}o\textsc{qm} and \textsc{ea-cm} is revealed. The result is presented in \textit{Proposition 3.1}. 

\begin{Proposition_3_1}
	The \textsc{w}o\textsc{qm} step $s_k$ at iterate $\theta_k$ can be expressed as
	\begin{equation}
	s_k = -\lambda^{-1}\bar B_k^{-1}g_k,\,\,\,\,\,\forall k \in\mathbb Z^+.
	\label{eqn_proposition_3_1_eqn}
	\end{equation}
\end{Proposition_3_1} 
\textit{Proof.} Relies on simple inductive argument. See supplementary material. $\square$

\textit{Proposition 3.1} tells us that the \textsc{w}o\textsc{qm} step is exactly the step obtained by using an EA curvature matrix $\bar B_k$ in a simple quadratic model of the form (\ref{eqn_quadratic_model_defn}). That is, \textsc{w}o\textsc{qm} (\ref{eqn_WoQM_in_terms_of_quadratic_model_defn})-(\ref{WoQM_defn_2_lambda}) is the principled optimization formulation of a \textsc{ea-cm} step\footnote{\textsc{w}o\textsc{qm} with $\rho = 0$ and $B_k$ based on quantities at $\theta_k$ only is also the principled optimization formulation of no-EA CM algorithms, with steps of the form (\ref{eqn_quadratic_model_defn}).}, when the \textsc{ea-cm} stepsize is constant and equal to $1/\lambda$. Thus, we see what \textsc{ea-cm} is actually doing from an optimization perspective: instead of perfectly solving for the local quadratic model, it solves for a trade-off between all previously encountered models, where the weights of the trade-off are (almost\footnote{Can modify the definition of $M^{(Q)}_0$ s.t.\ \textsc{w}o\textsc{qm} is a proper EA of quadratic models.}) given by an exponential average (older models receive exponentially less ``attention'').

There are two observations to make at this point. First, we began by noting that the justification for \textsc{ea-cm} is largely heuristic, but we ended up explaining \textsc{ea-cm} through some optimization model which involved an EA of local models (the \textsc{w}o\textsc{qm} model). Since the EA was the difficult part to justify in the first place, it might seem that we are sweeping the problem from under one rug to another. However, this is not the case. The \textsc{w}o\textsc{qm} formulation aims to reveal a different perspective on \textsc{ea-cm} algorithms, rather than explain the presence of EA itself. It is indeed true that we did not justify why one should use EA and thus get the \textsc{w}o\textsc{qm} family, but this is not required to enhance our understanding and draw conclusions. We can draw conclusions purely based on the established equivalence. This leads us to the second observation, which is why \textsc{ea-cm} improves stability from a stochastic optimization perspective. Rather than conferring stability because it \textit{``uses more data''} (\cite{KFAC,efficientbackprop}), \textsc{ea-cm} can alternatively be thought of as conferring stability because it uses the collection of all previous noisy local models to build a better model\footnote{Although it is still not clear why putting the previous local models together in an exponentially-averaged fashion is \textit{``right''} for conferring further stability - and this aspect remains heuristic. While this could be informally and partly explained by \textit{``older models should matter less''}, the complete explanation remains an open question.} (in terms of both noise and functional form).

Note that what we have discussed so far applies when using any curvature matrix $B_k$. In particular, \textit{Proposition 3.1} can be directly applied to establish an equivalence between \textsc{ea-ng} algorithms and \textsc{fisher-w}o\textsc{qm} algorithms (\textsc{w}o\textsc{qm} algorithms with $B_k = F_k$). In fact, we can replace the Fisher by any approximation and still have the equivalence holding, \textit{if the EA is done as in (\ref{EAFM_eqn})}.

\subsection{Fisher-WoQM and Practical K-FAC Equivalence}
We have seen (in \textit{Section 2.4}) that \textsc{k-fac} holds an EA for the \textit{Kronecker factors} (see (\ref{EA_KF_1})), rather than for the \textsc{k-fac} approximation to $F_k$ (as in (\ref{EAFM_eqn})). Thus, the EA scheme employed by \textsc{k-fac} is \textit{not} the same as (\ref{EAFM_eqn}) with $F_k \leftarrow F_k^{(KFAC)}$. Therefore, we cannot directly apply \textit{Proposition 3.1} to obtain an equivalence between \textsc{ea-kfac} and \textsc{kfac-w}o\textsc{qm} (\textsc{w}o\textsc{qm} with $B_k \leftarrow F_k^{(KFAC)}$). We can loosely establish this equivalence by viewing the \textit{EA over the Kronecker factors} as a convenient (but coarse) approximation to the \textit{EA over }$F^{(KFAC)}_k$ (see \textit{Section 4} in the \textit{supplementary material}). Indeed, carrying an EA for $F^{(KFAC)}_k$ is impractical. 

Our equivalence reveals that \textsc{ea-ng} is actually solving an exponentially decaying wake of quadratic models (\textsc{w}o\textsc{qm}) where the curvature matrix (meant to be a Hessian approximation) is taken to be an approximate Fisher. This is in contrast with the typical \textsc{ea-ng} interpretation which says that we take NG steps by solving quadratic models of the form (\ref{eqn_quadratic_model_defn}) with $B_k = F_k$, but then we further approximate $F_k$ as an EA based on $\{\hat F_j\}_{j=1}^{k-1}$. While \textsc{ea-kfac} does not do the exact same thing, it can be seen as a (very crude) approximation to it.

\subsubsection{Dealing with Infrequent Updates}
In practice, the curvature matrix is not computed at each location, and the EA update is typically performed every $N_u \approx 100$ steps\footnote{More complicated heuristics can be designed, see \cite{Kazuki_SPNGD}.} to save computation cost (at least in supervised learning\footnote{ In RL we may prefer updating the \textsc{k-fac} EA-matrix at each step \cite{ACKTR}.}) \cite{KFAC}. In this circumstance, the final implementation of \textsc{ea-ng} would actually be equivalent to a \textsc{w}o\textsc{qm} algorithm where we set
\vspace{-1ex}
\begin{equation}
B_k = \begin{cases}
\hat F_k,\,\,\,\,\text{if}\,\mod(k,N_u) = 0\\
\bar B_q,\,\text{where}\,q:=N_u\big\lfloor\frac{k}{N_u}\big\rfloor,\,\text{otherwise}
\end{cases}.
\label{eqn_KFAC_Nu_steps_period_as_WoQM}
\end{equation} 
In (\ref{eqn_KFAC_Nu_steps_period_as_WoQM}), $\hat F_k$ represents an approximation for the Fisher $F_k$, whose computation has an associated cost.
Recall that $\bar B_k = \sum_{j=0}^{k}\kappa(j)\rho^{k-j}B_j$. Note that (\ref{eqn_KFAC_Nu_steps_period_as_WoQM}) is well defined since $\bar B_0= B_0 = \hat F_0$, then $B_1 = B_2 = ... = B_{N_u-1} = B_0$, and hence $\bar B_1 = \bar B_2 =...= \bar B_{N_u-1}= B_0$, and so on. Further note that by (\ref{eqn_proposition_3_1_eqn}), defining our \textsc{w}o\textsc{qm} curvature matirx as in (\ref{eqn_KFAC_Nu_steps_period_as_WoQM}) gives us the \textsc{ea-ng} matrix that we want, since we can easily see that: $\bar B_{N_u} = \rho \bar B_0 + (1-\rho)\hat F_{N_u} = \rho \hat F_{0}+ (1-\rho)\hat F_{N_u}$. We can then easily extend the argument to show $\bar B_{q N_u} = \sum_{j=0}^{q}\kappa(j)\rho^{q-j}\hat F^{(KFAC)}_{jN_u}$ which is exactly the form of EA that \textsc{ea-ng} employs when updating statistics every $N_u$ steps. Note that we can apply \textit{Proposition 3.1} irrespectively of our choice of $B_k$, so in particular, it must hold for $B_k$ as defined in (\ref{eqn_KFAC_Nu_steps_period_as_WoQM}).

By extending our reasoning, we see that any heuristic which adapts $N_u$ as a function of observations up until $k$ can be transformed into an equivalent heuristic of picking between $B_k = \bar B_q$ (where $q$ is now more generally the previous location where we computed $\hat F_k$) and $B_k = \hat F_k$. Thus, the equivalence between \textsc{w}o\textsc{qm} and \textsc{ea-ng} holds irrespectively of the heuristic which decides when to update the \textsc{ea-ng} matrix. The exact same reasoning holds for generic \textsc{ea-cm} algorithms.

\subsection{Fisher-WoQM: A Different Perspective}
Since we have the following approximation\footnote{Which is exact in the limit as $\theta_k +s \to \theta_i$, but would nevertheless be very crude in practice, particularly for large $k-i$, since the steps taken might be relatively large.} for $k \geq i$ \cite{New_insights_and_perspectives}
\vspace{-1ex}
\begin{equation}
\mathbb D_{KL}(\theta_i, \theta_{k} + s) \approx \tilde{\mathbb D}_{KL}(\theta_i, \theta_{k} + s): = \frac{1}{2}\biggl(s + \sum_{j=i}^{k-1}s_j\biggr)^T F_i \biggl(s + \sum_{j=i}^{k-1}s_j\biggr),
\label{eqn_very_crude_approx_D_KL_WoQM}
\end{equation}
one might think that a \textsc{w}o\textsc{qm} model with $B_k = F_k$ and $\rho\in(0,1)$ would in fact be some form of approximate \textit{``KL-Divergence Wake-Regularized\footnote{Regularizing w.r.t.\ SKL divergences relative to all previous distributions, as opposed to just the most recent one - as the quadratic model associated with NG step does.} model''}. This is indeed true, as we can write our \textsc{fisher-w}o\textsc{qm} as
\vspace{-1ex}
\begin{equation}
\min_s \biggl[\sum_{i = 0}^k \rho^{k-i}g_i\biggr]^Ts + \lambda\sum_{i=0}^k\kappa(i)\rho^{k-i}\tilde{\mathbb D}_{KL}(\theta_i, \theta_{k} + s).
\label{WoQM_as_particular_case_KLD_WRM}
\end{equation}
{Note that $\tilde{\mathbb D}_{KL}(\theta_i, \theta_{k} + s)$ in (\ref{eqn_very_crude_approx_D_KL_WoQM}) is a second-order approximation of the \unskip\parfillskip 0pt \par }
\newpage 
\noindent SKL-Divergence between $p_{\theta_i}(x,y)$ and $p_{\theta_k+s}(x,y)$. Thus, we see that \textsc{fisher-w}o\textsc{qm} (same as \textsc{ea-ng} by \textit{Prop.\ 3.1}) is in fact solving a \textit{regularized linear model}, where the model gradient is taken to be the \textit{momentum gradient} (with parameter $\rho$), and the regularization term is an exponentially decaying wake of \textit{(crudely) approximate} SKL-divergences relative to previously encountered distributions. 

This equivalence raises scope for a new family of algorithms: perhaps using another model for the objective $f$, rather than a simple linear model based on momentum-gradient, and/or using a better approximation for the KL divergence could lead to better performance. This is the main topic of this paper, explored formally in \textit{Section 4} and numerically in \textit{Section 5}. We now note that \textsc{w}o\textsc{qm} (and thus also \textsc{ea-ng} by \textit{Proposition 3.1}, and approximately so, \textsc{ea-kfac}) is in fact a particular instantiation of the family proposed in the next section.

\section{A KL-Divergence Wake-Regularized Models Approach}
We now propose a new family of algorithms, which we call \textit{KL-Divergence Wake-Regularized Models (\textsc{kld-wrm}; reads: \textit{``Cold-Worm''})}. At each location $\theta_k$, the \textsc{kld-wrm} step $s_k$ is defined as the solution to the problem
\begin{equation}
\min_s M(s; \mathcal F_k) + \lambda\sum_{i=0}^k\zeta(i)\rho^{k-i}{\mathbb D}_{KL}(\theta_i,\theta_{k} + s),
\label{eqn_most_general_KLDWRM}
\end{equation}
where $\rho\in[0,1)$, $M(s; \mathcal F_k)$ is a model of the objective which uses \textit{at most} all the information ($\mathcal F_k$) encountered up until and including $\theta_k$, and $\zeta(i)$ allows for different '$\lambda$\textit{'s}' at different $i$'s. Simple choices would be $\zeta(i)\equiv 1$, or $\zeta(i)=\kappa(i)$.

The motivation behind \textsc{kld-wrm} is two-fold. First, a wake of SKL regularization allows us to stay close (in a KL sense) to all previously encountered distributions, rather than only to the most recent one. Thus, we might expect \textsc{kld-wrm} to give more conservative steps in terms of distribution ($p_\theta(y|x)$) change. Second, \textsc{kld-wrm} can be seen as a generalization\footnote{The linear model $\big[\sum_{i = 0}^k \rho^{k-i}g_i\big]^Ts$ in (\ref{WoQM_as_particular_case_KLD_WRM}) becomes arbitrary, and $\kappa(i)$ is replaced by a general $\zeta:\mathbb Z^+\to \mathbb R$.} of \textsc{ea-ng} (which is also \textsc{fisher-w}o\textsc{qm}), which also \textit{``undoes''} the approximation\footnote{For small $\norm{s}$ we have $\mathbb D(\theta, \theta+s) \approx\mathbb D(\theta||\theta+s) \approx\mathbb D(\theta + s|| \theta) \approx (1/2)s^TF(\theta) s$. Thus, the generalization towards our family could use $\xi\mathbb D(\theta||\theta+s) + (1-\xi) \mathbb D(\theta + s|| \theta)$ $\forall \xi\in\mathbb R$, instead of the SKL (i.e.\ $\xi = 1/2$). We choose SKL for simplicity.} of SKL. 
 
Note that (\ref{eqn_most_general_KLDWRM}) is the most general formulation of \textsc{kld-wrm}, but in order to obtain practical algorithms we have to make further approximations and definitions (of $M$ and $\zeta$). For example, we could set
\begin{equation}
M(s; \mathcal F_k) = \sum_{i=0}^k\nu^{k-i}M_i\biggl(s + \sum_{j=i}^{k-1}s_j\biggr)
\label{eqn_WoM_KLDWRM}
\end{equation}
where the models $M_i(s)$ are general models (not necessary quadratic), constructed only based on local information at $\theta_i$. This would be a \textit{Wake of Models \textsc{kld-wrm}} \textsc{(w}o\textsc{m-kld-wrm)}. We could make this even more particular, for example by considering instantiations where $\nu = 0$, but $\rho\in[0,1)$ in (\ref{eqn_WoM_KLDWRM}). This would give a \textit{Local-Model} \textsc{kld-wrm} \textsc{(lm-kld-wrm)}, whose steps $s_k$ are the solution to
\begin{equation}
\min_s M_k(s) + \lambda\sum_{i=0}^k\zeta(i)\rho^{k-i}{\mathbb D}_{KL}(\theta_i, \theta_{k} + s),
\label{eqn_particular_KLDWRM}
\end{equation}
In this paper, we focus on \textsc{lm-kld-wrm}  instantiations (a particular sub-family of \textsc{kld-wrm}), and leave the general case as future work. We give three instantiations of \textsc{lm-kld-wrm}  of increasing complexity, and discuss the links with already existing methods. We investigate their performance in \textit{Section 5}.
\subsection{Connection between KLD-WRM and Fisher-WoQM}
It is easy to see that setting $\nu=\rho$, $\zeta(i)=\kappa(i)$, $M_i(s) = s^Tg_i$ and approximating ${\mathbb D}_{KL}(\theta_i,\theta_{k} + s)\approx \tilde{\mathbb D}_{KL}(\theta_i, \theta_{k} + s)$ (defined in (\ref{eqn_very_crude_approx_D_KL_WoQM})) in a \textsc{w}o\textsc{m-kld-wrm} gives the \textsc{w}o\textsc{qm} family. Note that \textsc{w}o\textsc{qm} is \textit{not} an \textsc{lm-kld-wrm} model as $M_k(s)$ can only include information local to $\theta_k$ (eg. cannot include $g_{k-1}$). However, the simplest instantiation of \textsc{lm-kld-wrm}  takes steps which are formally similar to \textsc{fisher-w}o\textsc{qm} (and thus \textsc{ea-ng}) steps. We investigate this in \textit{Section 4.2}.

\subsection{Simplest KLD-WRM Instantiation: Smallest Order KLD-WRM}
The simplest practical instantiation of \textsc{kld-wrm}, \textit{Smallest Order \textsc{kld-wrm}} (\textsc{so-kld-wrm}), uses the most crude approximations to \textsc{lm-kld-wrm}  (\ref{eqn_particular_KLDWRM}) and sets $\zeta(i) = \kappa(i)$. The \textsc{so-kld-wrm} step $s_k$ (at $\theta_k$) is given by
\begin{equation}
\min_s g_k^Ts + \lambda\sum_{i=0}^k\kappa(i)\rho^{k-i}\tilde{\mathbb D}_{KL}(\theta_i, \theta_{k} + s),
\label{eqn_SO_KLD_WRM}
\end{equation}
It is trivial to see that \textsc{so-kld-wrm} differs from \textsc{fisher-w}o\textsc{qm} (\ref{WoQM_as_particular_case_KLD_WRM}) only through replacing $\big[\sum_{i = 0}^k \rho^{k-i}g_i\big]$ with $g_k$. Since \textsc{fisher-w}o\textsc{qm} is equivalent to \textsc{ea-ng}, one might expect that \textsc{so-kld-wrm} steps are \textit{formally} similar to \textsc{ea-ng} steps. \textit{Proposition 4.1} formalizes this result.

\begin{Proposition_4_1}
	The \textsc{so-kld-wrm} step $s_k$ at iterate $\theta_k$ can be expressed as 
	\begin{equation}
	s_k = -\lambda^{-1}\bar F_k^{-1}[g_k-\rho g_{k-1}],
	\label{eqn_proposition_4_1_eqn}
	\end{equation}
	$\forall k \in\mathbb Z^+$, where $\bar F_k := \sum_{i=0}^k\kappa(i)\rho^{k-i}F_i$ and we set $g_{-1}:=0$ by convention.
\end{Proposition_4_1} 

\textit{Proof.} By induction. See the \textit{supplementary material}. $\square$

Note that we set  $g_{-1}=0$ to avoid providing two separate cases (for $k=0$ and for $k\geq1$). \textit{Proposition 4.1} tells us that the \textsc{so-kld-wrm} step is \textit{formally similar} to the \textsc{fisher-w}o\textsc{qm} step (which we have seen is the \textsc{ea-ng} step). The only (formal) difference between the \textsc{so-kld-wrm} step and the \textsc{ea-ng} step is that $g_k$ gets replaced by $g_{k}-\rho g_{k-1}$ in (\ref{eqn_proposition_3_1_eqn}). By \textit{``formally''} here, we mean that the expressions look very similar. However, when considering the two implemented algorithms, the paths taken can be very different.
\newpage

We have seen that \textsc{ea-ng} (being equivalent to \textsc{fisher-w}o\textsc{qm})  algorithms are in fact a sub-family of the \textsc{kld-wrm} family, and obviously \textsc{so-kld-wrm} is also a sub-family of \textsc{kld-wrm}. Thus, the formal difference in the steps between \textsc{so-kld-wrm} and \textsc{ea-ng} tells us these two sub-families are distinct. The formal similarity between \textsc{so-kld-wrm} and \textsc{ea-ng}, combined with the fact that both are members of the \textsc{kld-wrm} family raises hopes that more accurate instantiations \textsc{kld-wrm} might lead to better performance than \textsc{ea-ng}.

Note that basing our \textsc{so-kld-wrm} step computation on \textit{Proposition 4.1} gives a tractable algorithm, while solving (\ref{eqn_SO_KLD_WRM}) directly gives an intractable algorithm. To see this, review definition (\ref{eqn_very_crude_approx_D_KL_WoQM}), and realize that solving equation (\ref{eqn_SO_KLD_WRM}) \textit{directly} requires storing all previous $\bar F_js_j$ matrix-vector products as a minimum (if the algorithm is written efficiently). Thus, we have an exploding number of vectors that need to be stored when solving  (\ref{eqn_SO_KLD_WRM}) \textit{directly}, which eventually will overflow the memory - giving an untractable algorithm. Conversely, using \textit{Proposition 4.1} only requires storing at most 2 matrices ($\bar F_k$ and $F_k$) and 2 gradient-shaped vectors at any one moment in time. Thus, using \textit{Proposition 4.1} with tractable approximations for $F_k$ gives a tractable algorithm.

\subsubsection{Reconsidering Gradient Momentum in EA-KFAC}
By comparing (\ref{WoQM_as_particular_case_KLD_WRM}) and (\ref{eqn_SO_KLD_WRM}), we see that \textsc{fisher-w}o\textsc{qm} (which is also \textsc{ea-ng}) can \textit{almost} be seen as \textsc{so-kld-wrm} with added momentum for the gradient. The equivalence would be exact if we would have a $\kappa(i)$ inside the sum of the linear term of (\ref{WoQM_as_particular_case_KLD_WRM}). In \textsc{ea-kfac}, gradient momentum is not added in the standard fashion. A different way to add momentum is proposed\footnote{More akin to a subspace method rather than a standard momentum method (see \cite{KFAC}).} and presented as successful, perhaps because trying to add gradient momentum in the standard fashion gives worse performance \cite{KFAC}. From our discussion, this could be because \textsc{ea-kfac} can be interpreted as an approximate \textsc{ea-ng}, and \textsc{ea-ng} is a \textsc{so-kld-wrm} algorithm which already has included gradient momentum. Thus, further adding momentum does not make sense. Note that by adding momentum to gradient in the \textit{standard fashion} we mean replacing $g_k$ by $\sum_{i=0}^k\kappa(i)\rho^{k-i}g_k$ (see \cite{ADAM}).

\subsubsection{SO-KLD-WRM in practice}
When implementing \textsc{so-kld-wrm} in practice, we use the \textsc{k-fac} approximation of the Fisher. Note that \textit{Proposition 4.1} tells us that we need \textit{not} store all previous \textsc{k-fac} matrices. Instead, we can only save the \textsc{ea-kfac} matrix. As we have discussed in \textit{Section 3}, we can skip computing the \textsc{k-fac} matrix at some locations to save on computation. To do that, we just pretend the new \textsc{k-fac} matrix at $\theta_k$ is the \textsc{ea-kfac} that we currently have stored. For example, if we want to compute \textsc{k-fac} matrices only once in $N_u$ steps, we use (\ref{eqn_KFAC_Nu_steps_period_as_WoQM}), but different heuristics can also be used (for eg.\ as in \cite{Kazuki_SPNGD}). Of course, in practice we store an EA for the Kronecker factors (instead of an EA for the \textsc{k-fac} matrix), as is standard with practical \textsc{k-fac} implementation \cite{KFAC}.

\subsection{Second KLD-WRM Instantiation: Q-KLD-WRM}
The \textit{Quadratic \textsc{kld-wrm} \textsc{(q-kld-wrm)}} is an \textsc{lm-kld-wrm} instantiation which uses a second-order approximation for both the Model and the SKLs, and sets $\zeta(i)=\kappa(i)$. The \textsc{q-kld-wrm} step $s_k$ (at $\theta_k$) solves
\begin{equation}
\min_s g_k^Ts + \frac{1}{2}s^TB_ks +  \lambda\sum_{i=0}^k\kappa(i)\rho^{k-i}\tilde{\mathbb D}_{KL}(\theta_i, \theta_{k} + s),
\label{eqn_Q_KLD_WRM}
\end{equation}
where $B_k$ is a curvature matrix which aims to approximate the Hessian $H_k$. That is, \textsc{q-kld-wrm} sets $M_k$ in (\ref{eqn_particular_KLDWRM}) to be a quadratic model. Since (\ref{eqn_Q_KLD_WRM}) is overall a quadratic, we can obtian an \textit{analytic solution} for the \textsc{q-kld-wrm} step. 

\begin{Proposition_4_2}
	The \textsc{q-kld-wrm} step $s_k$ at location $\theta_k$ is given by the solution to the problem
	\begin{equation}
	\min_s s^T\hat g_k + \frac{\lambda}{2}s^T\biggl[\bar F_k + \frac{1}{\lambda}B_k\biggr]s
	\label{eqn_prop_4_2_result_0}
	\end{equation}
	where $\hat g_k$ is given by the one-step recursion
	\begin{equation}
	\hat g_{k+1} = g_{k+1} + \rho(I-\hat M_k)\hat g_{k}- \rho g_k,\,\,\,\,\,\forall k\in \mathbb Z^+,
	\label{eqn_prop_4_2_result_2}
	\end{equation}
	\noindent with $\hat g_0 := g_0$, $\bar F_k := \sum_{i=0}^k\kappa(i)\rho^{k-i}F_i$, and $\hat M_k :=\big[I +\frac{1}{\lambda}B_k \bar F_k^{-1}\big]^{-1}$. That is, the \textsc{q-kld-wrm} step is formally given by $s_{k}= -\frac{1}{\lambda}\big[\bar F_k + \frac{1}{\lambda}B_k\big]^{-1}\hat g_k$.
\end{Proposition_4_2}
\textit{Proof.} By induction. See the \textit{supplementary material}. $\square$

By comparing \textit{Propositions 4.1} and \textit{4.2}, we see that, unlike \textsc{so-kld-wrm}, the \textsc{q-kld-wrm} step deviates significantly from the \textsc{fisher-w}o\textsc{qm} step (also the \textsc{ea-ng} step). Note that setting $B_k=0$ in \textit{Proposition 4.2} gives \textit{Proposition 4.1}. That is,  \textsc{so-kld-wrm} is a particular case of \textsc{q-kld-wrm} (with $B_k\equiv0$).

Note that $\{\hat g_k\}$ could explode with $k$, leading to divergence. A sufficient condition for $\{\hat g_k\}$ to stay bounded is $\rho\norm{I-\hat M_k}_2\leq\delta$, $\forall k\in\mathbb Z^+$, with $\delta\in(0,1)$ and that $\norm{\nabla f(\theta)}_2\leq K_g$, $\forall \theta$. Under this condition, one can see that $\norm{\hat g_{k}}_2\leq \norm{g_k - \rho g_{k-1}}_2+\delta \norm{\hat g_{k-1}}_2$. Applying this inequality recursively, and noting that $\norm{g_k - \rho g_{k-1}}_2\leq (1+\rho)K_g$, we get that our sufficient condition yields $\norm{\hat g_k}_2\leq \frac{2}{1-\delta}K_g$, $\forall k\in\mathbb Z^+$. The condition $\rho\norm{I-\hat M_k}_2 = \rho \norm{I - \big[I +\frac{1}{\lambda}B_k \bar F_k^{-1}\big]^{-1}}_2\leq\delta$ can always be achieved in practice, since taking $\lambda\to\infty$ or $\rho\to0$ gives $\delta = 0$.

In a similar fashion to the role of \textit{Proposition 4.1}, the role of \textit{Proposition 4.2} (besides highlighting any similarity or dissimilarity to \textsc{ea-ng}) is to give a tractable algorithm. Again, as with \textsc{so-kld-wrm}, \textsc{q-kld-wrm} is not tractable if we implement it by solving (\ref{eqn_Q_KLD_WRM}) directly for the same reasons. On the other hand, implementing \textit{Proposition 4.2} requires simultaneous storage of \textit{at most} 3 matrices and 3 gradient-shaped vectors at any one point in time - that is, the storage cost does not explode with $k$. However, because we now have two different sets of matrices involved: $\{F_k\}$ and $\{B_k\}$, the situation is more subtle. In particular, if $B_k$ and $F_k$\footnote{Of course, $\bar F_k$ will have the same structure as $F_k$.} are block-diagonal, then we can see that all involved matrices are block-diagonal, and thus tractably storable and invertable (eg.\ choose $B_k = F_k$, and approximate $F_k\approx \hat F_k^{(KFAC)}$\footnote{Using \textsc{k-fac} further reduces the storage and computation cost through the K-factors.}). On the contrary, $B_k$ and $F_k$ might be tractably storable and invertible in isolation, but if they have different structures, computing $s_k$ from \textit{Proposition 4.2} might be intractable.

\subsubsection{Q-KLD-WRM in practice}
In practice, we can employ one of two approximations for $B_k$. The first option is to use a \textit{BFGS} approximation (\cite{BFGS}, \cite{KFAC_BFGS_paper}). The second option is to replace $B_k$ by the \textsc{k-fac} matrix. This latter approximation can be justified through the qualified equivalence between the \textit{Fisher} and the \textit{Generalized Gauss-Newton (GGN)} matrix, the latter of which is an approximation to the Hessian. However, in order for the qualified equivalence to hold, we need our predictive distribution $p(y|h_\theta(x))$ to be in the exponential family with natural parameters $h_\theta(x)$ (thinking of each conditional $y|x$ as a different distribution with its own parameters; see \cite{New_insights_and_perspectives}). This qualified equivalence holds for most practically used models (see \cite{New_insights_and_perspectives}), so is not of huge practical concern.

As we have discussed, it is not obvious how one could get a tractable algorithm from \textit{Proposition 4.2} if the structures of $B_k$ and $F_k$ are dissimilar. Thus, in this paper we focus on instantiations where $B_k:=F_k\approx F_k^{(KFAC)}$. In our experiments, we will choose $p(y|h_\theta(x))$ s.t.\ the qualified equivalence between Fisher and GGN matrix holds, and thus use $B_k \approx F^{(KFAC)}_{k}$, $F_k \approx F^{(KFAC)}_{k}$. As is typical with \textsc{k-fac}, we also maintain an EA for the Kronecker factors, rather than for the block-diagonal matrix. With these choices, the \textsc{q-kld-wrm} step can be efficiently computed form \textit{Proposition 4.2} (see the \textit{supplementary material}).

\subsection{Third KLD-WRM Instantiation: QE-KLD-WRM}
The \textit{Quadratic Objective Approximation Exact SKL\footnote{Note there is no \textit{tilde} on $\mathbb D$ in (\ref{eqn_QE_KLD_WRM}}} \textsc{kld-wrm} (\textsc{qe-kld-wrm}) is the final instantiation of \textsc{lm-kld-wrm} which we propose here. The \textsc{qe-kld-wrm} step $s_k$ (at $\theta_k$) solves
\begin{equation}
\min_s g_k^Ts + \frac{1}{2}s^TB_ks +  \lambda\sum_{i=0}^k\zeta(i)\rho^{k-i}{\mathbb D}_{KL}(\theta_i, \theta_{k} + s),
\label{eqn_QE_KLD_WRM}
\end{equation}
where $B_k$ is a curvature matrix at $\theta_k$, treated the same as we did in \textsc{q-kld-wrm}. 
\subsubsection{Practical QE-KLD-WRM for Regression}
To be able to work with the exact SKL, we restrict ourselves to a class of $p_{\theta}(y|x)$ models where the SKL can be expressed in terms of euclidean norms of differences in the network output space. Consider equation (\ref{final_D_KL_we_care_about_2}). For predictive distributions of the form (which are used in regression)
\begin{equation}
p(y|h_{\theta_k}(x_j)) = \mathcal N(y|h_{\theta_k}(x_j); I),
\label{predictive_distribution_4_3}
\end{equation}
we have a special form for $D_{KL}\big(p(y|h_{\theta_1}(x_j)) , p(y|h_{\theta_2}(x_j))\big)$, namely
\begin{equation}
\mathbb D_{KL}\big(p(y|h_{\theta_1}(x_j)) , p(y|h_{\theta_2}(x_j))\big) = \frac{1}{2}\norm{h_{\theta_1}(x_j) - h_{\theta_2}(x_j) }_2^2.
\end{equation}
See the \textit{supplementary material} for derivation details. Note that predictive distributions of the form (\ref{predictive_distribution_4_3}) are the most frequently used in practice with regression. 
\noindent For this choice of predictive distribution, ${\mathbb D}_{KL}(\theta_1,\theta_2)$ becomes
\begin{equation}
{\mathbb D}_{KL}(\theta_1,\theta_2) = \frac{1}{2N}\sum_{j=0}^N\norm{h_{\theta_1}(x_j) - h_{\theta_2}(x_j) }_2^2.
\end{equation}
Thus, the \textsc{qe-kld-wrm} step solves 
\begin{equation}
\begin{split}
\min_s g_k^Ts + \frac{1}{2}s^TB_ks +  \frac{\lambda}{2N}\sum_{i=0}^k\zeta(i)\rho^{k-i}\sum_{j=0}^N\norm{h_{\theta_i}(x_j) - h_{\theta_k + s}(x_j) }_2^2.
\end{split}
\label{eqn_QE_KLD_WRM_instantiated}
\end{equation}
\subsubsection{Practical QE-KLD-WRM for Classification} A similar practical computation of $\mathbb D_{KL}$ is also available for classification (see the \textit{supplementary material}).

\subsubsection{QE-KLD-WRM in Practice}
While the KL-regularization term in (\ref{eqn_QE_KLD_WRM_instantiated}) is in principle computable, the amount of associated storage would explode with $k$ (storing old parameters). To bypass this problem, we have two options. We could either choose to discard very old regularization terms, or model them through the approximation (\ref{eqn_very_crude_approx_D_KL_WoQM}) (as we did for \textsc{q-kld-wrm}, but now only do so for old terms). The latter approach, while convenient, is not well principled - we should really use second-order approximation when we are close (so for recent distributions), not when we are far away. This can be improved, but is left as future work. In this paper, we focus on implementations that use the former approach. Note that we now need to iteratively solve (\ref{eqn_QE_KLD_WRM_instantiated}) (for eg.\ with \textsc{sgd}), and get an approximate \textsc{qe-kld-wrm} step. Further note that the \textsc{q-kld-wrm} step given by \textit{Proposition 4.2} is an approximation of the solution to (\ref{eqn_QE_KLD_WRM_instantiated}), where the SKL-divergence is approximated by its second-order Taylor expansion (\ref{eqn_very_crude_approx_D_KL_WoQM}). Thus, the \textsc{q-kld-wrm} step is a good initial guess for (\ref{eqn_QE_KLD_WRM_instantiated}), and we exploit this fact in practice.

\subsubsection{Extension to Variable Stepsize} 
For \textsc{w}o\textsc{qm}, \textsc{so-kld-wrm} and \textsc{q-kld-wrm}, we have so far considered only cases when the \textit{``step-size''} $1/\lambda$ is fixed across different locations $\theta_k$. Indeed, our established equivalence between \textsc{fisher-w}o\textsc{qm} and \textsc{ea-ng} holds for fixed $\lambda$ only. However, we may desire variable $\lambda\leftarrow\lambda^{(k)}$ in our \textsc{kld-wrm} algorithms. To incorporate variable $\lambda^{(k)}$ in \textsc{q-kld-wrm}, all one has to do is to merely replace (\ref{eqn_prop_4_2_result_2}) with $\hat g_{k+1} = g_{k+1} + (\lambda^{(k+1)}/\lambda^{(k)})\rho[\hat g_{k}-g_k - \hat M_k\hat g_{k}]$ and all $\lambda$'s in \textit{Proposition 4.2} with $\lambda^{(k)}$. The version of \textit{Proposition 4.2} which includes variable $\lambda^{(k)}$ can be found in the \textit{supplementary material}. \textit{Proposition 4.1} with variable $\lambda^{(k)}$ then follows trivially as a particular case with $B_k\equiv0$.

\subsection{Connection with Second Order Methods (SOM)} 
The particular \textsc{kld-wrm} instantiation in equation (\ref{eqn_Q_KLD_WRM}), most simply illustrates the connection between our proposed class of algorithms and SOM. Setting $\lambda\leftarrow 0$ in (\ref{eqn_Q_KLD_WRM}) reverses \textsc{q-kld-wrm} back to a simple second order algorithm. More generally, from (\ref{eqn_most_general_KLDWRM}) we see that relative to SOM, \textsc{kld-wrm} generalizes the local second order model to an arbitrary model that may also include previous information (in principle), and more importantly, it adds a wake of $\mathbb D_{KL}$ regularization. The connection between NG and Generalized Gauss-Newton can be found in \cite{New_insights_and_perspectives}. 
\section{Numerical Results}

We compare our proposed \textsc{kld-wrm} instantiations (\textsc{so}, \textsc{q} and \textsc{qe}) with \textsc{k-fac}, on the MNIST classification problem. We investigated 4 different hyper-parameter settings for each solver, but only present the best ones here. The complete results, implementation details, as well as more in depth discussion can be found in \textit{Section 8} of the \textit{supplementary material}. \textit{Figures 1 and 2} show \textit{test loss} and \textit{test accuracy} for our considered solvers. Ten runs were performed for each solver. \textit{Table 1} shows important summary statistics of these result. 

\textbf{Hyper-parameters:} The values of $\rho$ and $\lambda$ are specified in \textit{Figures 1} and \textit{2}. We used $\zeta(i)=\kappa(i)/330$ for \textsc{qe}, and $\zeta(i)=\kappa(i)$ for the other solvers, because the \textsc{qe} estimates of the SKL term were larger. The \textsc{qe-}specific hyper-parameters are: $\omega/\lambda$ -- the learning rate of the inner SGD solving (\ref{eqn_QE_KLD_WRM_instantiated}), $N_{\text{IS}}$ -- the number of inner SGD steps per iteration, and $N_{\text{CAP}}$ -- the total number of networks stored. For the results presented here, these were set to $7\cdot10^{-4}$, $10$ and $4$ respectively. 

\textbf{Test Accuracy and Loss:} All \textsc{kld-wrm} variants exceed $97.5\%$ mean test accuracy, outperforming \textsc{k-fac} by about $1.5\%$. The SD is 4-5 times lower for \textsc{kld-wrm} variants, which is desirable.  Analogous observations hold for the test loss.  Since higher variance may more frequently yield \textit{``favorable''} outliers, we show relevant metrics for this aspect in \textit{Columns 2-6} of \textit{Table 1}. We see that even from the \textit{favorable outliers} point of view, \textsc{kld-wrm} is mostly preferable.

{\textbf{Robustness and Overall Performance:} \textit{If we can only run the training a few times (perhaps once)}, it is preferable (in terms of epochs; both from a \textit{test} \textit{accuracy} and \textit{test-loss} point of view) to use \textsc{kld-wrm} rather than \textsc{k-fac}. \unskip\parfillskip 0pt \par }%

\begin{table}[b]
	\caption{MNIST results summary. \textsc{so}, \textsc{q} and \textsc{qe} refer to \textsc{kld-wrm} variants. \textit{``Accuracy''} and \textit{``loss''} are the test-set ones. $\mu_{acc}$ and $\sigma_{acc}$ are the mean and SD of the empirical distribution of \textit{accuracy} at the end of epoch 50. Notation is analogous for  $\mu_{loss}$ and $\sigma_{loss}$, which refer to the \textit{loss}. $\mathcal N_{\mathcal C}$ is the number of runs for which condition $\mathcal C$ is satisfied. }
	\label{MNIST_results_table}
	\centering

	\begin{tabular}{|c|c|c|c|c|c|c|c|c|c|}
		\hline
		\makecell{} & \makecell{ $\mathcal N_{acc\geq98\%}$} &\makecell{$\mathcal N_{acc>98\%}$} & \makecell{$\mathcal N_{acc\geq98.5\%}$} &  \makecell{$\mathcal N_{loss\leq0.25}$} & \makecell{$ \mathcal N_{loss\leq0.2}$} & $\mu_{acc}$ &\makecell{$\sigma_{acc}$} & \makecell{$\mu_{loss}$} & \makecell{$\sigma_{loss}$} \\
		\hline
		
		\textsc{k-fac} & 3 & 3 & \textbf{1} & 5 & \textbf{2} & $96.19\%$ & $3.2\%$ & 0.26 & 0.10\\
		\hline
		
		\textsc{so} & 4 & 2 & 0 & 7 & 0 & $97.60\%$ & $0.85\%$ & 0.25 & 0.04\\
		\hline
		
		\textsc{q} & 5 & \textbf{4} & 0 & 4 & 0 & $97.69\%$ & $0.69\%$& 0.26 & 0.04\\
		\hline
		
		\textsc{qe}  \ &\textbf{8} & \textbf{7} & \textbf{2} &\textbf{9}& 0 & $\mathbf{98.01\%}$ & $\mathbf{0.64\%}$ & \textbf{0.23} & \textbf{0.02}\\
		
		\hline
	\end{tabular}		
\end{table}
\newpage 
\begin{figure}[t]
	\centering
	\includegraphics[trim={0.0cm 0.1cm 0.0cm 0.95cm},clip,width=0.3125\textwidth]{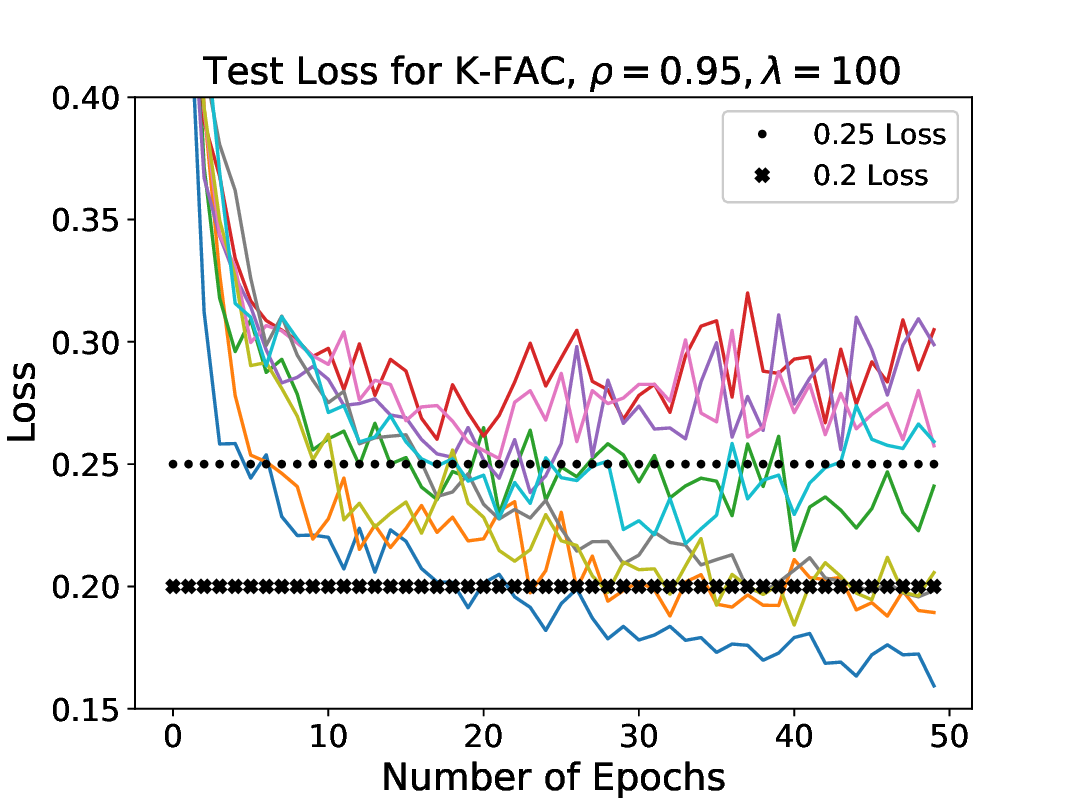}
	\includegraphics[trim={0.0cm 0.1cm 0.0cm 0.95cm},clip,width=0.3125\textwidth]{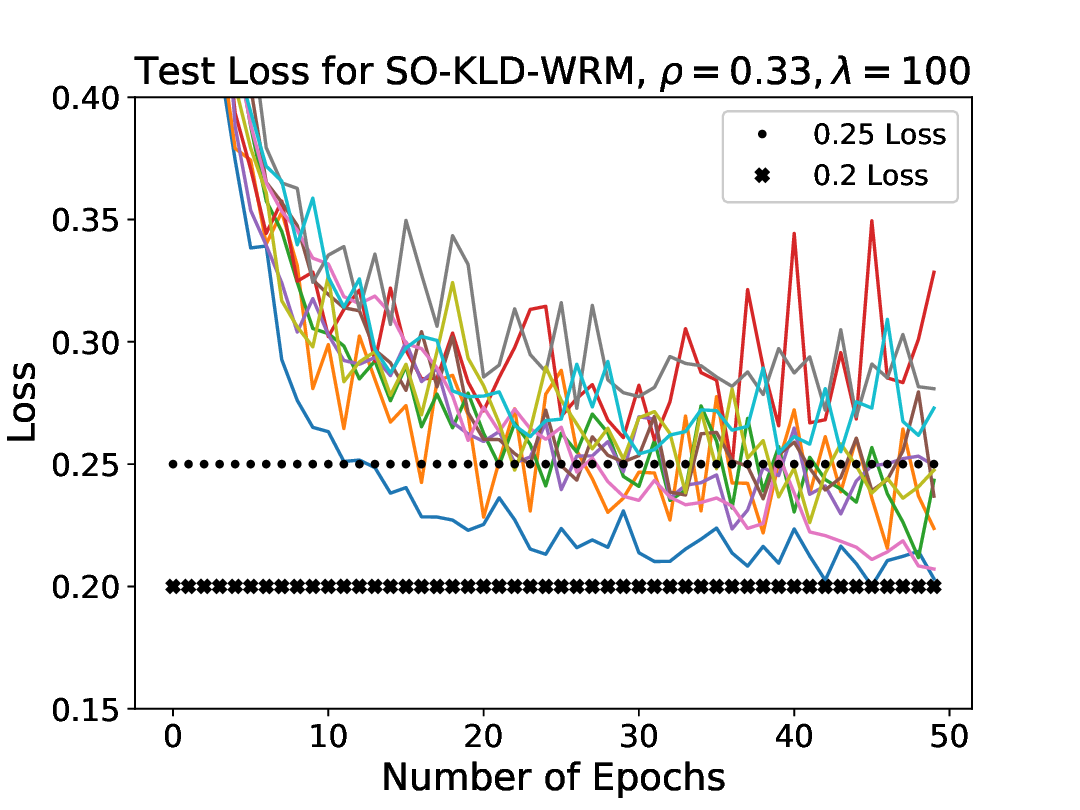}
	
	\includegraphics[trim={0.0cm 0.25cm 0.0cm 0.9cm},clip,width=0.3125\textwidth]{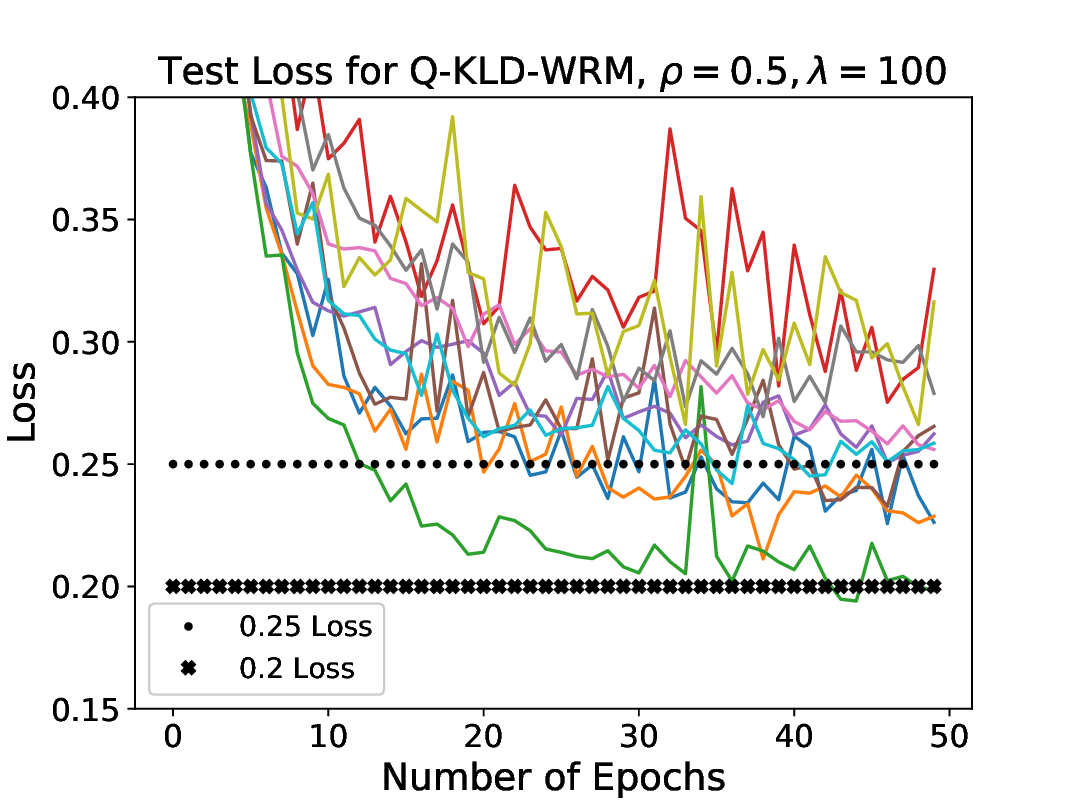}
	\includegraphics[trim={0.0cm 0.25cm 0.0cm 0.9cm},clip,width=0.3125\textwidth]{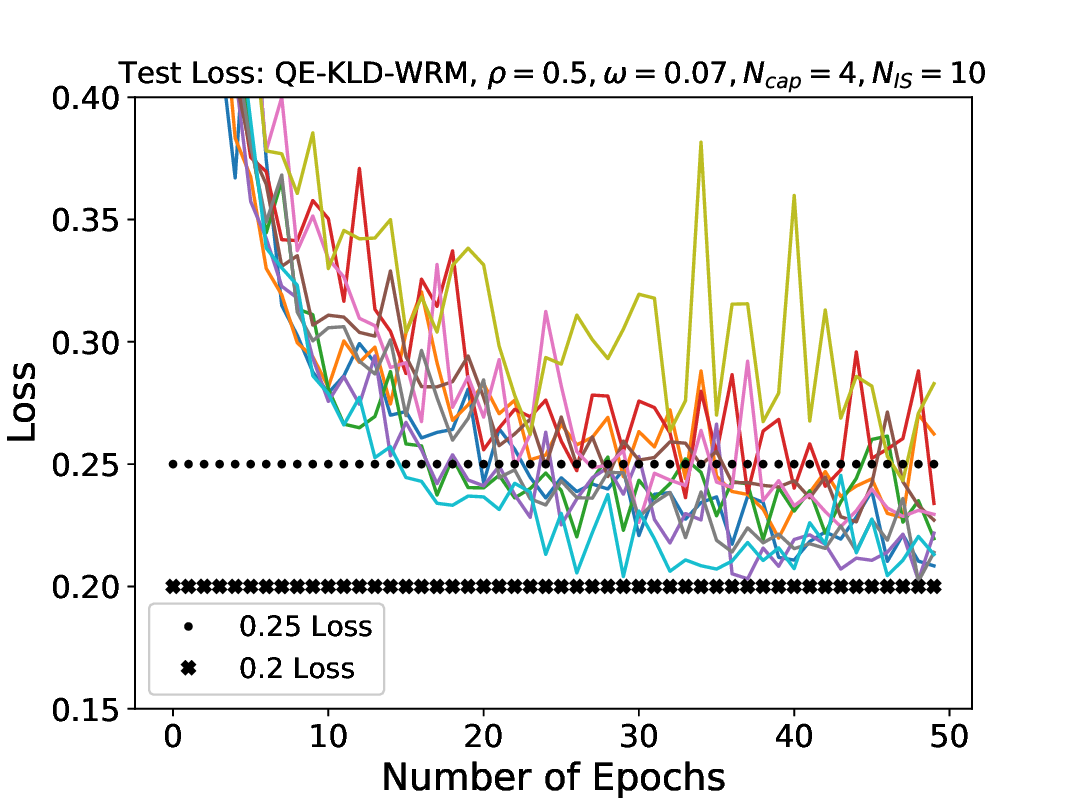}
	\caption{MNIST test-loss results for \textsc{k-fac}, and our three \textsc{kld-wrm} variants.} \label{MNIST_results_test_loss}
	\vspace{-2ex}
\end{figure}
\noindent That is because all our \textsc{kld-wrm} variants more robustly achieve good test metrics. Conversely, \textit{if we can run the training many times} (and choose the best run) \textsc{k-fac}'s large variance \textit{may} eventually play to our advantage (not guaranteed).

\textbf{\textsc{kld-wrm} variant selection:} \textsc{so-kld-wrm} and \textsc{q-kld-wrm} have virtually the same computation cost per epoch as \textsc{k-fac}. Conversely, \textsc{qe-kld-wrm} has the same \textit{data acquisition} and \textit{linear algebra} costs as \textsc{k-fac}, but 3-10 times higher oracle cost (fwd.\ and bwd.\ pass cost), owing to approximately solving (\ref{eqn_QE_KLD_WRM_instantiated}). Thus, when \textit{data cost is relatively low}, \textsc{so-kld-wrm} and \textsc{q-kld-wrm} will be preferable, as they will have the smallest wall-time per epoch (while having almost the same performance per epoch as \textsc{qe-kld-wrm}). Conversely, when \textit{data cost dominates}, all 4 solvers will have the same wall-time per epoch. In this case, \textsc{qe-kld-wrm} is preferable as it gives the best performance per epoch\footnote{Codes repo: https://github.com/ConstantinPuiu/Rethinking-EA-of-the-Fisher}.

\section{Conclusions and Future Work}
We established an equivalence between \textsc{ea-cm} algorithms (typically used in ML) and \textsc{w}o\textsc{qm} algorithms (which we defined in \textit{Section 3}). The equivalence revealed what \textsc{ea-cm} algorithms are doing from a model-based optimization point of view. Generalizing from \textsc{w}o\textsc{qm}, we defined a broader class of algorithms in \textit{Section 4}: \textsc{kld-wrm}. We then focused our attention on a different subclass of \textsc{kld-wrm}, \textsc{lm-kld-wrm}, and provided three practical instantiations of it. Numerical results on MNIST showed that performance-metrics distributions have better mean and lower variance for our \textsc{kld-wrm} algorithms, indicating they are preferable to \textsc{k-fac} in practice due to higher robustness.

\textbf{Future work:} (a) \textsc{kld-wrm} for VI BNNs and RL; (b) convergence theory; (c) investigate \textsc{q} and \textsc{qe} variants when $B_k$ and $F_k$ have different structures; (d) consider \textsc{kld-wrm} algorithms outside the \textsc{lm-kld-wrm} subfamily (include info.\ at $\{\theta_j\}_{j<k}$ in $M(s;\mathcal F_k)$; see (\ref{eqn_most_general_KLDWRM})); (e) consider arbitrary $\xi\in \mathbb R$ (see footnote 12). 
\newpage
\begin{figure}[t]
	\centering
	\includegraphics[trim={0.0cm 0.1cm 0.0cm 0.95cm},clip,width=0.385\textwidth]{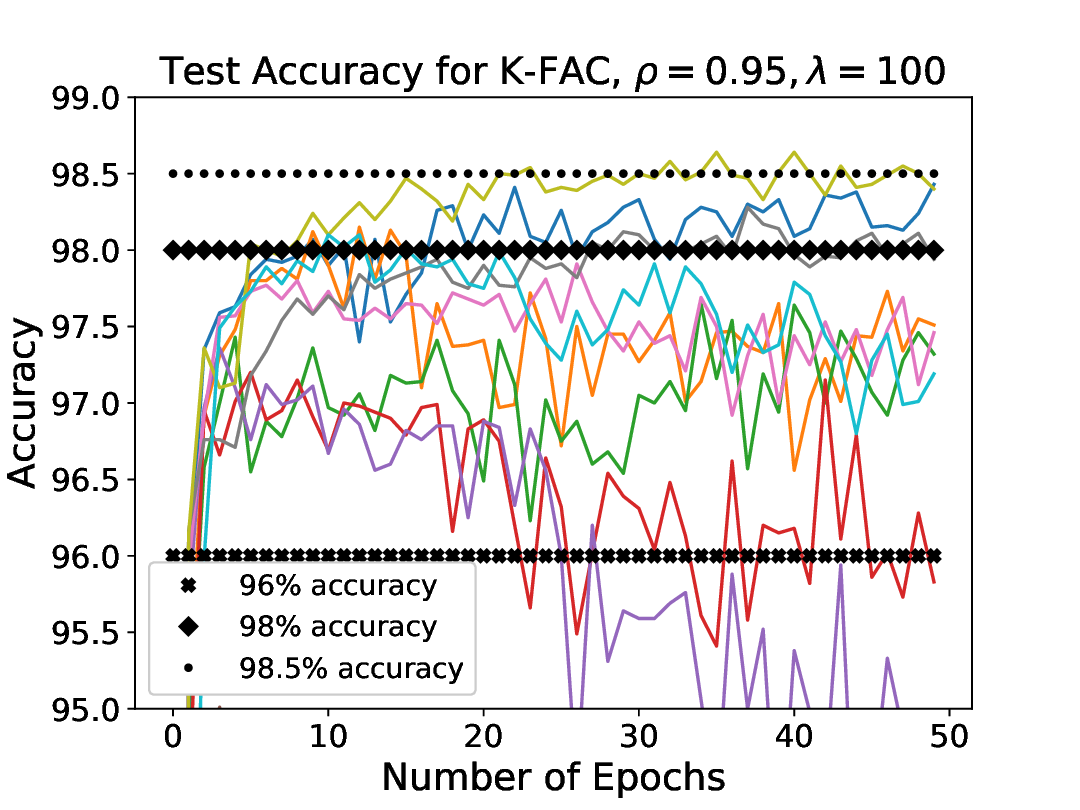}
	\includegraphics[trim={0.0cm 0.1cm 0.0cm 0.95cm},clip,width=0.385\textwidth]{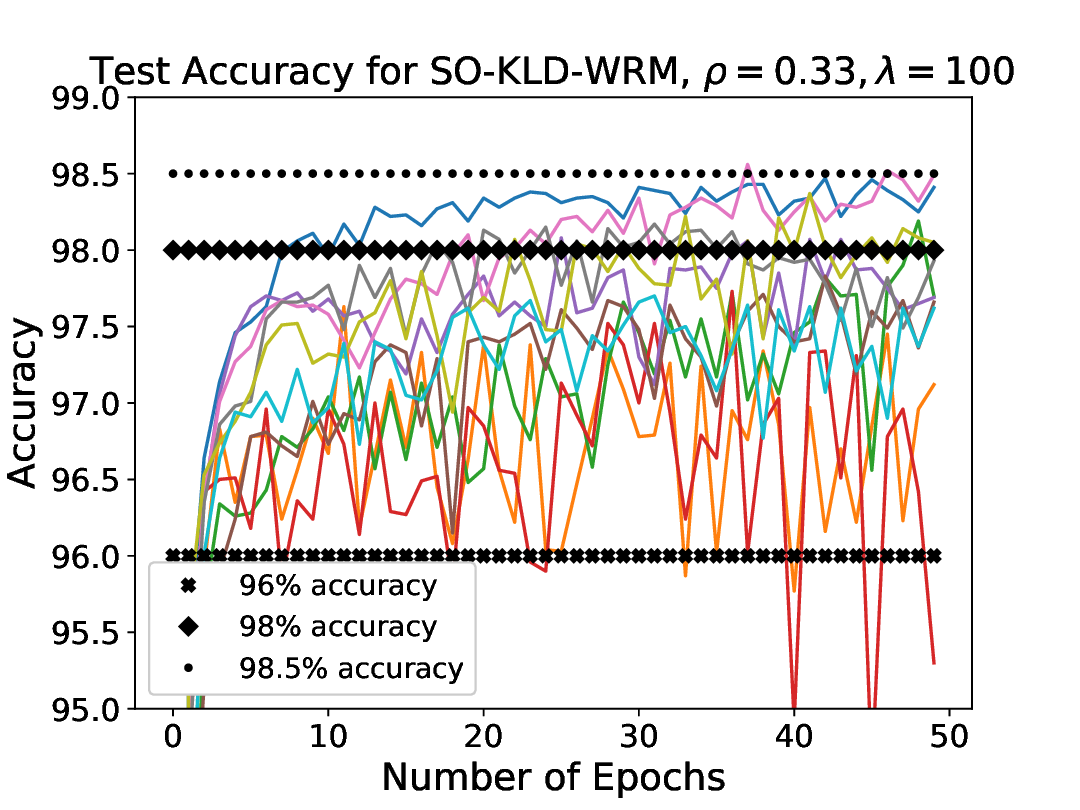}
	
	\includegraphics[trim={0.0cm 0.25cm 0.0cm 0.9cm},clip,width=0.385\textwidth]{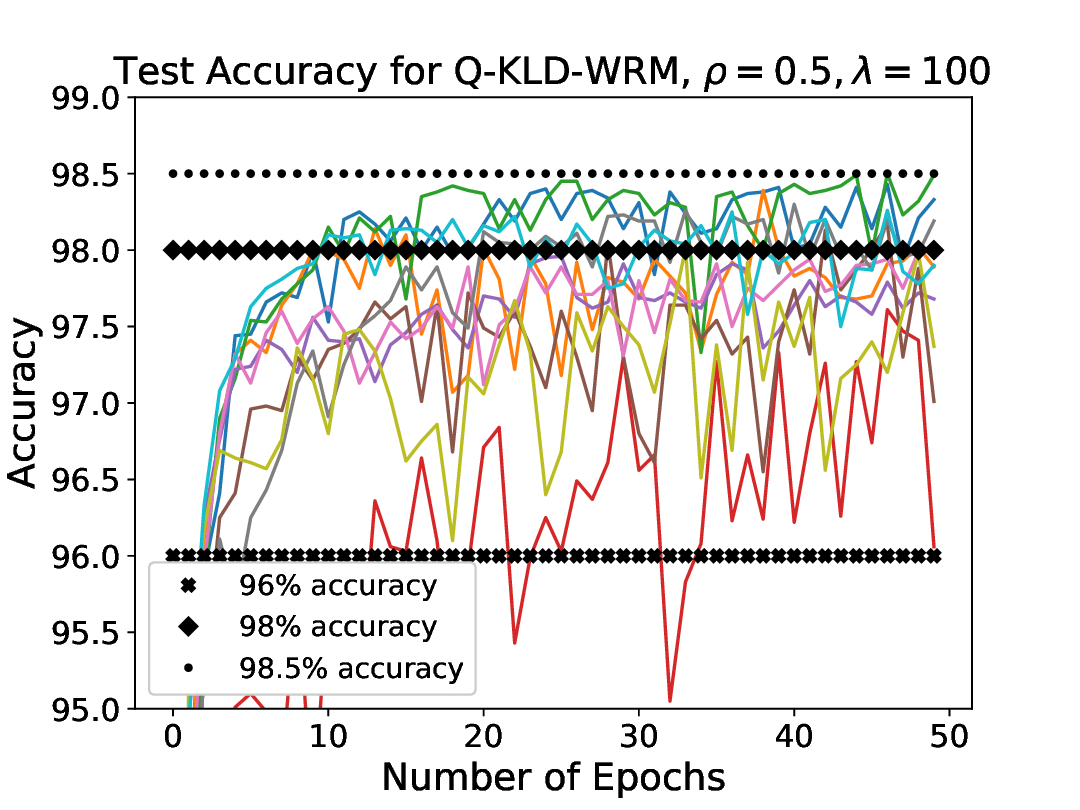}
	\includegraphics[trim={0.0cm 0.25cm 0.0cm 0.9cm},clip,width=0.385\textwidth]{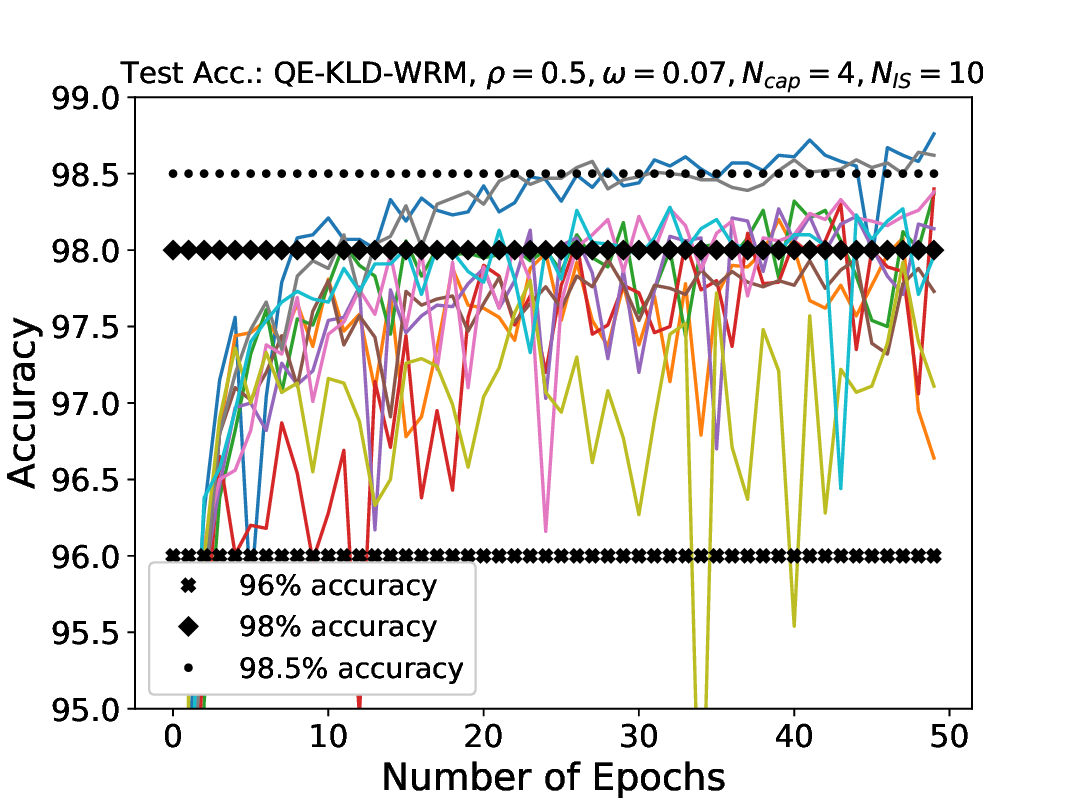}
	\caption{MNIST test-accuracy results for \textsc{k-fac}, and our three \textsc{kld-wrm} variants.} \label{MNIST_results_test_acc}
\end{figure}

\subsubsection{Acknowledgments}

Thanks to \textit{Jaroslav Fowkes} for very useful discussions. I am funded by the EPSRC CDT in InFoMM (EP/L015803/1) in collaboration with Numerical Algorithms Group and St.\ Anne's College (Oxford).

\end{document}


\newtheorem*{Proposition_3_1}{Proposition 3.1: Analytic Solution of \textit{WoQM} step}
\newtheorem*{Proposition_4_1}{Proposition 4.1: Analytic Solution of \textit{SO-KLD-WRM} step}
\newtheorem*{Proposition_4_2}{Proposition 4.2: Analytic Solution of \textit{Q-KLD-WRM} step}
\newtheorem*{Proposition_4_2_extension}{Proposition A.1 (Extension of Proposition 4.2): Analytic Solution of \textit{Q-KLD-WRM} step with variable $\lambda^{(k)}$}
\newcommand{\norm}[1]{\left\lVert#1\right\rVert}

\twocolumn[
\icmltitle{Supplementary Material: Rethinking Exponential Averaging of the Fisher}

\icmlsetsymbol{equal}{*}

\begin{icmlauthorlist}
\icmlauthor{Constantin Octavian Puiu}{}
\end{icmlauthorlist}

\icmlcorrespondingauthor{Constantin Octavian Puiu}{constantin.puiu@maths.ox.ac.uk}

\icmlkeywords{Natural Gradient, KL divergence regularization, Fisher momentum, statistics momentum, Fisher exponential average.}

\vskip 0.3in
]

\section{Proof of Proposition 3.1}
We reiterate the statement of \textit{Proposition 3.1} and the \textit{WoQM} definition for convenience. Recall that the \textit{WoQM} step at $\theta_k$ solves
\begin{equation}
\begin{split}
\min_s s^T\biggl[\sum_{i=0}^k\rho^{k-i}\biggl(g_i + \lambda \kappa(i) B_i\sum_{j=i}^{k-1}s_j\biggr)\biggr]\\
+\frac{\lambda}{2}s^T\biggl[\rho^kB_0 + (1-\rho)\sum_{i=1}^k\rho^{k-i}B_i\biggr]s,
\end{split}
\label{eqn_WoQM_for_particualr_model_appendix}
\end{equation}
\begin{Proposition_3_1}
	The \textit{WoQM} step $s_k$ at iterate $\theta_k$ can be expressed as \begin{equation}
	s_k = -\frac{1}{\lambda}\biggl(\rho^kB_0 + (1-\rho)\sum_{j=1}^k\rho^{k-j}B_j\biggr)^{-1}g_k,\,\,\,\,\,\forall k \in\mathbb N.
	\label{eqn_proposition_3_1_eqn}
	\end{equation}
\end{Proposition_3_1} 
\textit{Proof.} We use induction. 

\textit{Induction check. }First, check (\ref{eqn_proposition_3_1_eqn}) holds for $k=0$, $k=1$ and $k=2$. Since we have new sum terms starting to appear\footnote{Because $\kappa(i)=1$ for $i=0$, and $\kappa(i)=1-\rho$ for $i\geq1$ but the terms $(1-\rho)B_i\sum_{j=i}^{k-1}s_j$ only start appearing after $k\geq 2$.} at $k=1$ and also at $k=2$ (and keep on being present thereafter), we also need to check for $k=1$. 

For $k=0$ we have the model in (\ref{eqn_WoQM_for_particualr_model_appendix}) reducing to
\begin{equation}
\min_s s^Tg_0 +\frac{\lambda}{2}s^TB_0s,
\end{equation}
which has solution 
\begin{equation}
s_0 = -\frac{1}{\lambda}B_0^{-1}g_0,
\label{prop_3_1_proof_s_0}
\end{equation}
which satisfies (\ref{eqn_proposition_3_1_eqn}).

For $k=1$ we have the model in (\ref{eqn_WoQM_for_particualr_model_appendix}) reducing to
\begin{equation}
\min_s s^T[\rho g_0 + g_1 +\rho \lambda B_0 s_0] + \frac{\lambda}{2}s^T\big[\rho B_0 + (1-\rho) B_1\big]s.
\end{equation}
But, $\rho g_0 +\rho \lambda B_0 s_0 = 0$ by (\ref{prop_3_1_proof_s_0}). Thus, we have
\begin{equation}
s_1 = -\frac{1}{\lambda}\big[\rho B_0 + (1-\rho) B_1\big]^{-1}g_1,
\label{prop_3_1_proof_s_1}
\end{equation}
which satisfies (\ref{eqn_proposition_3_1_eqn}).

For $k=2$, we have the model in (\ref{eqn_WoQM_for_particualr_model_appendix}) reducing to
\begin{equation}
\begin{split}
&\min_s \frac{\lambda}{2}s^T\big[\rho^2 B_0 + (1-\rho)\rho B_1+ (1-\rho)B_2 \big]s+\\
s^T\biggl[&\rho^2 g_0 + \rho g_1 + g_2 + \rho^2\lambda B_0(s_0 + s_1) + \rho\lambda (1-\rho)B_1s_1\biggr].
\end{split}
\end{equation}
Now, $\rho^2g_0 + \rho^2\lambda B_0s_0 = 0$ using (\ref{prop_3_1_proof_s_0}). We also have $\rho g_1 + \rho\lambda(\rho B_0 + (1-\rho)B_1)s_1 = 0$ by (\ref{prop_3_1_proof_s_1}). Thus, we get
\begin{equation}
s_2 = -\frac{1}{\lambda}\big[\rho^2 B_0 + (1-\rho)\rho B_1+ (1-\rho)B_2 \big]^{-1}g_2,
\label{prop_3_1_proof_s_2}
\end{equation}
which satisfies (\ref{eqn_proposition_3_1_eqn}).

\textit{Induction main body.} Assume for the \textbf{inductive hypothesis} that we have for some $k\geq 2$
\begin{equation}
s_l = -\frac{1}{\lambda}\biggl(\rho^lB_0 + (1-\rho)\sum_{j=1}^l\rho^{l-j}B_j\biggr)^{-1}g_l,\,\,\,\,\,\forall l\leq k \in\mathbb N.
\label{inductive_hypothesis_proposition_3_1}
\end{equation}
We \textbf{need to prove} that at $k+1$, for all $k\geq 2$, we have 
\begin{equation}
s_{k+1} = -\frac{1}{\lambda}\biggl(\rho^{k+1}B_0 + (1-\rho)\sum_{j=1}^{k+1}\rho^{k+1-j}B_j\biggr)^{-1}g_{k+1}
\label{induction_target_prop_3_1},
\end{equation}
and since the inductive check holds at $k\in \{0,1,2\}$, this would imply (\ref{eqn_proposition_3_1_eqn}).

To begin, consider (\ref{eqn_WoQM_for_particualr_model_appendix}) evaluated at $k+1$. We have that the \textit{WoQM} step at $k+1$, $s_{k+1}$ is given by
\begin{equation}
\begin{split}
\min_s\,\,\, \frac{\lambda}{2}s^T\biggl[\rho^{k+1}B_0 + (1-\rho)\sum_{i=1}^{k+1}\rho^{k+1-i}B_i\biggr]s\\
+s^T\biggl[\sum_{i=0}^{k+1}\rho^{k+1-i}\biggl(g_i + \lambda \kappa(i) B_i\sum_{j=i}^{k}s_j\biggr)\biggr].
\end{split}
\label{eqn_WoQM_for_particualr_model_appendix_k_plus_1}
\end{equation}
Let us now consider the coefficient of the linear term in (\ref{eqn_WoQM_for_particualr_model_appendix_k_plus_1}). It reads
\begin{equation}
\begin{split}
\sum_{i=0}^{k+1}\rho^{k+1-i}\biggl(g_i + \lambda \kappa(i) B_i\sum_{j=i}^{k}s_j\biggr) =
\\ g_{k+1} + \rho\biggl[\sum_{i=0}^{k}\rho^{k-i}\biggl(g_i + \lambda \kappa(i) B_i\sum_{j=i}^{k}s_j\biggr)\biggl] = \\
g_{k+1} + \rho \biggl[\sum_{i=0}^{k}\rho^{k-i} g_i + \lambda \sum_{i=0}^{k}\kappa(i)\rho^{k-i}\biggl(B_i\sum_{j=i}^{k}s_j\biggr)\biggr].
\label{eqn_linear_term_prop_3_1_appendix_proof}
\end{split}
\end{equation}
Now we prove that the bracket on the third line of (\ref{eqn_linear_term_prop_3_1_appendix_proof}) is zero. Once we have that, we can very easily see from (\ref{eqn_WoQM_for_particualr_model_appendix_k_plus_1}) that we have the desired result. To show the term is zero, consider
\begin{equation}
\begin{split}
\lambda \sum_{i=0}^{k}\kappa(i)\rho^{k-i}\biggl(B_i\sum_{j=i}^{k}s_j\biggr) = \lambda \sum_{i=0}^{k}\sum_{j=i}^{k}\kappa(i)\rho^{k-i}B_i s_j 
\end{split}
\end{equation}
Now, inverting the $i$ and $j$ sums, we get
\begin{equation}
\begin{split}
\lambda \sum_{i=0}^{k}\kappa(i)\rho^{k-i}\biggl(B_i\sum_{j=i}^{k}s_j\biggr) = \lambda \sum_{j=0}^{k}\sum_{i=0}^{j}\kappa(i)\rho^{k-i}B_i s_j = \\
\lambda \sum_{j=0}^{k}\rho^{k-j}\biggl(\sum_{i=0}^{j}\rho^{j-i}\kappa(i)B_i\biggr) s_j.
\end{split}
\label{eqn_for_inductive_hypothesis_plugin_proposition_3_1}
\end{equation}
Now, explicitly writing that $\kappa(0)=1$ and $\kappa(i)=1-\rho$ for all $i\geq 1$, we get that 
\begin{equation}
\biggl(\sum_{i=0}^{j}\rho^{j-i}\kappa(i)B_i\biggr) s_j = \biggl(\rho^jB_0+(1-\rho)\sum_{i=1}^{j}\rho^{j-i}B_i\biggr) s_j.
\label{eqn_porp_3_1_big_bracket_extracted_term_proof}
\end{equation}
Substituting the inductive hypothesis (\ref{inductive_hypothesis_proposition_3_1}) into (\ref{eqn_porp_3_1_big_bracket_extracted_term_proof}), we get
\begin{equation}
\biggl(\sum_{i=0}^{j}\rho^{j-i}\kappa(i)B_i\biggr) s_j = -\frac{1}{\lambda}g_j
\label{final_formula_for_lengthy_inner_term_prop_3_1}
\end{equation}
Plugging (\ref{final_formula_for_lengthy_inner_term_prop_3_1}) into (\ref{eqn_for_inductive_hypothesis_plugin_proposition_3_1}), we see that
\begin{equation}
\lambda \sum_{i=0}^{k}\kappa(i)\rho^{k-i}\biggl(B_i\sum_{j=i}^{k}s_j\biggr) = -\sum_{j=0}^k \rho^{k-j}g_j.
\label{eqn_prop_3_1_B_sum_linear_term_expressioN_solved}
\end{equation}
Finally, plugging (\ref{eqn_prop_3_1_B_sum_linear_term_expressioN_solved}) into the last line of (\ref{eqn_linear_term_prop_3_1_appendix_proof}), we see that 
\begin{equation}
\begin{split}
\sum_{i=0}^{k+1}\rho^{k+1-i}\biggl(g_i + \lambda B_i\sum_{j=i}^{k}s_j\biggr) = g_{k+1},
\label{to_be_plugged_into_final_result_prop_3_1}
\end{split}
\end{equation}
which means the linear coefficient in (\ref{eqn_WoQM_for_particualr_model_appendix_k_plus_1}) is just $g_{k+1}$. Thus, differentiating (\ref{eqn_WoQM_for_particualr_model_appendix_k_plus_1}) and setting to zero gives us the global solution\footnote{Since $B_k$ are positive definite by assumption.} of the \textit{WoQM} model as

\begin{equation}
s_{k+1} = -\frac{1}{\lambda}\biggl[\rho^{k+1}B_0 + (1-\rho)\sum_{i=1}^{k+1}\rho^{k+1-i}B_i\biggr]^{-1}g_{k+1}
\label{induction_completed_prop_3_1}.
\end{equation}
Equation (\ref{induction_completed_prop_3_1}) is the same as (\ref{induction_target_prop_3_1}), which completes the proof $\square$.

\section{Proof of Proposition 4.1}
We reiterate the statement of \textit{Proposition 4.1} and the \textit{SO-KLD-WRM} definition for convenience. Recall that the \textit{SO-KLD-WRM} step at $\theta_k$ solves
\begin{equation}
 \min_s g_k^Ts + \lambda\sum_{i=0}^k\kappa(i)\rho^{k-i}\tilde{\mathbb D}_{KL}(\theta_i||\theta_{k} + s),
 \label{eqn_SO_KLD_WRM}
\end{equation}

That is, the \textit{SO-KLD-WRM} step at $\theta_k$ solves
\begin{equation}
\begin{split}
\min_s s^T\biggl[g_k + \sum_{i=0}^{k-1}\rho^{k-i}\biggl(\lambda \kappa(i) F_i\sum_{j=i}^{k-1}s_j\biggr)\biggr]\\
+\frac{\lambda}{2}s^T\biggl[\sum_{i=0}^k\kappa(i)\rho^{k-i}F_i\biggr]s.
\end{split}
\label{eqn_SO_KLD_WRM_2}
\end{equation}

\begin{Proposition_4_1}
 	The \textit{SO-KLD-WRM} step $s_k$ at iterate $\theta_k$ can be expressed as 
 	\begin{equation}
 	s_k = -\frac{1}{\lambda}\biggl(\rho^kF_0 + (1-\rho)\sum_{j=1}^k\rho^{k-j}F_j\biggr)^{-1}[g_k-\rho g_{k-1}],
 	\label{eqn_proposition_4_1_eqn}
 	\end{equation}
 	$	\forall k \in\mathbb N$, where we set $g_{-1}:=0$ by convention.
\end{Proposition_4_1} 
\textit{Proof.} We use induction. The proof is very similar to the prof of \textit{Proposition 3.1}.

\textit{Induction check. }First, check (\ref{eqn_proposition_4_1_eqn}) holds for $k=0$, $k=1$ and $k=2$. Since we have new sum terms starting to appear\footnote{Because $\kappa(i)=1$ for $i=0$, and $\kappa(i)=1-\rho$ for $i\geq1$ but the terms $(1-\rho)B_i\sum_{j=i}^{k-1}s_j$ only start appearing after $k\geq 2$.} at $k=1$ and also at $k=2$ (and keep on being present thereafter), we also need to check for $k=1$. 

For $k=0$ we have the model in (\ref{eqn_SO_KLD_WRM_2}) reducing to
\begin{equation}
\min_s s^Tg_0 +\frac{\lambda}{2}s^TF_0s,
\end{equation}
which has solution 
\begin{equation}
s_0 = -\frac{1}{\lambda}F_0^{-1}g_0,
\label{prop_4_1_proof_s_0}
\end{equation}
which satisfies (\ref{eqn_proposition_4_1_eqn}).

For $k=1$ we have the model in (\ref{eqn_WoQM_for_particualr_model_appendix}) reducing to
\begin{equation}
\min_s s^T[g_1 +\rho \lambda F_0 s_0] + \frac{\lambda}{2}s^T\big[\rho F_0 + (1-\rho) F_1\big]s.
\end{equation}
But, $g_1 +\rho \lambda B_0 s_0 = g_1 - \rho g_0$ by (\ref{prop_4_1_proof_s_0}). Thus, we have
\begin{equation}
s_1 = -\frac{1}{\lambda}\big[\rho F_0 + (1-\rho) F_1\big]^{-1}[g_1 - \rho g_0],
\label{prop_4_1_proof_s_1}
\end{equation}
which satisfies (\ref{eqn_proposition_4_1_eqn}).

For $k=2$, we have the model in (\ref{eqn_WoQM_for_particualr_model_appendix}) reducing to
\begin{equation}
\begin{split}
&\min_s \frac{\lambda}{2}s^T\big[\rho^2 F_0 + (1-\rho)\rho F_1+ (1-\rho)F_2 \big]s+\\
&s^T\biggl[g_2 + \rho^2\lambda F_0(s_0 + s_1) + \rho\lambda (1-\rho)F_1s_1\biggr].
\end{split}
\label{eqn_s_2_prop_4_1}
\end{equation}
Now, $\rho^2\lambda F_0s_0 = -\rho^2 g_0$ using (\ref{prop_4_1_proof_s_0}). We also have $\rho\lambda[\rho F_0+(1-\rho)F_1]s_1 = -\rho[g_1-\rho g_0]$ by (\ref{eqn_proposition_4_1_eqn}). Thus, we get that the linear term of (\ref{eqn_s_2_prop_4_1}) is $s^T[g_2-\rho g_1]$. That is, 
\begin{equation}
s_2 = -\frac{1}{\lambda}\big[\rho^2 F_0 + (1-\rho)\rho F_1+ (1-\rho)F_2 \big]^{-1}[g_2-\rho g_1],
\label{prop_4_1_proof_s_2}
\end{equation}
which satisfies (\ref{eqn_proposition_4_1_eqn}).

\textit{Induction main body.} Assume for the \textbf{inductive hypothesis} that we have for some $k\geq 2$
\begin{equation}
s_l = -\frac{1}{\lambda}\biggl(\sum_{j=0}^l\kappa(j)\rho^{l-j}F_j\biggr)^{-1}[g_l-\rho g_{l-1}],\,\,\,\,\,\forall l\leq k.
\label{inductive_hypothesis_proposition_4_1}
\end{equation}
We \textbf{need to prove} that at $k+1$, for all $k\geq 2$, we have 
\begin{equation}
s_{k+1} =  -\frac{1}{\lambda}\biggl(\sum_{j=0}^{k+1}\kappa(j)\rho^{k+1-j}F_j\biggr)^{-1}[g_{k+1}-\rho g_{k}]
\label{induction_target_prop_4_1},
\end{equation}
and since the inductive check holds at $k\in \{0,1,2\}$, this would imply (\ref{eqn_proposition_4_1_eqn}).

To begin, consider (\ref{eqn_SO_KLD_WRM_2}) evaluated at $k+1$. We have that the \textit{SO-KLD-WRM} step at $k+1$, $s_{k+1}$ is given by
\begin{equation}
\begin{split}
\min_s\,\,\, \frac{\lambda}{2}s^T\biggl[\rho^{k+1}F_0 + (1-\rho)\sum_{i=1}^{k+1}\rho^{k+1-i}F_i\biggr]s\\
+s^T\biggl[g_{k+1}+ \sum_{i=0}^{k}\rho^{k+1-i}\biggl(\lambda \kappa(i) F_i\sum_{j=i}^{k}s_j\biggr)\biggr].
\end{split}
\label{eqn_SOKLDWRM_for_particualr_model_appendix_k_plus_1}
\end{equation}
Let us now consider the coefficient of the linear term in (\ref{eqn_SOKLDWRM_for_particualr_model_appendix_k_plus_1}). It reads
\begin{equation}
\begin{split}
g_{k+1}+ \sum_{i=0}^{k}\rho^{k+1-i}\biggl(\lambda \kappa(i) F_i\sum_{j=i}^{k}s_j\biggr) =\\
g_{k+1}+ \lambda\sum_{i=0}^{k}\sum_{j=i}^{k}\rho^{k+1-i} \kappa(i) F_i s_j = \\
g_{k+1}+ \lambda\sum_{j=0}^{k}\biggl[\biggl(\sum_{i=0}^{j}\rho^{k+1-i} \kappa(i) F_i\biggr)s_j\biggr].
\label{eqn_linear_term_prop_4_1_appendix_proof}
\end{split}
\end{equation}
In the last line, we interchanged the summation over $i$ and $j$. But the square bracket in the second term of (\ref{eqn_linear_term_prop_4_1_appendix_proof}) is given by
\begin{equation}
\begin{split}
\biggl(\sum_{i=0}^{j}\rho^{k+1-i} \kappa(i) F_i\biggr) s_j = \rho^{k+1 - j} \biggl(\sum_{i=0}^{j}\rho^{j-i} \kappa(i) F_i\biggr) s_j
\label{eqn_prop_4_1_to_plug_in_inductive_hypothesis}
\end{split}
\end{equation}
Now, plugging the inductive hypothesis (\ref{inductive_hypothesis_proposition_4_1}) in (\ref{eqn_prop_4_1_to_plug_in_inductive_hypothesis}) - which we can do as the outer-sum over $j$ in (\ref{eqn_linear_term_prop_4_1_appendix_proof}) is up until $k$ - we get
\begin{equation}
\begin{split}
\biggl(\sum_{i=0}^{j}\rho^{k+1-i} \kappa(i) F_i\biggr) s_j = -\frac{1}{\lambda}\rho^{k+1 - j}[g_j-\rho g_{j-1}]
\end{split}
\label{eqn_prop_4_1_telescopic_sum_element_formula}
\end{equation}
Finally, plugging (\ref{eqn_prop_4_1_telescopic_sum_element_formula}) into (\ref{eqn_linear_term_prop_4_1_appendix_proof}), we get
\begin{equation}
\begin{split}
&g_{k+1}+ \sum_{i=0}^{k}\rho^{k+1-i}\biggl(\lambda \kappa(i) F_i\sum_{j=i}^{k}s_j\biggr) =\\
g_{k+1} + \lambda &\sum_{j=0}^k\biggl[-\frac{1}{\lambda}\rho^{k+1 - j}[g_j-\rho g_{j-1}]\biggr] = g_{k+1} - \rho g_{k}.
\label{eqn_final_linear_term_prop_4_1_appendix_proof}
\end{split}
\end{equation}
To get the last equality in (\ref{eqn_final_linear_term_prop_4_1_appendix_proof}) we solved the telescopic sum, and in doing that we recalled our convention $g_{-1} = 0$.

Finally, plugging (\ref{eqn_final_linear_term_prop_4_1_appendix_proof}) into (\ref{eqn_SOKLDWRM_for_particualr_model_appendix_k_plus_1}), we see that 
\begin{equation}
s_{k+1} = -\frac{1}{\lambda}\biggl[\sum_{i=0}^{k+1}\kappa(i)\rho^{k+1-i}F_i\biggr]^{-1}[g_{k+1}-\rho g_k]
\label{induction_completed_prop_4_1}.
\end{equation}
Equation (\ref{induction_completed_prop_4_1}) is the same as (\ref{induction_target_prop_4_1}), which completes the proof $\square$.

\section{Proof of Proposition 4.2}
We reiterate the \textit{Q-KLD-WRM} step definition as well as \textit{Proposition 4.2} here for convenience. The \textit{Q-KLD-WRM} step at $\theta_k$ is the solution to
\begin{equation}
\min_s g_k^Ts + \frac{1}{2}s^TB_ks +  \lambda\sum_{i=0}^k\kappa(i)\rho^{k-i}\tilde{\mathbb D}_{KL}(\theta_i||\theta_{k} + s),
\label{eqn_Q_KLD_WRM}
\end{equation}
That is, the \textit{Q-KLD-WRM} step at $\theta_k$ solves
\begin{equation}
\begin{split}
\min_s s^T\biggl[g_k + \sum_{i=0}^{k-1}\rho^{k-i}\biggl(\lambda \kappa(i) F_i\sum_{j=i}^{k-1}s_j\biggr)\biggr]\\
+\frac{\lambda}{2}s^T\biggl[\sum_{i=0}^k\kappa(i)\rho^{k-i}F_i + \frac{1}{\lambda}B_k\biggr]s.
\end{split}
\label{eqn_Q_KLD_WRM_2}
\end{equation}
Of course, we have $\bar F_k = \sum_{i=0}^k\kappa(i)\rho^{k-i}F_i$, and we will use that notation most of the times.

\textit{Proposition 4.2} is given again below.
\begin{Proposition_4_2}
	The \textit{Q-KLD-WRM} step $s_k$ at location $\theta_k$ is given by the solution to the problem
	\begin{equation}
	\min_s s^T\hat g_k + \frac{\lambda}{2}s^T\biggl[\bar F_k + \frac{1}{\lambda}B_k\biggr]s
	\label{eqn_prop_4_2_result_0}
	\end{equation}
	
	where $\hat g_k$ is given by the one-step recursion
	\begin{equation}
	\hat g_{k+1} = g_{k+1} + \rho(I-\hat M_k)\hat g_{k}- \rho g_k,\,\,\,\,\,\forall k\in \mathbb Z^+,
	\label{eqn_prop_4_2_result_2}
	\end{equation}
	with $\hat g_0 := g_0$, $\bar F_k := \sum_{i=0}^k\kappa(i)\rho^{k-i}F_i$, and 
	\begin{equation}
	\hat M_k :=\big[I +\frac{1}{\lambda}B_k \bar F_k^{-1}\big]^{-1}.
	\label{eqn_prop_4_2_result_3}
	\end{equation}
	That is, the \textit{Q-KLD-WRM} step is formally given by
	\begin{equation}
	s_{k}= -\frac{1}{\lambda}\biggl[\bar F_k + \frac{1}{\lambda}B_k\biggr]^{-1}\hat g_k.
	\label{eqn_prop_4_2_result_1}
	\end{equation}
\end{Proposition_4_2}

\textit{Proof.} By induction. The induction is mainly focused around finding a form for the linear term of the \textit{Q-KLD-WRM} formulation at arbitrary $k$. The step analytic formula then follows easily.

\textit{Induction check.}
For $k=0$, we have the model in (\ref{eqn_Q_KLD_WRM_2}) reducing to
\begin{equation}
\min_s s^Tg_0 + \frac{\lambda}{2}s^T\biggl[\bar F_0 +\frac{1}{\lambda}B_0\biggr]s
\end{equation}
Now, noting that $\hat g_0 = g_0$ by definition, we see that the \textit{Q-KLD-WRM} linear\footnote{The Quadratic term obviously also obeys \textit{Propositon 4.2 formulation} - but this fact follows trivially from the original \textit{Q-KLD-WRM} definition, as we can see.} term obeys \textit{Proposition 4.2} at $k=0$. Thus, 
\begin{equation}
s_0 = -\frac{1}{\lambda}\biggl[\bar F_0 +\frac{1}{\lambda}B_0\biggr]^{-1}\hat g_0.
\label{eqn_prop_4_2_s_0_ind_check}
\end{equation}

For $k=1$, we have the model in (\ref{eqn_Q_KLD_WRM_2}) reducing to
\begin{equation}
\min_s s^T[g_1 +\rho \lambda F_0 s_0] + \frac{\lambda}{2}s^T\biggl[\bar F_1+\frac{1}{\lambda}B_1\biggr]s.
\end{equation}
Now, by using (\ref{eqn_prop_4_2_s_0_ind_check}), we have that

\begin{equation}
	\rho \lambda F_0 s_0 = -\rho \big[F_0^{-1}\big]^{-1} \biggl[\bar F_0 +\frac{1}{\lambda}B_0\biggr]^{-1}\hat g_0 
\end{equation}
\begin{equation}
\rho \lambda F_0 s_0 = -\rho \biggl[\bar F_0 F_0^{-1} +\frac{1}{\lambda}B_0 F_0^{-1}\biggr]^{-1}\hat g_0 = -\rho \hat M_0\hat g_0
\end{equation}
where the latter equality holds by the definition of $\hat M_k$ evaluated at $k=0$ and the fact that $\bar F_0 = F_0$. Thus,
\begin{equation}
g_1 +\rho \lambda F_0 s_0 = g_1 - \rho\hat M_0 \hat g_0 = g_1  + \rho(I-\hat M_0)\hat g_0 -\rho g_0.\label{introducing_sneaky_terms_prop_4_2_ind_check_s_1}
\end{equation}
The latter equality in (\ref{introducing_sneaky_terms_prop_4_2_ind_check_s_1}) holds because $g_0=\hat g_0$. Consequently, using the recursive definition of $\hat g_{k+1}$ for $k=0$, we get
\begin{equation}
g_1 +\rho \lambda F_0 s_0 = \hat g_1\label{hat_g_1_prop_4_2_ind_check_s_1},
\end{equation}
which shows us that the \textit{Q-KLD-WRM} linear term obeys the formulation in \textit{Proposition 4.2}. Thus the whole \textit{Q-KLD-WRM} formulation obeys formulation (\ref{eqn_prop_4_2_result_0}). In particular, we have
\begin{equation}
s_1 = -\frac{1}{\lambda}\biggl[\bar F_1+\frac{1}{\lambda}B_1\biggr]^{-1}\hat g_1.
\label{eqn_prop_4_2_s_1ind_check}
\end{equation}

For $k=2$, we have the model in (\ref{eqn_Q_KLD_WRM_2}) reducing to
\begin{equation}
\begin{split}
&\min_s \frac{\lambda}{2}s^T\biggl[\bar F_2 + \frac{1}{\lambda}B_2\biggr]s+\\
s^T\biggl[g_2 + &\rho^2\lambda F_0(s_0 + s_1) + \rho\lambda (1-\rho)F_1s_1\biggr].
\end{split}
\label{eqn_s_2_prop_4_2}
\end{equation}
The linear term in (\ref{eqn_s_2_prop_4_2}) can be rearranged as
\begin{equation}
\begin{split}
s^T\biggl[g_2 + &\rho^2\lambda F_0 s_0  + \rho\lambda [\rho F_0 + (1-\rho)F_1]s_1\biggr] = \\
&s^T\biggl[g_2 + \rho^2 \lambda \bar F_0 s_0  + \rho\lambda \bar F_1s_1\biggr] = \\
s^T\biggl[g_2 +&\rho(g_1 + \rho \lambda \bar F_0 s_0)  + \rho\lambda \bar F_1s_1 - \rho g_1 \biggr] ,
\end{split}
\label{eqn_s_2_prop_4_2_part_2}
\end{equation}
where to get the second equality we used the definition of $\bar F_k$ applied at $k=0$ and $k=1$. We have already seen when doing the check for $k=1$ that
\begin{equation}
g_1+ \rho \lambda \bar F_0 s_0  = \hat g_1
\label{eqn_g_1_hat_2nd_time_prop_4_2}
\end{equation}
Thus, we only need to work out the $\rho\lambda \bar F_1s_1$ term. By using (\ref{eqn_prop_4_2_s_1ind_check}), this reads
\begin{equation}
\rho\lambda \bar F_1s_1 = - \rho \big[\bar F_1^{-1}\big]^{-1} \biggl[\bar F_1+\frac{1}{\lambda}B_1\biggr]^{-1}\hat g_1
\end{equation}
\begin{equation}
\rho \lambda \bar F_1 s_1 = -\rho \biggl[\bar F_1 \bar F_1^{-1} +\frac{1}{\lambda}B_1 \bar F_1^{-1}\biggr]^{-1}\hat g_1 = -\rho \hat M_1\hat g_1
\label{prop_4_2_ind_check_s_2_M_hat_term}
\end{equation}
where the latter equality holds by the definition of $M_k$ evaluated at $k=1$. Thus, plugging in (\ref{prop_4_2_ind_check_s_2_M_hat_term}) and (\ref{eqn_g_1_hat_2nd_time_prop_4_2})\footnote{Or indeed can use (\ref{hat_g_1_prop_4_2_ind_check_s_1}), we wrote it again for clarity.} in the linear term expression (\ref{eqn_s_2_prop_4_2_part_2}),
we get that the linear term reads
\begin{equation}
\begin{split}
s^T\biggl[g_2 + &\rho^2\lambda F_0 s_0  + \rho\lambda [\rho F_0 + (1-\rho)F_1]s_1\biggr] = \\ 
&s^T\biggl[g_2 + \rho(I-\hat M_1)\hat g_1 - \rho g_1\biggr] = s^T \hat g_2,
\end{split}
\label{eqn_s_2_linear_term_final_prop_4_2_part_2}
\end{equation}
where in the last equality we used the recursive definition of $\hat g_{k+1}$ from \textit{Proposition 4.2}, evaluated at $k=1$. Thus, (\ref{eqn_s_2_linear_term_final_prop_4_2_part_2}) shows us that the \textit{Q-KLD-WRM} linear term obeys the formulation in \textit{Proposition 4.2}. Thus the whole \textit{Q-KLD-WRM} formulation obeys formulation (\ref{eqn_prop_4_2_result_0}). In particular, if we now plug the linear term in (\ref{eqn_s_2_linear_term_final_prop_4_2_part_2}) back into the full model (\ref{eqn_s_2_prop_4_2}), we get that the \textit{Q-KLD-WRM} step satisfies
\begin{equation}
s_2 = -\frac{1}{\lambda} \biggl[\bar F_2 + \frac{1}{\lambda}B_2\biggr]^{-1}\hat g_2.
\label{eqn_prop_4_2_s_2ind_check}
\end{equation}
We are done with the induction check.

\textit{Induction main body.} 
Assume for the \textbf{inductive hypothesis} that we have for some $k\geq 2$
\begin{equation}
s_{l}= -\arg\min_s s^T\hat g_l + \frac{\lambda}{2}s^T\biggl[\bar F_l + \frac{1}{\lambda}B_l\biggr]s,\,\forall l\in\{0,..,k\}
\label{eqn_prop_4_2_result_1_inductive_hypothesis}
\end{equation}
\begin{equation}
\hat g_{l} = g_{l} + \rho(I-\hat M_{l-1})\hat g_{l-1}- \rho g_{l-1},\,\,\,\forall l\in\{1,...,k\},
\label{eqn_prop_4_2_result_2_inductive_hypothesis}
\end{equation}
with $\hat g_0 := g_0$, $\bar F_k := \sum_{i=0}^k\kappa(i)\rho^{k-i}F_i$, and 
\begin{equation}
\hat M_l :=\big[I +\frac{1}{\lambda}B_l \bar F_l^{-1}\big]^{-1}.
\label{eqn_prop_4_2_result_3_inductive_hypothesis}
\end{equation}
We \textbf{need to prove} that at $k+1$, for all $k\geq 2$, we have
\begin{equation}
 \hat g_{k+1} = g_{k+1} + \rho(I-\hat M_k)\hat g_{k}- \rho g_{k},
 \label{eqn_prop_4_2_result_2_inductive_target_1}
 \end{equation}
with $\hat M_k$ defined as in (\ref{eqn_prop_4_2_result_3_inductive_hypothesis}), and
\begin{equation}
s_{k+1}= \arg\min_s s^T\hat g_{k+1} + \frac{\lambda}{2}s^T\biggl[\bar F_{k+1} + \frac{1}{\lambda}B_{k+1}\biggr]s.
\label{eqn_prop_4_2_result_1_inductive_target_2}
\end{equation}
Since the inductive check holds at $k\in \{0,1,2\}$, this would imply \textit{Proposition 4.2} holds.

To begin, consider (\ref{eqn_Q_KLD_WRM_2}) evaluated at $k+1$. We have that the \textit{Q-KLD-WRM} step at $k+1$, $s_{k+1}$ is the solution to
\begin{equation}
\begin{split}
\min_s\,\,\, \frac{\lambda}{2}s^T\biggl[\bar F_{k+1} +\frac{1}{\lambda}B_{k+1}\biggr]s\\
+s^T\biggl[g_{k+1}+ \sum_{i=0}^{k}\rho^{k+1-i}\biggl(\lambda \kappa(i) F_i\sum_{j=i}^{k}s_j\biggr)\biggr].
\end{split}
\label{eqn_QKLDWRM_for_particualr_model_appendix_k_plus_1}
\end{equation}
Let us now consider the coefficient of the linear term in (\ref{eqn_QKLDWRM_for_particualr_model_appendix_k_plus_1}). It reads
\begin{equation}
\begin{split}
g_{k+1}+ \sum_{i=0}^{k}\rho^{k+1-i}\biggl(\lambda \kappa(i) F_i\sum_{j=i}^{k}s_j\biggr) =\\
g_{k+1}+ \lambda\sum_{i=0}^{k}\sum_{j=i}^{k}\rho^{k+1-i} \kappa(i) F_i s_j = \\
g_{k+1}+ \lambda\sum_{j=0}^{k}\biggl[\biggl(\sum_{i=0}^{j}\rho^{k+1-i} \kappa(i) F_i\biggr)s_j\biggr] =\\
 g_{k+1} + \lambda \sum_{j=0}^k\rho^{k+1-j}\bar F_j s_j =  g_{k+1} + \rho  \sum_{j=0}^k\rho^{k-j}\lambda \bar F_j s_j .
\label{eqn_linear_term_prop_4_2_appendix_proof}
\end{split}
\end{equation}
To get from the third to the fourth line in (\ref{eqn_linear_term_prop_4_2_appendix_proof}) we used the definition of $\bar F_k$ evaluated at $k=j$. We can further write the linear coeffcient of (\ref{eqn_QKLDWRM_for_particualr_model_appendix_k_plus_1}) as
\begin{equation}
\begin{split}
g_{k+1}+ \sum_{i=0}^{k}\rho^{k+1-i}\biggl(\lambda \kappa(i) F_i\sum_{j=i}^{k}s_j\biggr) =\\ 
 g_{k+1} + \rho  \sum_{j=0}^k\rho^{k-j}\lambda \bar F_j s_j = \\
 g_{k+1} + \rho  \sum_{j=0}^{k-1}\rho^{k-j}\lambda \bar F_j s_j +\rho\lambda \bar F_ks_k.
 \label{prop_4_2_over_all_eqn_to_plug_particular_terms_in}
\end{split}
\end{equation}
Now, considering the same linear term of our \textit{Q-KLD-WRM} formulation, \textit{but at arbitrary} $l\in[1,k]\cap \mathbb Z$ (rather than at $k+1$) we get can apply\footnote{We can do that because our calculations (\ref{eqn_QKLDWRM_for_particualr_model_appendix_k_plus_1}) - (\ref{prop_4_2_over_all_eqn_to_plug_particular_terms_in}) hold for arbitrary $k\in \mathbb Z^+$, and so does the \textit{Q-KLD-WRM} formulation (\ref{prop_4_2_over_all_eqn_to_plug_particular_terms_in}).} equation (\ref{prop_4_2_over_all_eqn_to_plug_particular_terms_in}) to get
 \begin{equation}
\begin{split}
g_{l}+ \sum_{i=0}^{l-1}\rho^{l-i}\biggl(\lambda \kappa(i) F_i\sum_{j=i}^{k}s_j\biggr) =  
g_{l} + \sum_{j=0}^{l-1}\rho^{l-j}\lambda \bar F_j s_j.
\end{split}
\end{equation}
But the inductive hypothesis, (\ref{eqn_prop_4_2_result_1_inductive_hypothesis}) and (\ref{eqn_prop_4_2_result_2_inductive_hypothesis}), tells us that this linear term must also be $\hat g_l$. That is, we have
\begin{equation}
	\hat g_l = g_{l} + \sum_{j=0}^{l-1}\rho^{l-j}\lambda \bar F_j s_j,
\end{equation}
and therefore 
\begin{equation}
\sum_{j=0}^{l-1}\rho^{l-j}\lambda \bar F_j s_j = \hat g_l - g_{l}.
\label{eqn_ghat_g_l_minus_g_l}
\end{equation}
In particular (\ref{eqn_ghat_g_l_minus_g_l}) must also hold for $l=k$, because the induction hypothesis holds up until $\hat g_k$. We have 
\begin{equation}
\sum_{j=0}^{k-1}\rho^{k-j}\lambda \bar F_j s_j = \hat g_k - g_{k},
\label{eqn_ghat_g_k_minus_g_k}
\end{equation}
Now, plugging (\ref{eqn_ghat_g_k_minus_g_k}) into (\ref{prop_4_2_over_all_eqn_to_plug_particular_terms_in}), we have
\begin{equation}
\begin{split}
g_{k+1}+ \sum_{i=0}^{k}\rho^{k+1-i}\biggl(\lambda \kappa(i) F_i\sum_{j=i}^{k}s_j\biggr) =\\ 
g_{k+1} + \rho(\hat g_k - g_{k}) +\rho\lambda \bar F_ks_k,
\label{prop_4_2_over_all_eqn_to_plug_particular_terms_in_more_final}
\end{split}
\end{equation}
so all is left is to compute the term $\rho\lambda \bar F_ks_k$. To do that, we use the induction hypothesis again. In particular, we use (\ref{eqn_prop_4_2_result_1_inductive_hypothesis}) for $l=k$ and solve exactly for $s_k$, to get
\begin{equation}
s_k = -\frac{1}{\lambda}\biggl[\bar F_k+\frac{1}{\lambda}B_k\biggr]^{-1}\hat g_k
\label{s_k_proof_prop_4_2}.
\end{equation}
Using (\ref{s_k_proof_prop_4_2}), we have
\begin{equation}
\rho\lambda \bar F_ks_k = -\rho \big[\bar F_k^{-1}\big]^{-1}\biggl[\bar F_k+\frac{1}{\lambda}B_k\biggr]^{-1}\hat g_k = -\rho \hat M_k\hat g_k.
\label{eqn_72}
\end{equation}
The last equality of (\ref{eqn_72}) is obtained by combining the two inverses in the inverse of a single matrix, and then using the definition of $\hat M_k$. 

Plugging (\ref{eqn_72}) into (\ref{prop_4_2_over_all_eqn_to_plug_particular_terms_in_more_final}) gives us that our linear term coefficient of the \textit{Q-KLD-WRM} formulation at $k+1$ is given by
\begin{equation}
\begin{split}
g_{k+1}+ \sum_{i=0}^{k}\rho^{k+1-i}\biggl(\lambda \kappa(i) F_i\sum_{j=i}^{k}s_j\biggr) =\\ 
g_{k+1} + \rho(\hat g_k - g_{k}) -\rho\hat M_k\hat g_k =\\ 
g_{k+1} +\rho(I - \hat M_k)\hat g_k-\rho g_k.
\label{prop_4_2_over_all_eqn_to_plug_particular_terms_final}
\end{split}
\end{equation}
Equation (\ref{prop_4_2_over_all_eqn_to_plug_particular_terms_final}), combined with (\ref{eqn_QKLDWRM_for_particualr_model_appendix_k_plus_1}) shows us that the induction target (\ref{eqn_prop_4_2_result_2_inductive_target_1})-(\ref{eqn_prop_4_2_result_1_inductive_target_2}) is satisfied. Thus, the proof is complete $\square$.

\section{The error between EA over K-Factors and EA over K-FAC approximated Fisher}
Our discussion is based on the case when we hold an EA of the Fisher (or any approximation thereof). However, when using \textit{K-FAC} (and also our \textit{KLD-WRM} instantations which use \textit{K-FAC} as a platform), we do \textit{not} hold an EA over the \textit{K-FAC} approximated Fisher like
\begin{equation}
\bar{F}^{(KFAC)}_k = \sum_{i=0}^k \kappa(i) \rho^{k-i}{F}^{(KFAC)}_k
\label{ea_over_Kfac_fisher}
\end{equation}
but rather hold an EA over each of the K-Factors as:
\begin{equation}
\begin{split}
\bar A_k^{(l)} :=\sum_{i=0}^k\kappa(i) \rho^{k-i}A_i^{(l)},\\
\bar \Gamma_k^{(l)} := \sum_{i=1}^k\kappa(i) \rho^{k-i}\Gamma_i^{(l)},
\end{split}
\label{EA_KF_1}
\end{equation}
where $\kappa(i)  = 1-\rho$ if $i\geq 1$, and $\kappa(0) = 1$. This is standard implementation procedure as using (\ref{ea_over_Kfac_fisher}) would undo any benefit from the Kronecker factorization approach.

Thus, strictly speaking, \textit{K-FAC}, as well as the proposed \textit{KLD-WRM} instantiation algorithms lay outside the KLD-WRM family. However, they do approximate algorithms in the \textit{KLD-WRM} family. We now briefly analyze the error. Formal bounds are future work.
\subsection{The Error when using EA over Kronecker Factors}
Let us focus on a single layer $l$. We drop the layer index superscript for convenience. The error between using an EA over the K-FAC approximated Fisher and using an EA over the Kronecker factors is 
\begin{equation}
\begin{split}
\text{err}_k& = \sum_{i=0}^k \rho^{k-i}\kappa(i)A_i\otimes G_i \\&- \biggl(\sum_{i=0}^k \rho^{k-i}\kappa(i)A_i \biggr) \otimes \biggl(\sum_{j=0}^k \rho^{k-j}\kappa(j)G_j \biggr)
\end{split}
\end{equation}
Using the distributive property of the Kronecker Factor, we have
\begin{equation}
\begin{split}
\text{err}_k& = \sum_{i=0}^k \rho^{k-i}\kappa(i)A_i\otimes \biggl( G_i - \sum_{j=0}^k \rho^{k-j}\kappa(j)G_j \biggr)\\
& = \sum_{j=0}^k \rho^{k-j}\kappa(j)\biggl(A_j - \sum_{i=0}^k \rho^{k-i}\kappa(i)A_i \biggr)\otimes G_j.
\end{split}
\label{err_ea_over_kfac_vs_ea_over_fisher_kfac}
\end{equation}
Thus, we see that if either one of the following holds: (1) $A_k$ does not change much, (2) $G_k$ does not change much, or (3) $\rho\approx0$, then the error is small (going to zero if either $A_i$ is constant, $G_i$ is constant or $\rho\to 0 $). While one may expect (1) and/or (2) to hold in practice for very small step-size (at least if the whole\footnote{To avoid changes in $A_i$ and $G_i$ due to picking different batches, in which case the change will only be determined by the change in $\theta$. Of course, when using a batch estimate to $A_i$ and $G_i$, different batches give different mean estimations. However this change would merely be induced by sub-sampling, and would not be present in the exact quantities (or the distribution of the estimates). It is only the change in the distribution of estimates that is of interest, rather than the change in their realization.} data set is used to approximate the expectation), (3) is unrealistic. Further formal investigation of this aspect is required and will be future work.

\section{Efficient Computation of Q-KLD-WRM Step when $B_k = F_k^{(KFAC)}$}
\subsection{Computation of $\hat g_{k+1}$}
To compute the \textit{Q-KLD-WRM} step ($s_{k+1}$) using \textit{Proposition 4.2}, we must first compute 
\begin{equation}
\hat g_{k+1} = g_{k+1} + \rho(I-\hat M_k)\hat g_{k}- \rho g_k,
\label{g_k_hat}
\end{equation}
where
\begin{equation}
\hat M_k :=\big[I +\frac{1}{\lambda}B_k \bar F_k^{-1}\big]^{-1}.
\label{M_k_har}
\end{equation}
Clearly, the bottleneck lies in computing
\begin{equation}
\hat M_{k}\hat g_{k} = \big[I +\frac{1}{\lambda}B_k \bar F_k^{-1}\big]^{-1}\hat g_k,
\label{practical_computation_hat_g_k_plus_1}
\end{equation}
as the remaining operations in (\ref{g_k_hat}) are just vector additions.
First, note that $\hat M_k$ can be written as
\begin{equation}
\hat M_k = \bar F_k \big[\bar F_k +\frac{1}{\lambda}B_k \big]^{-1}
\end{equation}
We hold an EA of the Kronecker-factors, rathar than an EA of the K-FAC matrix, which means $\bar F_k$ is of the form
\begin{equation}
\bar F_k = \text{block-diag}\biggl(\big\{\bar A^{(l)}_k \otimes \bar \Gamma^{(l)}_k\big \}_{l\in\{1,2,...N_l\}}\biggr),
\label{bar_F_k_shape}
\end{equation}
where $N_l$ is the number of layers $\bar A^{(l)}_k$, $\bar \Gamma^{(l)}_k$ are the exponential averages of the standard Kronecker factors (see \cite{KFAC}). Similarly, we chose $B_k = F_k^{(KFAC)}$, so 
\begin{equation}
\bar B_k = \text{block-diag}\biggl(\big\{ A^{(l)}_k \otimes  \Gamma^{(l)}_k\big \}_{l\in\{1,2,...N_l\}}\biggr).
\label{B_k_shape}
\end{equation}
Thus (\ref{practical_computation_hat_g_k_plus_1}) reduces to performing an inversion of the form (for each layer)
\begin{equation}
\big[\hat M_{k}\hat g_{k}\big]_l = (\bar A^{(l)}_k \otimes  \bar \Gamma^{(l)}_k)\biggl[\bar A^{(l)}_k \otimes  \bar \Gamma^{(l)}_k +\frac{1}{\lambda}(A^{(l)}_k \otimes  \Gamma^{(l)}_k)\biggr]^{-1}v_l,
\label{Layer_hat_m_k_g_k}
\end{equation}
There are two ways to approach solving (\ref{Layer_hat_m_k_g_k}).  

The first one is to recognize that computing 
\begin{equation}
u_l = \biggl[\bar A^{(l)}_k \otimes  \bar \Gamma^{(l)}_k +\frac{1}{\lambda}(A^{(l)}_k \otimes  \Gamma^{(l)}_k)\biggr]^{-1}v_l
\end{equation}
amounts to solving a generalized Stein equation, and then use the methods proposed in Appendix B of \cite{KFAC} to get $u_l$. We then obtain $\big[\hat M_{k}\hat g_{k}\big]_l = (\bar A^{(l)}_k \otimes  \bar \Gamma^{(l)}_k) u_l = \Gamma^{(l)}_k U_l \bar A^{(l)}_k$, with $u_l = \text{vec}(U_l)$.

The second approach uses another approximation - in the same spirit as with replacing an EA for the Fisher with an EA over the Kronecker factors. For this approach, we note that when $B_k = F_k$, $\bar F_k +\frac{1}{\lambda}B_k = \sum_i^{k-1}\kappa(i)\rho^{k-i}F_i + \frac{(1-\rho)\lambda + 1}{\lambda}F_k$ is just a re-weighting of the terms in the $\bar F_k$ sum. Thus, instead of carrying EA-averaged Kronecker factors for $\bar F_k$, and then try to work with $\bar F_k + \frac{1}{\lambda}B_k$, we can merely edit the weight of the last Kronecker factor and get a Kronecker-decomposed estimate for $\bar F_k + \frac{1}{\lambda}B_k$ directly. This would be much easier to work with. That is, we save EA-Kronecker factors for two previous levels, and use 
\begin{equation}
\hat A_k^{(l)} = \rho\bar A_{k-1}^{(l)} + \frac{(1-\rho)\lambda+1}{\lambda}A_k^{(l)},
\label{sneaky_trick_A_hat}
\end{equation}
\begin{equation}
\hat \Gamma_k^{(l)} = \rho\bar \Gamma_{k-1}^{(l)} + \frac{(1-\rho)\lambda+1}{\lambda}\Gamma_k^{(l)}
\label{sneaky_trick_G_hat}
\end{equation}
These should be contrasted with the running EA for Kronecker factors
\begin{equation}
\bar A_k^{(l)} = \rho\bar A_{k-1}^{(l)} + (1-\rho)A_k^{(l)},
\end{equation}
\begin{equation}
\bar \Gamma_k^{(l)} = \rho\bar \Gamma_{k-1}^{(l)} + (1-\rho)\Gamma_k^{(l)}.
\end{equation}
We thus have the approximation\footnote{Note that the approximation would be exact if holding EA for the Kronecker factors was the same as holding EA for the big matrix - which is never true in practice.}
\begin{equation}
\big[\hat M_{k}\hat g_{k}\big]_l \approx (\bar A^{(l)}_k \otimes  \bar \Gamma^{(l)}_k)\big[\hat A^{(l)}_k \otimes  \hat \Gamma^{(l)}_k\big]^{-1}v_l.
\end{equation}
We have
\begin{equation}
\begin{split}
\big[\hat M_{k}\hat g_{k}\big]_l \approx (\bar A^{(l)}_k \otimes  \bar \Gamma^{(l)}_k)\big[\hat A^{(l)}_k \otimes  \hat \Gamma^{(l)}_k\big]^{-1}v_l =\\
 (\bar A^{(l)}_k \otimes  \bar \Gamma^{(l)}_k)\big[(\hat A^{(l)}_k)^{-1} \otimes  (\hat \Gamma^{(l)}_k)^{-1}\big]v_l =\\
 \text{vec}\biggl[\Gamma^{(l)}_k (\hat \Gamma^{(l)}_k)^{-1} V_l (\hat A^{(l)}_k)^{-1} ( A^{(l)}_k)\biggr],
\end{split}
\label{easy_hatM_g_computation}
\end{equation}
where $v_l = \text{vec}(V_l)$ - i.e. $V_l\in\mathbb R^{n_l\times n_{l-1}}$ is the matrix format of $v_l$. That is, $v_l$ maps to $V_l$ in the same way $\text{vec}(W_l)$ maps to $W_l$, where $W_l$ is the weight matrix\footnote{As ever with K-FAC, bias is included in the weight matrix through appending a 1 at each layer output - see \cite{KFAC}.} at layer $l$. 

Equation (\ref{easy_hatM_g_computation}) gives us an efficient way of estimating $\hat M_k g_k$. Thus, we can efficiently assemble $\hat g_{k+1}$ using (\ref{g_k_hat}).
\subsection{Computing $s_{k+1}$}
All that is now left, is to compute
\begin{equation}
s_{k+1} = -\frac{1}{\lambda}\biggl[\bar F_{k+1}+\frac{1}{\lambda}B_{k+1}\biggr]^{-1}\hat g_{k+1}.
\end{equation}
In the previous subsection, we have already discussed about the two possible approaches one can take to invert $ F_{k+1}+\frac{1}{\lambda}B_{k+1}$ when $B_{k+1} = F_k^{(KFAC)}$ and we approximate $\bar F_{k+1}$ by holding an EA for the Kronecker factors. Any of the two approaches apply directly. For example, taking the second approach, we get
\begin{equation}
[s_{k+1}]_l \approx= -\frac{1}{\lambda}\text{vec}\biggl[(\hat \Gamma_{k+1}^{(l)})^{-1} \hat G^{(l)}_{k+1}(\hat A^{(l)}_{k+1})^{-1} \biggr],
\label{eqn_Q_kld_wrm_step}
\end{equation}
where $\text{vec} (\hat G_{k+1}^{(l)}) = [\hat g_{k+1}]_l$. That is, $[\hat g_{k+1}]_l$ maps to $\hat G_{k+1}^{(l)}$ in the same way $\text{vec}(W_l)$ maps to $W_l$, where $W_l$ is the weight matrix at layer $l$.

Equation (\ref{eqn_Q_kld_wrm_step}) gives us an efficient way to estimate the \textit{Q-KLD-WRM} step provided we have $\hat g_{k+1}$ (which we can efficiently get as in (\ref{easy_hatM_g_computation})).

\section{KL-Divergence in terms of Network Output}
\subsection{Regression with Gaussian Predictive Distributions of Fixed Variance}
In this section, we derive the fact that the symmetric KL-Divergence is
\begin{equation}
\begin{split}
\mathbb D_{KL}\big(p(y|h_{\theta_1}(x_j)) , p(y|h_{\theta_2}(x_j))\big) = \\ \frac{1}{2}\norm{h_{\theta_1}(x_j) - h_{\theta_2}(x_j) }_2^2.
\end{split}
\end{equation}
when the predictive distribution of our net is of the form
\begin{equation}
p(y|h_{\theta_k}(x_j)) = \mathcal N(y|h_{\theta_k}(x_j); I).
\label{predictive_distribution_4_3}
\end{equation}
That is, $p(y|h_{\theta_k}(x_j))$ is a normal distribution with mean $h_{\theta_1}(x_j)$ and covariance matrix $I$. This boils down to computing the KL-divergence between two multivariate Gaussians having the same covariance matrix. Let us denote for convenience $\mu_{1,j}:=h_{\theta_1}(x_j)$, $\mu_{2,j}:=h_{\theta_2}(x_j)$. Then, we have the forward KL divergence
\begin{equation}
\begin{split}
\mathbb D_{KL}\big(p(y|\mu_{1,j}) \,\big|\big|\, p(y|\mu_{2,j})\big) =\\  \mathbb E_{y\sim \mathcal N (\mu_{1,j}| I)}\biggl[-\frac{1}{2}\norm{y- \mu_{1,j}}^2_2 + \frac{1}{2}\norm{y- \mu_{2,j}}^2_2\biggr]= \\
-\frac{1}{2}\mathbb E_{y\sim \mathcal N (\mu_{1,j}| I)}\biggl[\mu_{1,j}^T\mu_{1,j} + 2y^T(\mu_{2,j} - \mu_{1,j}) - \mu_{2,j}^T\mu_{2,j}\biggr] = \\
-\frac{1}{2}\biggl[\mu_{1,j}^T\mu_{1,j} + 2\mu_{1,j}^T(\mu_{2,j} - \mu_{1,j}) - \mu_{2,j}^T\mu_{2,j}\biggr].
\end{split}
\end{equation}
Thus, we have 
\begin{equation}
\mathbb D_{KL}\big(p(y|\mu_{1,j}) \,\big|\big|\, p(y|\mu_{2,j})\big) = \frac{1}{2}\norm{\mu_{2,j} - \mu_{1,j}}_2^2.
\label{eqn_kl_div_symmetric_fwd_and_bwd_appendix}
\end{equation}
By exchanging the indices in (\ref{eqn_kl_div_symmetric_fwd_and_bwd_appendix}) we see that the backwards KL-divergence is equal to the forwards KL divergence
\begin{equation}
\begin{split}
\mathbb D_{KL}\big(p(y|\mu_{1,j}) \,\big|\big|\, p(y|\mu_{2,j})\big) = &\mathbb D_{KL}\big(p(y|\mu_{2,j}) \,\big|\big|\, p(y|\mu_{1,j})\big) = \\&\frac{1}{2}\norm{\mu_{2,j} - \mu_{1,j}}_2^2.
\end{split}
\end{equation}
Thus, we have
\begin{equation}
\begin{split}
\mathbb D_{KL}\big(p(y|\mu_{1,j}), p(y|\mu_{2,j})\big) &=\, \mathbb D_{KL}\big(p(y|\mu_{1,j}) \,\big|\big|\, p(y|\mu_{2,j})\big)\\ & =\frac{1}{2}\norm{\mu_{2,j} - \mu_{1,j}}_2^2
\end{split},
\label{eqn_kl_div_symmetric_KL_normal_distrib_I_variance}
\end{equation}
which is our claim $\square$.
\subsection{Classification: Categorical Distribution}
In this case, our predictive distribution is given by
\begin{equation}
p(y|h_{\theta_k}(x_j)) = \mathcal S\big(h_{\theta_k}(x_j)\big)[y],
\label{predictive_distribution_classif}
\end{equation}
where $h_{\theta_k}(x_j)\in \mathbb R^{n_c}$ is the output of the net, $\mathcal S(\cdot)$ is the Softmax function, $n_c$ is the number of possible classes, and $\mathcal S[y]$ represents the $y^{\text{th}}$ entry of (probability) vector $\mathcal S$. We have the forward KL-Divergence
\begin{equation}
\begin{split}
\mathbb D_{KL}\big(p(y|h_{\theta_1}(x_j) \,\big|\big|\, p(y|h_{\theta_2}(x_j)\big) =\\ \sum_{i=1}^{n_c}p(i|h_{\theta_1}(x_j)\log \frac{p(i|h_{\theta_1}(x_j))}{p(i|h_{\theta_2}(x_j))},
\end{split}
\label{KL_div_fwd_categorical}
\end{equation}
and the forward KL-Divergence
\begin{equation}
\begin{split}
\mathbb D_{KL}\big(p(y|h_{\theta_2}(x_j) \,\big|\big|\, p(y|h_{\theta_1}(x_j)\big) =\\ \sum_{i=1}^{n_c}p(i|h_{\theta_2}(x_j)\log \frac{p(i|h_{\theta_2}(x_j))}{p(i|h_{\theta_1}(x_j))}.
\end{split}
\label{KL_div_bwd_categorical}
\end{equation}
Since we can get the vectors $ h_{\theta_1}(x_j)$ and $ h_{\theta_2}(x_j)$ by merely saving the old parameter $\theta_1$ (and we have $\theta_2$ as the current parameter) and performing 2 forward passes, we can compute both the forward KL divergence in (\ref{KL_div_fwd_categorical}) and the backward KL-divergence in (\ref{KL_div_bwd_categorical}) easily (as $n_c$ is typically very small). We can then obtain the symmetric KL divergence as
\begin{equation}
\begin{split}
\mathbb D_{KL}\big(p(y|h_{\theta_1}(x_j) , p(y|h_{\theta_1}(x_j)\big) =\\
\frac{1}{2}\sum_{i=1}^{n_c}\mathcal S\big(h_{\theta_1}(x_j)\big)[i]\cdot\log \frac{\mathcal S\big(h_{\theta_1}(x_j)\big)[i]}{\mathcal S\big(h_{\theta_2}(x_j)\big)[i]} \\ 
+\frac{1}{2} \sum_{i=1}^{n_c}\mathcal S\big(h_{\theta_2}(x_j)\big)[i]\cdot \log \frac{\mathcal S\big(h_{\theta_2}(x_j)\big)[i]}{\mathcal S\big(h_{\theta_1}(x_j)\big)[i]}.
\end{split}
\label{KL_div_sym_categorical}
\end{equation}

\section{Extension of KLD-WRM to variable step-size}
Note that for \textsc{w}o\textsc{qm}, \textsc{so-kld-wrm} and \textsc{q-kld-wrm}, the parameter $1/\lambda$ plays the role of a step-size. In the main body, we considered only cases when $\lambda$ is fixed across different locations $\theta_k$, that is, a fixed step-size. For \textsc{kld-wrm}, a fixed $\lambda$ at all steps can also be interpreted as a fixed `strength' of KL-wake-regularization.

Our established equivalence between \textsc{fisher-w}o\textsc{qm} and \textsc{ea-ng} holds for fixed step-size only, and in fact, one can check that it does not hold if we consider varying $\lambda$ with $\theta_k$. However, one might want to consider variable $\lambda$ (thus variable step-size/KL-wake-regularization strength) in our \textsc{kld-wrm} algorithms. That is, one might want to consider replacing $\lambda$ in \textit{equation (23) of the main body }with $\lambda^{(k)}$, whose schedule can be set as the inverse of the desired learning-rate schedule. It turns out that if we do this, \textit{Propositions 4.1} and \textit{4.2} only change mildly. To incorporate variable $\lambda^{(k)}$ in \textsc{q-kld-wrm}, all one has to do is to replace (\ref{eqn_prop_4_2_result_2}) with $\hat g_{k+1} = g_{k+1} + (\lambda^{(k+1)}/\lambda^{(k)})\rho[\hat g_{k}-g_k - \hat M_k\hat g_{k}]$ and all $\lambda$'s in \textit{Proposition 4.2} with $\lambda^{(k)}$. This fact allows us to easily use variable $\lambda^{(k)}$ in practice. The change to \textit{Proposition 4.1} can be directly obtained by noting that it is a particular case of \textit{Proposition 4.2} with $B_k=0$. \textit{Proposition 4.2} modified to allow for variable $\lambda^{(k)}$, as well as its proof are presented in the next subsection.

\subsection{Extension of Proposition 4.2 to Variable $\lambda$}
Before we begin, we restate the \textit{Q-KLD-WRM} formulation with variable $\lambda$ for clarity. The \textit{Q-KLD-WRM} step at $\theta_k$, with variable $\lambda^{(k)}$, is the solution to
\begin{equation}
\min_s g_k^Ts + \frac{1}{2}s^TB_ks +  \lambda^{(k)}\sum_{i=0}^k\kappa(i)\rho^{k-i}\tilde{\mathbb D}_{KL}(\theta_i||\theta_{k} + s),
\label{appendix_eqn_Q_KLD_WRM_variable_lambda_0}
\end{equation}
where $B_k$ is a curvature matrix which aims to approximate the Hessian $H_k$, and $\lambda^{(k)}$ is the level of KL-wake-regularization (or the inverse step-size) which is allowed to vary with $k$. Equation (\ref{appendix_eqn_Q_KLD_WRM_variable_lambda_0}) tells us that the \textit{Q-KLD-WRM} step at $\theta_k$ (with variable $\lambda^{(k)}$) solves
\begin{equation}
\begin{split}
\min_s s^T\biggl[g_k + \sum_{i=0}^{k-1}\rho^{k-i}\biggl(\lambda^{(k)} \kappa(i) F_i\sum_{j=i}^{k-1}s_j\biggr)\biggr]\\
+\frac{\lambda^{(k)}}{2}s^T\biggl[\sum_{i=0}^k\kappa(i)\rho^{k-i}F_i + \frac{1}{\lambda^{(k)}}B_k\biggr]s.
\end{split}
\label{appendix_eqn_Q_KLD_WRM_variable_lambda_}
\end{equation}
\begin{Proposition_4_2_extension}
	The \textit{Q-KLD-WRM} step $s_k$ at location $\theta_k$, with variable $\lambda^{(k)}$, is given by the solution to the problem
	\begin{equation}
	\min_s s^T\hat g_k + \frac{\lambda^{(k)}}{2}s^T\biggl[\bar F_k + \frac{1}{\lambda^{(k)}}B_k\biggr]s
	\label{eqn_prop_4_2_result_0_extension}
	\end{equation}
	
	where $\hat g_k$ is given by the one-step recursion
	\begin{equation}
	\hat g_{k+1} = g_{k+1} + \frac{\lambda^{(k+1)}}{\lambda^{(k)}}\rho\biggl[\hat g_{k}-g_k - \hat M_k\hat g_{k}\biggr],\,\,\forall k\in \mathbb Z^+,
	\label{eqn_prop_4_2_result_2_extension}
	\end{equation}
	with $\hat g_0 := g_0$, $\bar F_k := \sum_{i=0}^k\kappa(i)\rho^{k-i}F_i$, and 
	\begin{equation}
	\hat M_k :=\big[I +\frac{1}{\lambda^{(k)}}B_k \bar F_k^{-1}\big]^{-1}.
	\label{eqn_prop_4_2_result_3_extension}
	\end{equation}
	That is, the \textit{Q-KLD-WRM} step is formally given by
	\begin{equation}
	s_{k}= -\frac{1}{\lambda^{(k)}}\biggl[\bar F_k + \frac{1}{\lambda^{(k)}}B_k\biggr]^{-1}\hat g_k.
	\label{eqn_prop_4_2_result_1_extension}
	\end{equation}
\end{Proposition_4_2_extension}

\textit{Proof.} By induction. The induction is very similar to the proof of \textit{Proposition 4.2}.

\textit{Induction check.}
For $k=0$, we have the model in (\ref{appendix_eqn_Q_KLD_WRM_variable_lambda_}) reducing to
\begin{equation}
\min_s s^Tg_0 + \frac{\lambda^{(0)}}{2}s^T\biggl[\bar F_0 +\frac{1}{\lambda^{(0)}}B_0\biggr]s
\end{equation}
Now, noting that $\hat g_0 = g_0$ by definition, we see that the \textit{Q-KLD-WRM} linear\footnote{The Quadratic term obviously also obeys \textit{Propositon A.1 formulation} - but this fact follows trivially from the original \textit{Q-KLD-WRM} definition, as we can see.} term obeys \textit{Proposition A.1} at $k=0$. Thus, 
\begin{equation}
s_0 = -\frac{1}{\lambda^{(0)}}\biggl[\bar F_0 +\frac{1}{\lambda^{(0)}}B_0\biggr]^{-1}\hat g_0.
\label{eqn_prop_4_2_s_0_ind_check_variable_lambda}
\end{equation}

For $k=1$, we have the model in (\ref{appendix_eqn_Q_KLD_WRM_variable_lambda_}) reducing to
\begin{equation}
\min_s s^T[g_1 +\rho \lambda^{(1)} F_0 s_0] + \frac{\lambda^{(1)}}{2}s^T\biggl[\bar F_1+\frac{1}{\lambda^{(1)}}B_1\biggr]s.
\end{equation}
Now, by using (\ref{eqn_prop_4_2_s_0_ind_check_variable_lambda}), we have that
\begin{equation}
\rho \lambda^{(1)} F_0 s_0 = -\rho \frac{\lambda^{(1)}}{\lambda^{(0)}} \big[F_0^{-1}\big]^{-1} \biggl[\bar F_0 +\frac{1}{\lambda^{(0)}}B_0\biggr]^{-1}\hat g_0 
\end{equation}
\begin{equation}
\begin{split}
\rho \lambda^{(1)} F_0 s_0 = &-\rho \frac{\lambda^{(1)}}{\lambda^{(0)}} \biggl[\bar F_0 F_0^{-1} +\frac{1}{\lambda^{(0)}}B_0 F_0^{-1}\biggr]^{-1}\hat g_0 = \\ &-\rho \frac{\lambda^{(1)}}{\lambda^{(0)}} \hat M_0\hat g_0
\end{split}
\end{equation}
where the latter equality holds by the definition of $\hat M_k$ evaluated at $k=0$ and the fact that $\bar F_0 = F_0$. Thus,
\begin{equation}
g_1 +\rho \lambda^{(1)} F_0 s_0 = g_1 - \rho\frac{\lambda^{(1)}}{\lambda^{(0)}} \hat M_0 \hat g_0, \label{introducing_sneaky_terms_prop_4_2_ind_check_s_1_variable_lambda}
\end{equation}
\begin{equation}
g_1 +\rho \lambda^{(1)} F_0 s_0 = g_1 + \rho\frac{\lambda^{(1)}}{\lambda^{(0)}} \biggl[\hat g_0 - g_0 - \hat M_0 \hat g_0 \biggr] \label{introducing_sneaky_terms_prop_4_2_ind_check_s_1_variable_lambda_2}
\end{equation}
The latter equality in (\ref{introducing_sneaky_terms_prop_4_2_ind_check_s_1_variable_lambda_2}) holds because $g_0=\hat g_0$. Consequently, using the recursive definition of $\hat g_{k+1}$ for $k=0$, we get
\begin{equation}
g_1 +\rho \lambda^{(1)} F_0 s_0 = \hat g_1\label{hat_g_1_prop_4_2_ind_check_s_1_variable_lambda},
\end{equation}
which shows us that the \textit{Q-KLD-WRM} linear term obeys the formulation in \textit{Proposition A.1}. Thus the whole \textit{Q-KLD-WRM} formulation obeys formulation (\ref{eqn_prop_4_2_result_0_extension}). In particular, we have
\begin{equation}
s_1 = -\frac{1}{\lambda^{(1)}}\biggl[\bar F_1+\frac{1}{\lambda^{(1)}}B_1\biggr]^{-1}\hat g_1.
\label{eqn_prop_4_2_s_1ind_check_variable_lambda}
\end{equation}

For $k=2$, we have the model in (\ref{appendix_eqn_Q_KLD_WRM_variable_lambda_}) reducing to
\begin{equation}
\begin{split}
&\min_s \frac{\lambda^{(2)}}{2}s^T\biggl[\bar F_2 + \frac{1}{\lambda^{(2)}}B_2\biggr]s+\\
s^T\biggl[g_2 + &\rho^2\lambda^{(2)} F_0(s_0 + s_1) + \rho\lambda^{(2)} (1-\rho)F_1s_1\biggr].
\end{split}
\label{eqn_s_2_prop_4_2_variable_lambda}
\end{equation}
The linear term in (\ref{eqn_s_2_prop_4_2_variable_lambda}) can be rearranged as
\begin{equation}
\begin{split}
s^T\biggl[g_2 + &\rho^2\lambda^{(2)} F_0 s_0  + \rho\lambda^{(2)} [\rho F_0 + (1-\rho)F_1]s_1\biggr] = \\
&s^T\biggl[g_2 + \rho^2 \lambda^{(2)} \bar F_0 s_0  + \rho\lambda^{(2)} \bar F_1s_1\biggr],
\end{split}
\label{eqn_s_2_prop_4_2_part_2_variable_lambda}
\end{equation}
where to get the second equality we used the definition of $\bar F_k$ applied at $k=0$ and $k=1$. We have already seen when doing the check for $k=1$ (from (\ref{hat_g_1_prop_4_2_ind_check_s_1_variable_lambda})) that
\begin{equation}
\rho^2  \bar F_0 s_0  = \rho \frac{1}{\lambda^{(1)}}(\hat g_1 - g_1 )
\label{eqn_g_1_hat_2nd_time_prop_4_2_variable_lambda}
\end{equation}
Thus, we have 
\begin{equation}
\rho^2\lambda^{(2)}\bar F_0 s_0 = \rho \frac{\lambda^{(2)}}{\lambda^{(1)}} (\hat g_1 - g_1).
\label{eqn_s_2_prop_4_2_part_2_variable_lambda_2}
\end{equation}
So we only need to work out the $\rho\lambda^{(2)} \bar F_1s_1$ term in (\ref{eqn_s_2_prop_4_2_part_2_variable_lambda}). By using (\ref{eqn_prop_4_2_s_1ind_check_variable_lambda}), this reads
\begin{equation}
\rho\lambda^{(2)} \bar F_1s_1 = - \rho \frac{\lambda^{(2)}}{\lambda^{(1)}} \big[\bar F_1^{-1}\big]^{-1} \biggl[\bar F_1+\frac{1}{\lambda^{(1)}}B_1\biggr]^{-1}\hat g_1
\end{equation}
\begin{equation}
\begin{split}
\rho \lambda^{(2)} \bar F_1 s_1 =& -\rho \frac{\lambda^{(2)}}{\lambda^{(1)}} \biggl[\bar F_1 \bar F_1^{-1} +\frac{1}{\lambda^{(1)}}B_1 \bar F_1^{-1}\biggr]^{-1}\hat g_1 =\\
& -\rho \frac{\lambda^{(2)}}{\lambda^{(1)}} \hat M_1\hat g_1
\label{prop_4_2_ind_check_s_2_M_hat_term_variable_lambda}
\end{split}
\end{equation}
where the latter equality holds by the definition of $M_k$ evaluated at $k=1$. Thus, plugging in (\ref{prop_4_2_ind_check_s_2_M_hat_term_variable_lambda}) and (\ref{eqn_s_2_prop_4_2_part_2_variable_lambda_2}) in the linear term expression (\ref{eqn_s_2_prop_4_2_part_2_variable_lambda}),
we get that the linear term reads
\begin{equation}
\begin{split}
s^T\biggl[g_2 + &\rho^2\lambda F_0 s_0  + \rho\lambda [\rho F_0 + (1-\rho)F_1]s_1\biggr] = \\ 
&s^T\biggl[g_2 + \rho\frac{\lambda^{(2)}}{\lambda^{(1)}}(\hat g_1 - g_1-\hat M_1\hat g_1) \biggr] = s^T \hat g_2,
\end{split}
\label{eqn_s_2_linear_term_final_prop_4_2_part_2_variable_lambda}
\end{equation}
where in the last equality we used the recursive definition of $\hat g_{k+1}$ from \textit{Proposition A.1}, evaluated at $k=1$. Thus, (\ref{eqn_s_2_linear_term_final_prop_4_2_part_2}) shows us that the \textit{Q-KLD-WRM} linear term obeys the formulation in \textit{Proposition 4.2}. Thus the whole \textit{Q-KLD-WRM} formulation obeys formulation (\ref{eqn_prop_4_2_result_0_extension}). In particular, if we now plug the linear term in (\ref{eqn_s_2_linear_term_final_prop_4_2_part_2_variable_lambda}) back into the full model (\ref{eqn_s_2_prop_4_2_variable_lambda}), we get that the \textit{Q-KLD-WRM} step satisfies
\begin{equation}
s_2 = -\frac{1}{\lambda} \biggl[\bar F_2 + \frac{1}{\lambda}B_2\biggr]^{-1}\hat g_2.
\label{eqn_prop_4_2_s_2ind_check_variable_lambda}
\end{equation}
We are done with the induction check.

\textit{Induction main body.} 
Assume for the \textbf{inductive hypothesis} that we have for some $k\geq 2$
\begin{equation}
s_{l}= -\arg\min_s s^T\hat g_l + \frac{\lambda^{(l)}}{2}s^T\biggl[\bar F_l + \frac{1}{\lambda^{(l)}}B_l\biggr]s,\,\forall l\in\{0,..,k\}
\label{eqn_prop_4_2_result_1_inductive_hypothesis_variable_lambda}
\end{equation}
\begin{equation}
\hat g_{l} = g_{l} + \frac{\lambda^{(l)}}{\lambda^{(l-1)}}\rho\biggl[\hat g_{l-1}-g_{l-1} - \hat M_{l-1}\hat g_{l-1}\biggr]\,\,\,\forall l\in\{1,...,k\},
\label{eqn_prop_4_2_result_2_inductive_hypothesis_variable_lambda}
\end{equation}
with $\hat g_0 := g_0$, $\bar F_k := \sum_{i=0}^k\kappa(i)\rho^{k-i}F_i$, and 
\begin{equation}
\hat M_l :=\big[I +\frac{1}{\lambda}B_l \bar F_l^{-1}\big]^{-1}.
\label{eqn_prop_4_2_result_3_inductive_hypothesis_variable_lambda}
\end{equation}
We \textbf{need to prove} that at $k+1$, for all $k\geq 2$, we have
\begin{equation}
\hat g_{k+1} = g_{k+1} + \frac{\lambda^{(k+1)}}{\lambda^{(k)}}\rho\biggl[\hat g_{k}-g_k - \hat M_k\hat g_{k}\biggr]
\label{eqn_prop_4_2_result_2_inductive_target_1_variable_lambda}
\end{equation}
with $\hat M_k$ defined as in (\ref{eqn_prop_4_2_result_3_inductive_hypothesis_variable_lambda}), and
\begin{equation}
\begin{split}
s_{k+1}= &\arg\min_s s^T\hat g_{k+1} \\ +& \frac{\lambda^{(k+1)}}{2}s^T\biggl[\bar F_{k+1} + \frac{1}{\lambda^{(k+1)}}B_{k+1}\biggr]s.
\end{split}
\label{eqn_prop_4_2_result_1_inductive_target_2_variable_lambda}
\end{equation}
Since the inductive check holds at $k\in \{0,1,2\}$, this would imply \textit{Proposition A.1} holds.

To begin, consider (\ref{appendix_eqn_Q_KLD_WRM_variable_lambda_}) evaluated at $k+1$. We have that the \textit{Q-KLD-WRM} step at $k+1$, $s_{k+1}$ is the solution to
\begin{equation}
\begin{split}
\min_s\,\,\, \frac{\lambda^{(k+1)}}{2}s^T\biggl[\bar F_{k+1} +\frac{1}{\lambda^{(k+1)}}B_{k+1}\biggr]s\\
+s^T\biggl[g_{k+1}+ \sum_{i=0}^{k}\rho^{k+1-i}\biggl(\lambda^{(k+1)} \kappa(i) F_i\sum_{j=i}^{k}s_j\biggr)\biggr].
\end{split}
\label{eqn_QKLDWRM_for_particualr_model_appendix_k_plus_1_variable_lambda}
\end{equation}
Let us now consider the coefficient of the linear term in (\ref{eqn_QKLDWRM_for_particualr_model_appendix_k_plus_1}). It reads
\begin{equation}
\begin{split}
g_{k+1}+ \sum_{i=0}^{k}\rho^{k+1-i}\biggl(\lambda^{(k+1)} \kappa(i) F_i\sum_{j=i}^{k}s_j\biggr) =\\
g_{k+1}+ \lambda^{(k+1)}\sum_{i=0}^{k}\sum_{j=i}^{k}\rho^{k+1-i} \kappa(i) F_i s_j = \\
g_{k+1}+ \lambda^{(k+1)}\sum_{j=0}^{k}\biggl[\biggl(\sum_{i=0}^{j}\rho^{k+1-i} \kappa(i) F_i\biggr)s_j\biggr] =\\
g_{k+1} + \lambda^{(k+1)} \sum_{j=0}^k\rho^{k+1-j}\bar F_j s_j = \\
 g_{k+1} + \rho  \sum_{j=0}^k\rho^{k-j}\lambda^{(k+1)} \bar F_j s_j .
\label{eqn_linear_term_prop_4_2_appendix_proof_variable_lambda}
\end{split}
\end{equation}
To get from the third to the fourth line in (\ref{eqn_linear_term_prop_4_2_appendix_proof_variable_lambda}) we used the definition of $\bar F_k$ evaluated at $k=j$. We can further write the linear coeffcient of (\ref{eqn_QKLDWRM_for_particualr_model_appendix_k_plus_1_variable_lambda}) as
\begin{equation}
\begin{split}
g_{k+1}+ \sum_{i=0}^{k}\rho^{k+1-i}\biggl(\lambda^{(k+1)} \kappa(i) F_i\sum_{j=i}^{k}s_j\biggr) =\\ 
g_{k+1} + \rho  \sum_{j=0}^k\rho^{k-j}\lambda^{(k+1)} \bar F_j s_j = \\
g_{k+1} + \rho  \frac{\lambda^{(k+1)}}{\lambda^{(k)}}\sum_{j=0}^{k-1}\rho^{k-j}\lambda^{(k)} \bar F_j s_j +\rho\lambda^{(k+1)} \bar F_ks_k.
\label{prop_4_2_over_all_eqn_to_plug_particular_terms_in_variable_lambda}
\end{split}
\end{equation}
Now, considering the same linear term of our \textit{Q-KLD-WRM} formulation, \textit{but at arbitrary} $l\in[1,k]\cap \mathbb Z$ (rather than at $k+1$) we get can apply\footnote{We can do that because our calculations (\ref{eqn_QKLDWRM_for_particualr_model_appendix_k_plus_1_variable_lambda}) - (\ref{prop_4_2_over_all_eqn_to_plug_particular_terms_in_variable_lambda}) hold for arbitrary $k\in \mathbb Z^+$, and so does the \textit{Q-KLD-WRM} formulation (\ref{prop_4_2_over_all_eqn_to_plug_particular_terms_in_variable_lambda}).} equation (\ref{prop_4_2_over_all_eqn_to_plug_particular_terms_in_variable_lambda}) to get
\begin{equation}
\begin{split}
g_{l}+ \sum_{i=0}^{l-1}\rho^{l-i}\biggl(\lambda^{(l)} \kappa(i) F_i\sum_{j=i}^{k}s_j\biggr) =  
g_{l} + \sum_{j=0}^{l-1}\rho^{l-j}\lambda^{(l)} \bar F_j s_j.
\end{split}
\end{equation}
But the inductive hypothesis, (\ref{eqn_prop_4_2_result_1_inductive_hypothesis_variable_lambda}) and (\ref{eqn_prop_4_2_result_2_inductive_hypothesis_variable_lambda}), tells us that this linear term must also be $\hat g_l$. That is, we have
\begin{equation}
\hat g_l = g_{l} + \sum_{j=0}^{l-1}\rho^{l-j}\lambda^{(l)} \bar F_j s_j,
\end{equation}
and therefore 
\begin{equation}
\sum_{j=0}^{l-1}\rho^{l-j}\lambda^{(l)} \bar F_j s_j = \hat g_l - g_{l}.
\label{eqn_ghat_g_l_minus_g_l_variable_lambda}
\end{equation}
In particular (\ref{eqn_ghat_g_l_minus_g_l_variable_lambda}) must also hold for $l=k$, because the induction hypothesis holds up until $\hat g_k$. We have 
\begin{equation}
\sum_{j=0}^{k-1}\rho^{k-j}\lambda^{(k)} \bar F_j s_j = \hat g_k - g_{k},
\label{eqn_ghat_g_k_minus_g_k_variable_lambda}
\end{equation}
Now, plugging (\ref{eqn_ghat_g_k_minus_g_k_variable_lambda}) into (\ref{prop_4_2_over_all_eqn_to_plug_particular_terms_in_variable_lambda}), we have
\begin{equation}
\begin{split}
g_{k+1}+ \sum_{i=0}^{k}\rho^{k+1-i}\biggl(\lambda^{(k+1)} \kappa(i) F_i\sum_{j=i}^{k}s_j\biggr) =\\ 
g_{k+1} + \rho\frac{\lambda^{(k+1)}}{\lambda^{(k)}}(\hat g_k - g_{k}) +\rho\lambda^{(k+1)} \bar F_ks_k,
\label{prop_4_2_over_all_eqn_to_plug_particular_terms_in_more_final_variable_lambda}
\end{split}
\end{equation}
so all is left is to compute the term $\rho\lambda^{(k+1)} \bar F_ks_k$. To do that, we use the induction hypothesis again. In particular, we use (\ref{eqn_prop_4_2_result_1_inductive_hypothesis_variable_lambda}) for $l=k$ and solve exactly for $s_k$, to get
\begin{equation}
s_k = -\frac{1}{\lambda^{(k)}}\biggl[\bar F_k+\frac{1}{\lambda^{(k)}}B_k\biggr]^{-1}\hat g_k
\label{s_k_proof_prop_4_2_variable_lambda}.
\end{equation}
Using (\ref{s_k_proof_prop_4_2}), we have
\begin{equation}
\begin{split}
\rho\lambda^{(k+1)} \bar F_ks_k &= -\rho \frac{\lambda^{(k+1)}}{\lambda^{(k)}}\big[\bar F_k^{-1}\big]^{-1}\biggl[\bar F_k+\frac{1}{\lambda^{(k)}}B_k\biggr]^{-1}\hat g_k \\ &= -\rho \frac{\lambda^{(k+1)}}{\lambda^{(k)}} \hat M_k\hat g_k.
\label{eqn_72_variable_lambda}
\end{split}
\end{equation}
The last equality of (\ref{eqn_72_variable_lambda}) is obtained by combining the two inverses in the inverse of a single matrix, and then using the definition of $\hat M_k$, (\ref{eqn_prop_4_2_result_3_extension}). 

Plugging (\ref{eqn_72_variable_lambda}) into (\ref{prop_4_2_over_all_eqn_to_plug_particular_terms_in_more_final_variable_lambda}) gives us that our linear term coefficient of the \textit{Q-KLD-WRM} formulation at $k+1$ is given by
\begin{equation}
\begin{split}
g_{k+1}+ \sum_{i=0}^{k}\rho^{k+1-i}\biggl(\lambda \kappa(i) F_i\sum_{j=i}^{k}s_j\biggr) =\\ 
g_{k+1} + \rho\frac{\lambda^{(k+1)}}{\lambda^{(k)}}(\hat g_k - g_{k}) -\rho\frac{\lambda^{(k+1)}}{\lambda^{(k)}}\hat M_k\hat g_k =\\ 
g_{k+1} + \rho\frac{\lambda^{(k+1)}}{\lambda^{(k)}}\big[\hat g_k - g_{k} -\hat M_k\hat g_k\big].
\label{prop_4_2_over_all_eqn_to_plug_particular_terms_final_variable_lambda}
\end{split}
\end{equation}
Equation (\ref{prop_4_2_over_all_eqn_to_plug_particular_terms_final_variable_lambda}), combined with (\ref{eqn_QKLDWRM_for_particualr_model_appendix_k_plus_1_variable_lambda}) shows us that the induction target (\ref{eqn_prop_4_2_result_2_inductive_target_1_variable_lambda})-(\ref{eqn_prop_4_2_result_1_inductive_target_2_variable_lambda}) is satisfied. Thus, the proof is complete $\square$.

\section{Further Numerical Experiments}
In this section, we first give the implementation details separately for each of the four considered solvers (K-FAC, SO-KLD-WRM, Q-KLD-WRM, QE-KLD-WRM). We then give the network architecture and finally present our numerical results in more detail than in the main body (looking at more metrics and more in-depth comments). The considered problem was MNIST classification.
\subsection{Optimizer Implementation Details}
\textbf{Kronecker Factors Regularization:} Typically, a Levenberg-Marquardt style regularization is used with the Kronecker factors in K-FAC to implicitly decide the step size (see \textit{Section 6} in \cite{KFAC}) . For our network architecture, the Kronecker factors were full rank as long as the batch size was reasonably large. We found the optimal batch size to be 512, and this was well above the limit at which the Kronecker factors became low rank. Thus, we avoided using the complicated Kronecker-factor regularization method described in the original K-FAC paper (\cite{KFAC}). 

Instead, we simply used a constant learning rate, and added a constant regularization factor of $0.01$ to the Kronecker factor's eigenvalues (when doing the eigen-decomposition) in order to avoid infinite stepsize. This performed well enough in practice for all solvers including K-FAC (note that all our examined KLD-WRM implementations are based on the Kronecker factors in K-FAC).

\textbf{Learning Rates and $\lambda$:} The learning rate for \textit{K-FAC} was set to $0.01$. In all KLD-WRM cases, we set $\lambda = 100$ (which is roughly equivalent\footnote{We can see that by thinking of K-FAC as a practical implementation of EA-NG, which is of the form shown in equation (22) in the main body: in this case $1/\lambda$ is the learning rate.} to a learning rate of $0.01$). 

\textbf{Step/Gradient Clipping:} For K-FAC and SO-KLD-WRM the clipping strategy of our \textit{``K-FAC library codes\footnote{These are the codes we built our codes on. In particular, we used the \textit{kfac.py} file at \url{https://github.com/ikostrikov/pytorch-a2c-ppo-acktr-gail/blob/master/a2c_ppo_acktr/algo/kfac.py}.}''} was used, with a clip parameter of $0.1$. For Q-KLD-WRM and QE-KLD-WRM we used a slightly different clipping strategy: clip each parameter group individually if $\norm{\theta_{G_i}}_2/\sqrt{|\theta_{G_i}|} > \tau$, where $\theta_{G_i}$ is the parameters in group $G_i$ and $\tau$ is a clipping threshold. In our experiments we set $\tau = 2$ throughout.

The reason for using different strategies with different solvers was to try altering the step by the least amount when clipping in each case, while avoiding blow-ups. From our numerical trials, it seemed that the outlined choices fit our purpose. Note that clipping matters very little for our results. One reason for that is because it only very rarely activates, and whenever it does so, it will be in the initial phase of the training. We hereby focus on performance metrics that have to do much more with mid and end-training results - in other words, we do not claim (nor investigate whether) KLD-WRM implementations perform better than K-FAC in extremely low-epoch budget regime.

\textbf{Updating the Kronecker Factors and Computing their Inverse:} It is typical that the eigen-decomposition (i.e.\ computing inverses) of the Kronecker factors is performed with lower frequency than updating the Kronecker factors (the statistics) for maximal computation efficiency \cite{KFAC,Kazuki_SPNGD}. However, we performed both of them with the same frequency (every 30 steps) across all our four investigated solvers. We chose to do so in order to avoid the complications which arise with KLD-WRM instantiations when having these two frequencies different (which we did not discuss here). While K-FAC would perform better if we increase the statistics update frequency and keep the eigen-decomposition frequency constant, this would also happen for KLD-WRM. Thus, we believe that our comparison is fair, even if we set the two update frequencies to be equal (which is empirically observed to be sub-optimal).

\textbf{Other Details:} We used \textit{weight decay} with parameter of $0.001$, and a batch size of $512$ throughout.

\subsubsection{QE-KLD-WRM Specific Implementation Details}
For K-FAC, SO-KLD-WRM and Q-KLD-WRM there are no specific implementation details, and all these fall under the generic details we have just described. However, for QE-KLD-WRM we need to (approximately) solve the QE-KLD-WRM sub-problem in equation (36) of the main body, which defines the QE-KLD-WRM step. This brings in further hyper-parameters: initial guess, \textit{``inner''} optimizer choice, and the number of inner optimization steps.

We use the Q-KLD-WRM step as the initial guess. This way, even if our inner optimizer makes little progress, we can still get a reasonable step.  We choose the inner optimizer to be SGD with no momentum. We let $\omega$ be the ratio between the inner optimizer learning rate and outer optimizer ``learning rate\footnote{Which is $1/\lambda=0.01$ here for all solvers.}''. Further, we let $N_{IS}$ and $N_{CAP}$ be the number of inner optimization steps, and the maximum number of \textit{``old''} networks saved\footnote{To approximate the $\mathbb D_{KL}$ term in equation (36) of the main body.} respectively. Note that when we have more than $N_{CAP}$ networks stored (including the current one), we delete the oldest one. 

We have to choose $N_{CAP}$ large enough s.t.\ we have a good approximation of the $\mathbb D_{KL}$ term in equation (36) of the main body, but small enough s.t.\ the memory does not overflow. A simple rule (which is the one we followed) is to pick $N_{IS}$ to be the smallest positive integer s.t.\ $\rho^{N_{IS}} < 0.1$. The over-flow problem might limit us from making this choice\footnote{In this case, we have to choose between reducing $rho$ or using a coarser approximation of $\mathbb D_KL$.} for larger nets if TPUs are not available. However, for our net, it was not a concern even on a standard CPU/GPU. We let $\rho$, $\omega$, $N_{IS}$ be variable, and try four different parameters configurations (see \textit{Figures 1,2,3} and \textit{4}). Note that $\rho$ is not specific to QE, but is present in all solvers.

Since the exact KL divergence estimate tended to be 2.5 orders o magnitude larger than the loss, we used $\zeta(i)=\kappa(i)/330$ for QE-KLD-WRM, rather than $\zeta(i) = \kappa(i)$ as for the other solvers.

\subsection{Implementation Detail: Network Architecture}

\textit{Conv1: seven $5 \times 5$ filters, ReLu activation, $2 \times 2$ MaxPool; Conv2: five $5 \times 5$ filters, ReLu activation, $2 \times 2$ MaxPool; Fully-Connected 1: 112
nodes and ReLu activation; Fully-Connected 2: 30 nodes}. The output dimension is equal to the number of classes, which is $n_c = 10$. This gives a total of $4712$ parameters. We also add dropout at the very last layer to reduce over-fit. The output of the net represents the logits, which are then passed through softmax activation function to obtain class probabilities. The loss is taken to be cross-entropy (using the softmax output), as is typical in classification.

Both the network architecture and the optimizer details can be directly observed from the codes available at: GIVE CODE LINK.

\subsection{Further Numerical Results and Comments}

In this section, we present numerical results for all investigated parameters (as opposed to only the best ones as in the main body). Even though we do not focus on the training loss, we also show it here for completeness.

\subsubsection{Considered Metrics Explanation}

We present results for K-FAC, SO-KLD-WRM, Q-KLD-WRM and QE-KLD-WRM in \textit{Figures 1, 2, 3} and \textit{4} respectively. Four different hyper-parameter settings are considered for each solver, and these vary across columns for each figure. \textit{Test accuracy}, \textit{test loss} and \textit{train loss} are considered going down the rows. Ten runs are considered for each \textit{[optimizer, parameter setting]} pair. To simplify the interpretation of these results, we consider extracting some \textit{summary statistics\footnote{Which are also performance metrics.}} out of the raw results (shown in \textit{Figures 1 - 4}). These summary statistics are shown in \textit{Table 1}.

Firstly, we consider the distribution of \textit{test accuracy} and \textit{test loss} after 50 training epochs. We show the \textit{empirical mean} and \textit{standard deviation} in the final columns of \textit{Table 1}. In principle, low standard deviation is desirable, because it indicates more robustness (w.r.t.\ randomness) in obtaining good results. However, having ``\textit{favourable'' outliers} (which are more likely to happen under large variance) is beneficial.

 Secondly, we count the \textit{number of runs that exceed a certain goodness threshold} for both \textit{test accuracy} (higher is better) and \textit{test loss} (lower is better). These metrics are meant to give an idea about \textit{``favourable''} \textit{outliers}, as well as about the robustness of achieving good results with a solver. There are 2 important points about the counting procedure:
\vspace{-3ex}
\begin{enumerate}
	\item \textit{Counting is done even if the metric degrades afterwards}. This is because we assume one would periodically save checkpoints - and choose the best at the end of training (this is easily implementable in the code at the expense of some communication cost and disk storage cost);
	\item \textit{Counting is not done unless the metric} either exceeds the threshold significantly once, or does so multiple times. We do this to try to reduce the noise.
\end{enumerate}

\subsubsection{Interpretation of Results}

\textbf{Test Accuracy:} For K-FAC, mean accuracy is around $96\%$, whereas for our proposed solvers it gets as high as $97.5\%$ for all of them, with some getting close to, and even exceeding, $98\%$. The standard deviation is also much smaller for our proposed algorithms (desirable). This can also be observed by looking at \textit{Figures 1-4}, and counting the number of runs that exceed a certain threshold (avoid counting a line if it appears to be crossed only due to noise). These counting results are also summarized in \textit{Table 1}. 

We see that, while K-FAC typically has one outlier with more than $98.5\%$ test accuracy, it only has at most 4 runs where the test accuracy is greater than or equal to $98\%$, and 3 runs where this is strictly greater than $98\%$. In contrast, our proposed KLD-WRM instantiations have at least 6 runs where the test accuracy is greater than or equal to $98\%$, and at least 4 runs where this is strictly greater than $98\%$ for most hyper-parameter values. In fact, by observing columns 2 and 3 in \textit{Table 1}, we see that for some hyper-parameter values, our proposed solvers can get as many as 7-8 runs where the test accuracy is greater than or equal to $98\%$, and 5 runs where this is strictly greater than $98\%$. Furthermore, even in terms of favourable outliers, we have a KLD-WRM variant that performs on par with K-FAC: QE-KLD-WRM (see column 4 in \textit{Table 1}).

\textbf{Test Loss:} Test loss results are roughly analogous to the test accuracy results. There are three main points to note here. First, there are 1-2 K-FAC \textit{``favourable'' outliers} (test loss better than 0.2) rather than just 1, as was the case for test accuracy. Second, there are no such outliers for KLD-WRM variants (not even for QE, as was the case with test accuracy results). Third, very good test loss is not always equivalent to very good test accuracy. This observation applies to all four solvers, but it seems that the equivalence is slightly more fragile for KLD-WRM variants. The meaning and reasons behind this are outside our scope and touch ideas like the fact that \textit{the test set is just a sample from the distribution we care about}, and \textit{comparing accuracy versus robustness to adversarial attacks}.

\textbf{Robustness and Overall Performance:} Overall, our results show that if we can only afford to run the training a few times (perhaps once), it is preferable (in terms of epochs; both from a \textit{test accuracy} and \textit{test-loss} point of view) to use one of our KLD-WRM variants rather than K-FAC, because these more robustly\footnote{Better mean and lower variance.} achieve good results. Conversely, if we can afford to run the algorithms many times (and take the best solution), the high variance of K-FAC results metrics \textit{may eventually} play to our advantage (but this is not guaranteed). Thus, we believe that based on the presented results, it is preferable (in terms of epochs) to use a KLD-WRM variant rather than using K-FAC.

\textbf{Choosing an Appropriate KLD-WRM Variant and Winners: }Which variant of KLD-WRM should we choose depends on the problem. Recall that SO-KLD-WRM and Q-KLD-WRM have virtually the same computational\footnote{Includes both data acquisition cost and optimizer cost (optimizer cost includes linear algebra and oracle cost - Forward pass and automatic differentiation cost).} cost per step as K-FAC. Conversely, QE-KLD-WRM has the same data acquisition and linear-algebra\footnote{The linear algebra cost is marginally larger for QE-KLD-WRM than for K-FAC, but this has no real influence on the observed wall time (as with the SO and Q variants).} costs as K-FAC, but higher oracle\footnote{Here we use oracle cost to refer to the forward pass and backward pass computation cost.} cost by a factor of $N_{IS} + 1$, due to the SGD steps taken to (approximately) solve the QE-KLD-WRM optimization subproblem (which defines the QE step).

Thus, for \textit{problems where the data acquisition cost is relatively low\footnote{Eg.\ classification problems with data on our disk.}}, one should choose between Q-KLD-WRM and SO-KLD-WRM, as these would (most of the time) give the best results with virtually identical wall-time as K-FAC. The best picks are SO-KLD-WRM with $\rho = 0.33$ and Q-KLD-WRM with $\rho \in \{0.33,0.5\}$. The latter performs slightly better, but the former gives almost the same performance for a negligible implementation effort (assuming K-FAC codes are available) and might thus be preferred for prototyping.

For \textit{problems where the data acquisition cost is sufficiently high\footnote{Eg.\ reinforcement learning with high data simulation cost.} (dominating the linear algebra and oracle cost)}, the wall time per epoch will be virtually the same across all four solvers. In this case, one should choose QE-KLD-WRM. The best parameters we found (on MNIST classification) were $\rho = 0.5$,  $N_{IS} = 10$, $\omega = 0.07$, $N_{\text{cap}} = 4$.

\textbf{Best Hyper-parameter Settings for each Solver:} By investigating \textit{Figures 1, 2, 3, 4,} and \textit{Table 1}, one would (arguably) choose the following hyper-paratemers settings to be the best: \textit{(1) K-FAC}: $\rho = 0.95$ (which is what also the original author recommends), \textit{(2) SO-KLD-WRM}: $\rho = 0.33$, \textit{(3) Q-KLD-WRM}: $\rho = 0.33$ or $\rho = 0.5$, \textit{(4) QE-KLD-WRM:} $\rho = 0.5$,  $N_{IS} = 10$, $\omega = 0.07$, $N_{\text{cap}} = 4$. These are the values we used for comparison in the main body.

\newpage
\begin{figure*}[hp]
	\centering
	\includegraphics[trim={0.07cm 0.25cm 1.8cm 0.85cm},clip,width=0.24\textwidth]{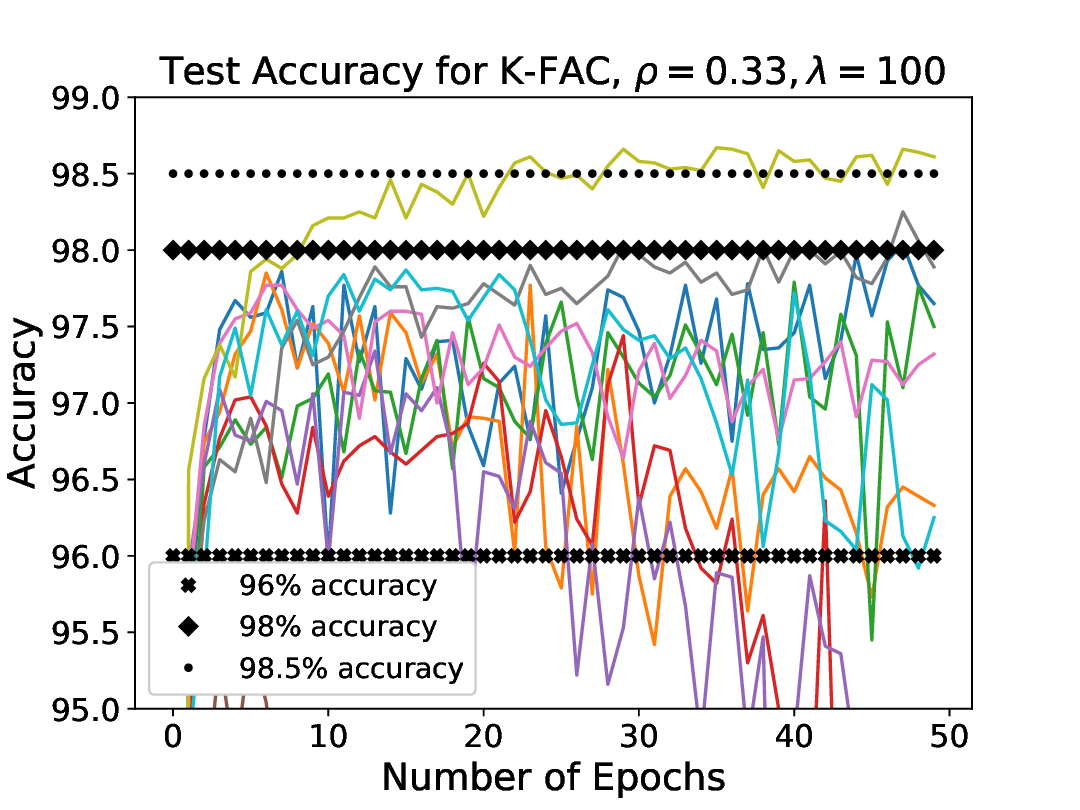}
	\includegraphics[trim={0.07cm 0.25cm 1.8cm 0.85cm},clip,width=0.24\textwidth]{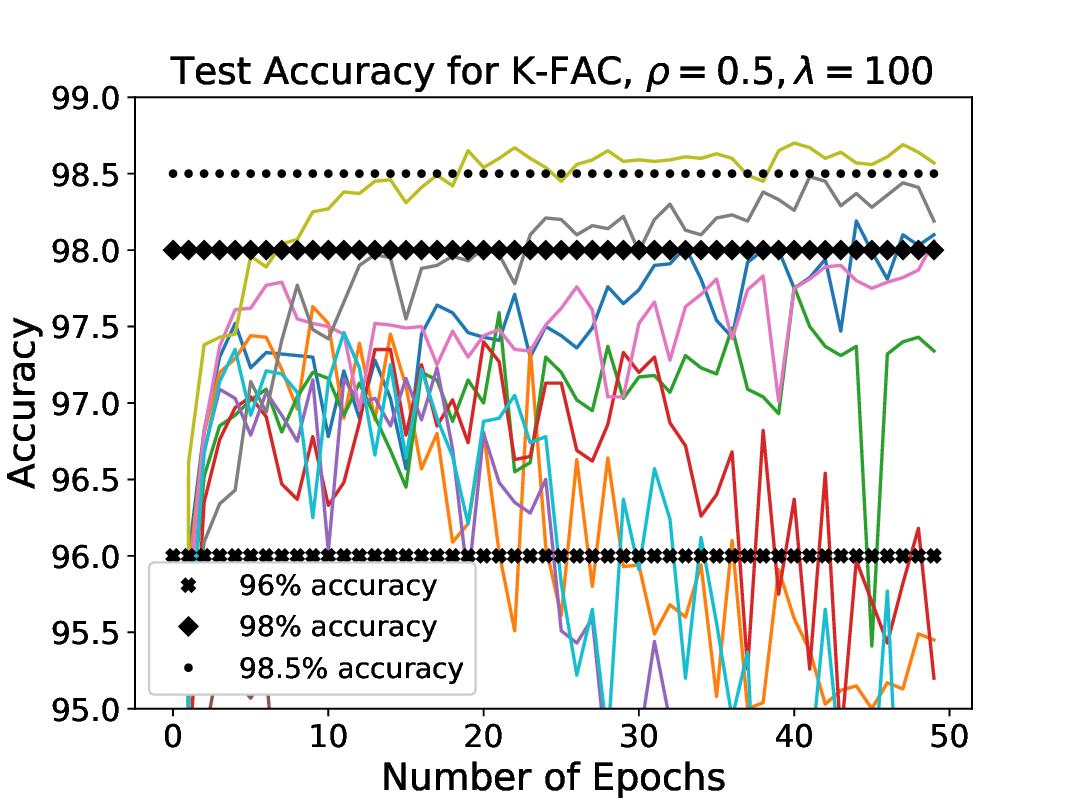}
	\includegraphics[trim={0.07cm 0.25cm 1.8cm 0.85cm},clip,width=0.24\textwidth]{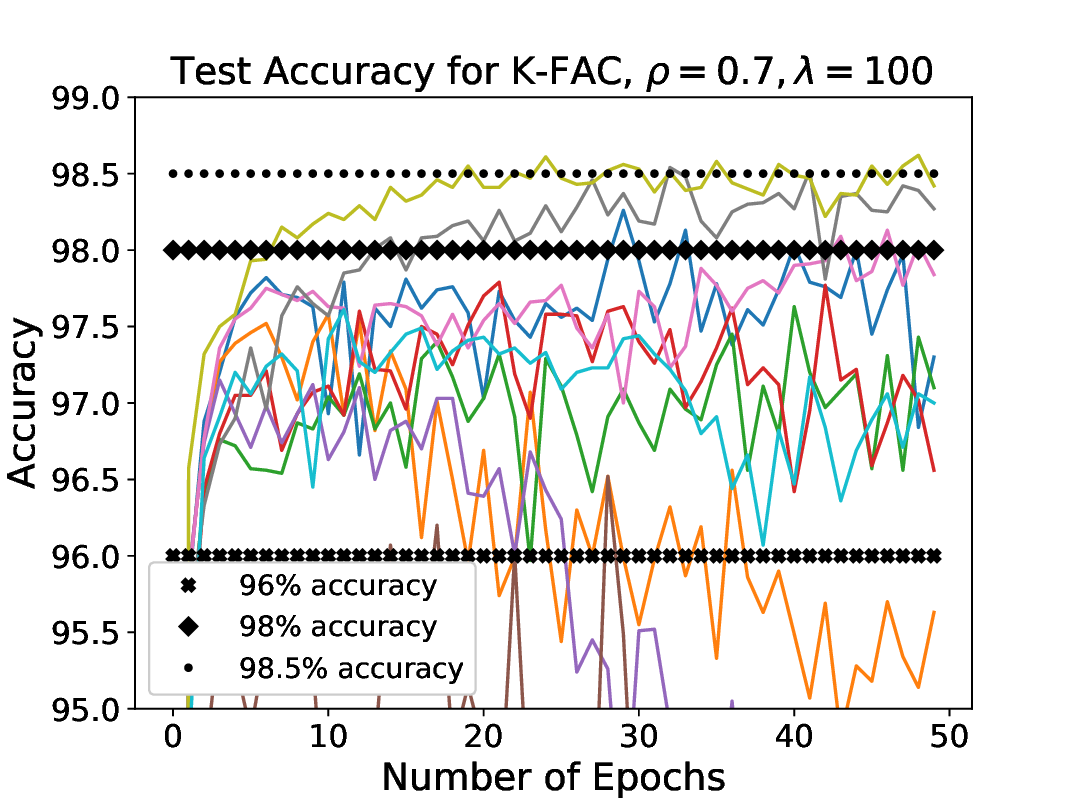}
	\includegraphics[trim={0.07cm 0.25cm 1.8cm 0.85cm},clip,width=0.24\textwidth]{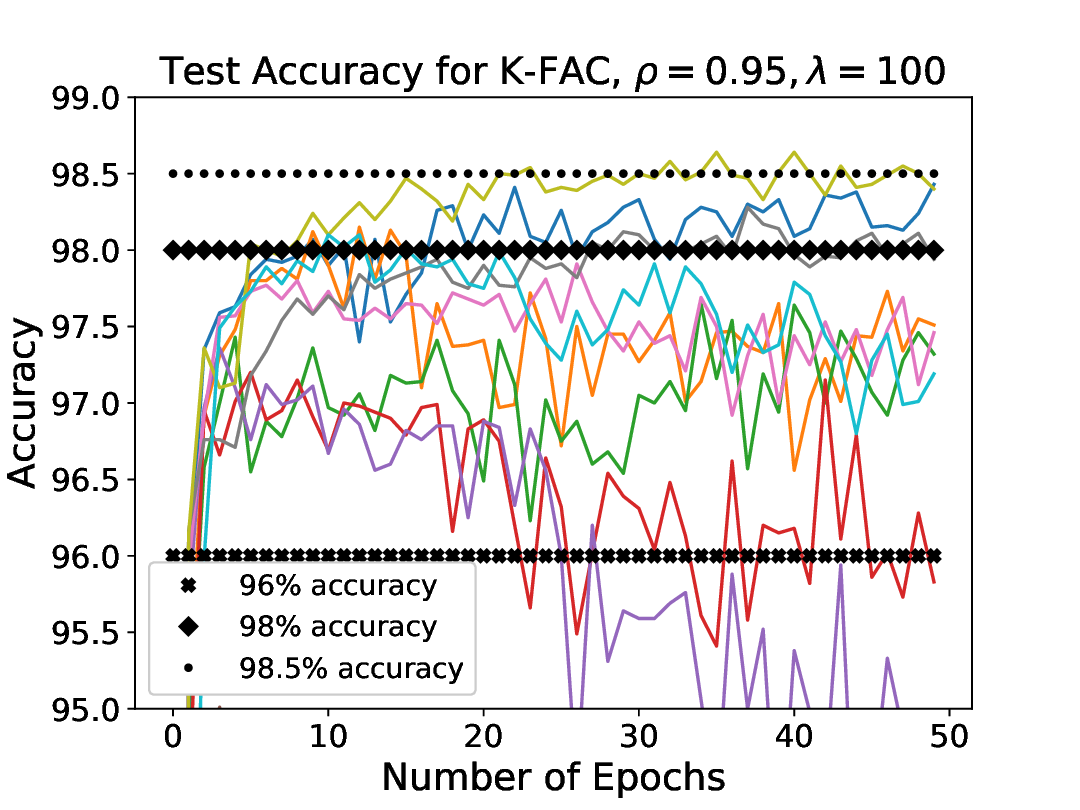}
	
	\vspace{+0.5ex}
	\includegraphics[trim={0.07cm 0.25cm 1.8cm 0.85cm},clip,width=0.24\textwidth]{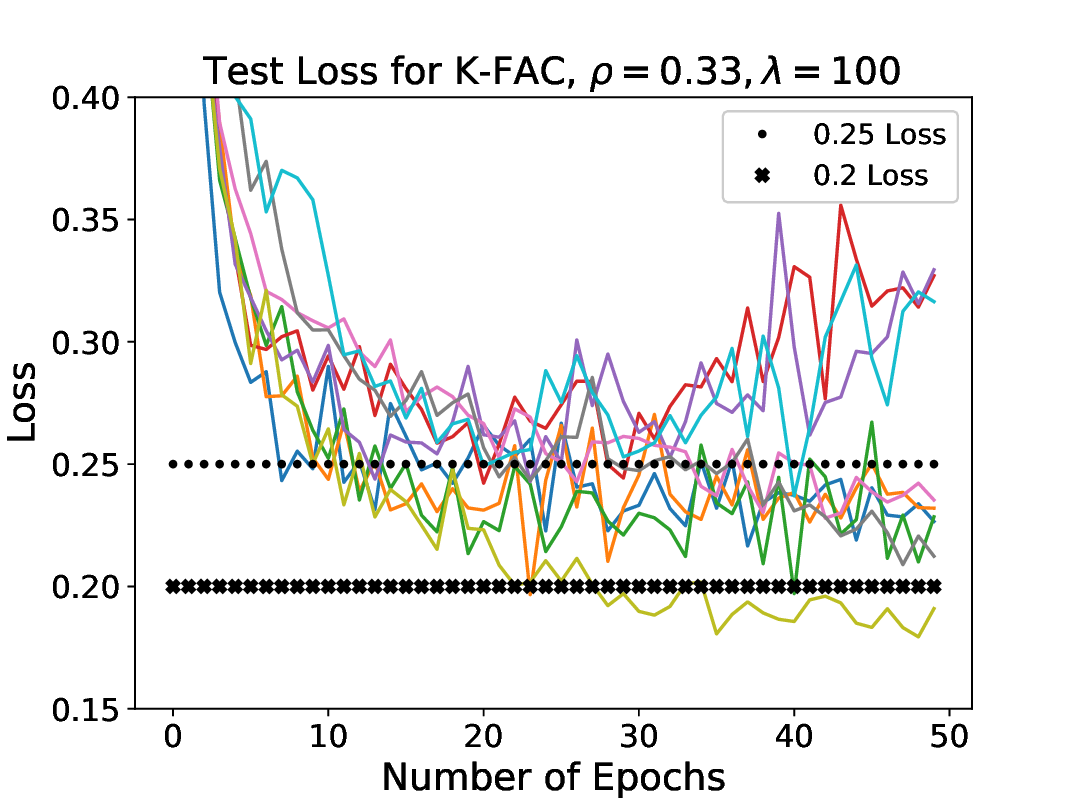}
	\includegraphics[trim={0.07cm 0.25cm 1.8cm 0.85cm},clip,width=0.24\textwidth]{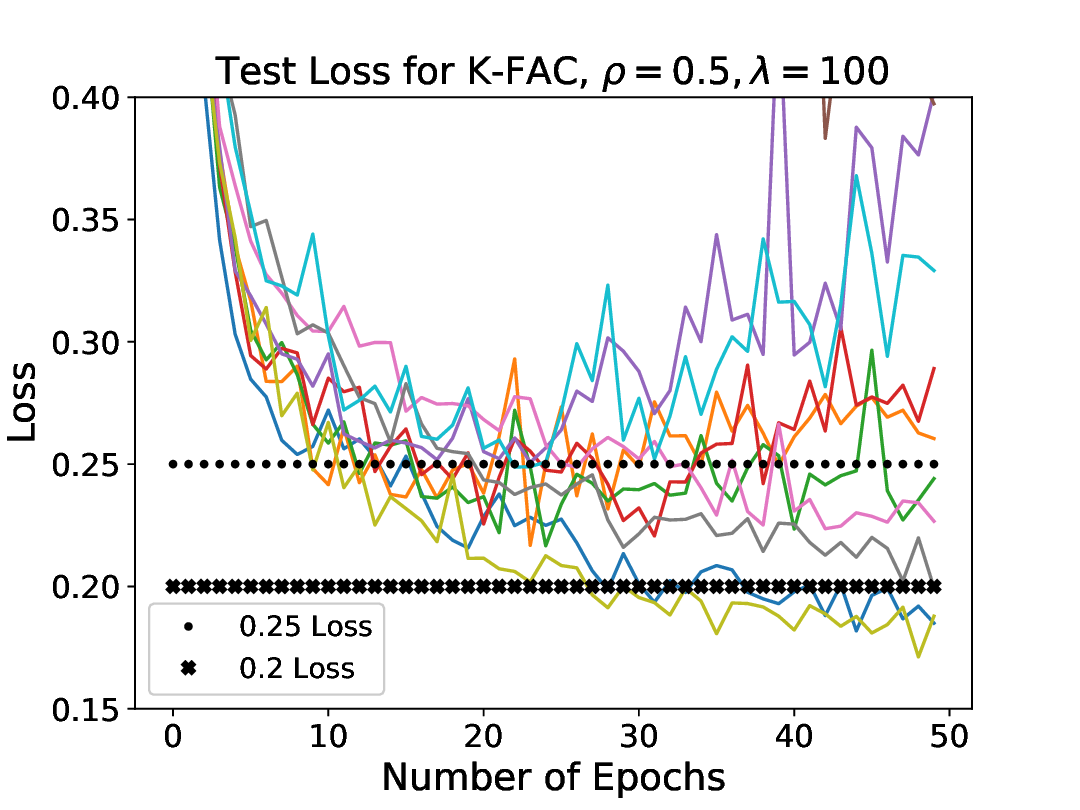}
	\includegraphics[trim={0.07cm 0.25cm 1.8cm 0.85cm},clip,width=0.24\textwidth]{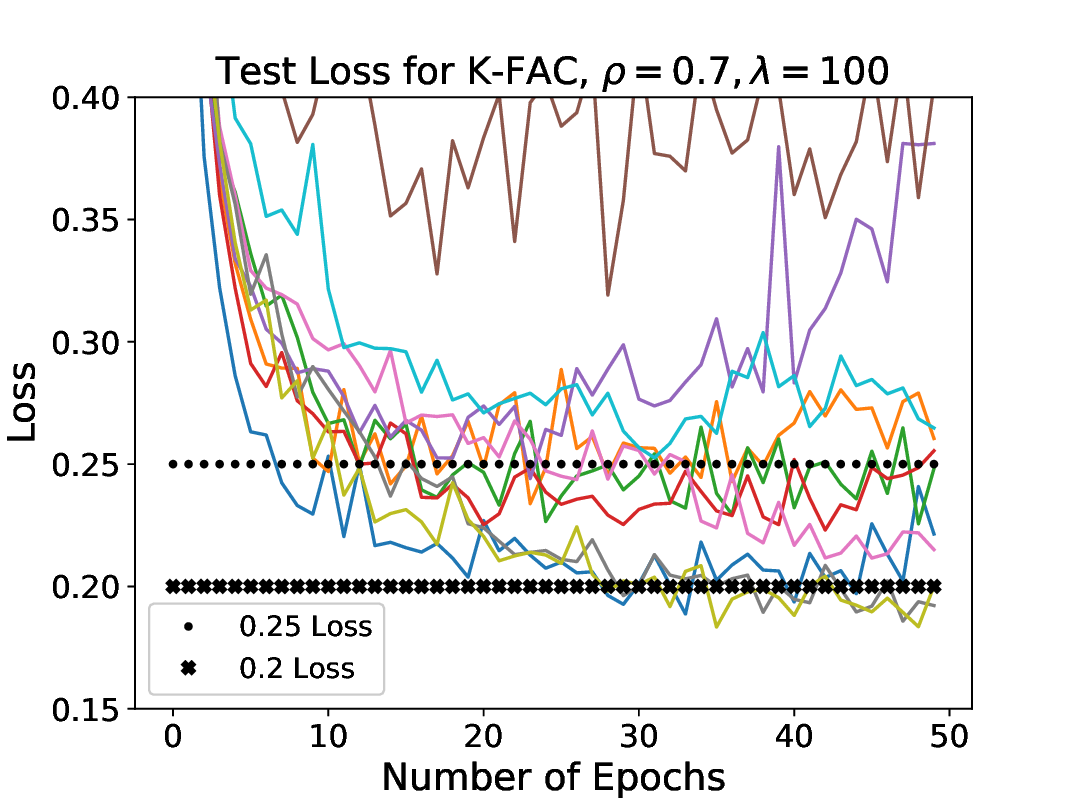}
	\includegraphics[trim={0.07cm 0.25cm 1.8cm 0.85cm},clip,width=0.24\textwidth]{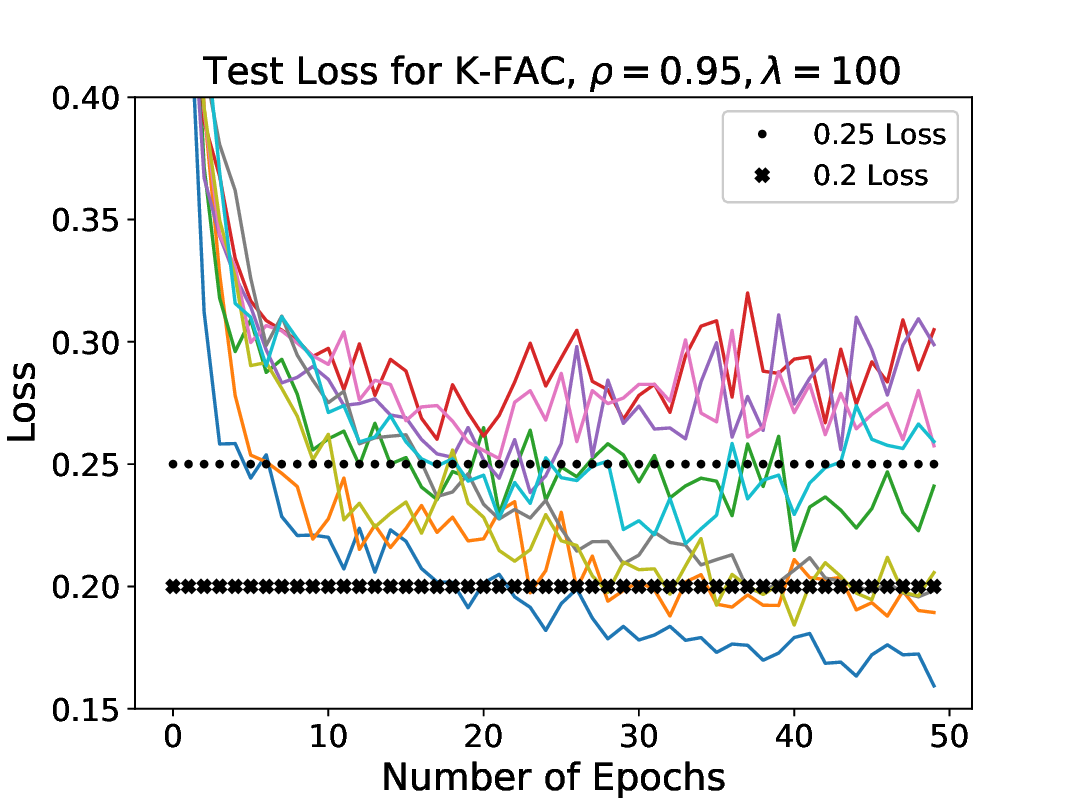}
	
	\vspace{+0.5ex}
	\includegraphics[trim={0.07cm 0.25cm 1.8cm 0.85cm},clip,width=0.24\textwidth]{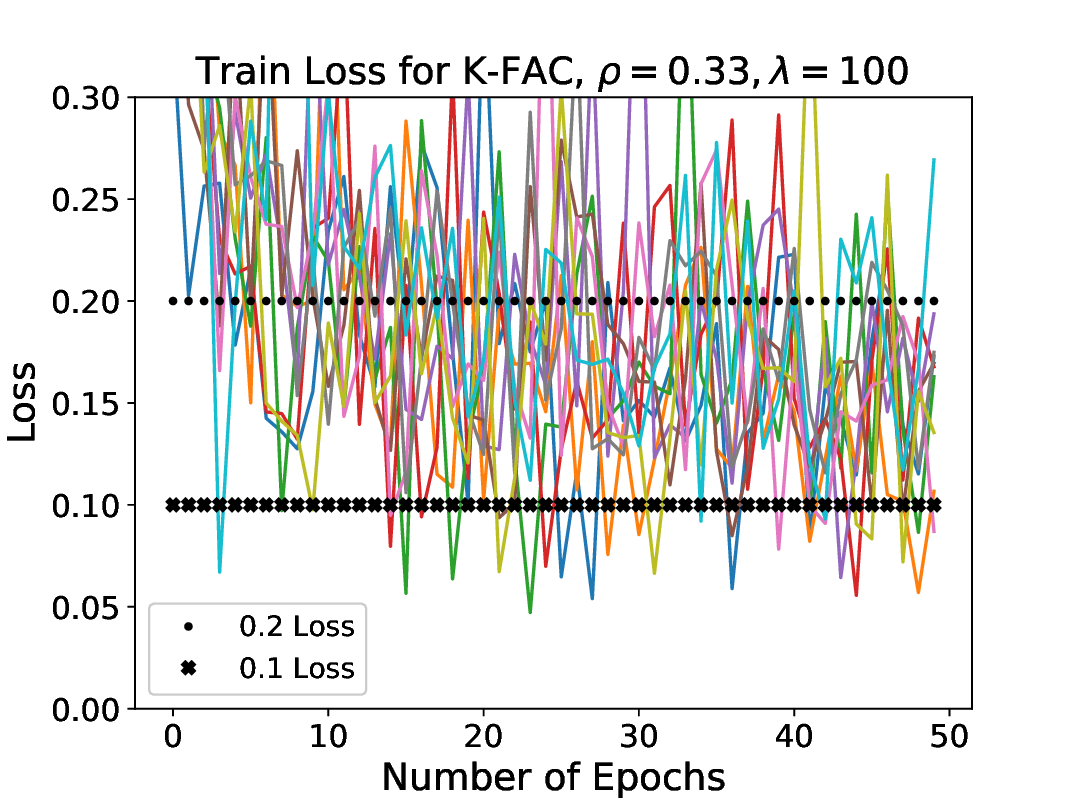}
	\includegraphics[trim={0.07cm 0.25cm 1.8cm 0.85cm},clip,width=0.24\textwidth]{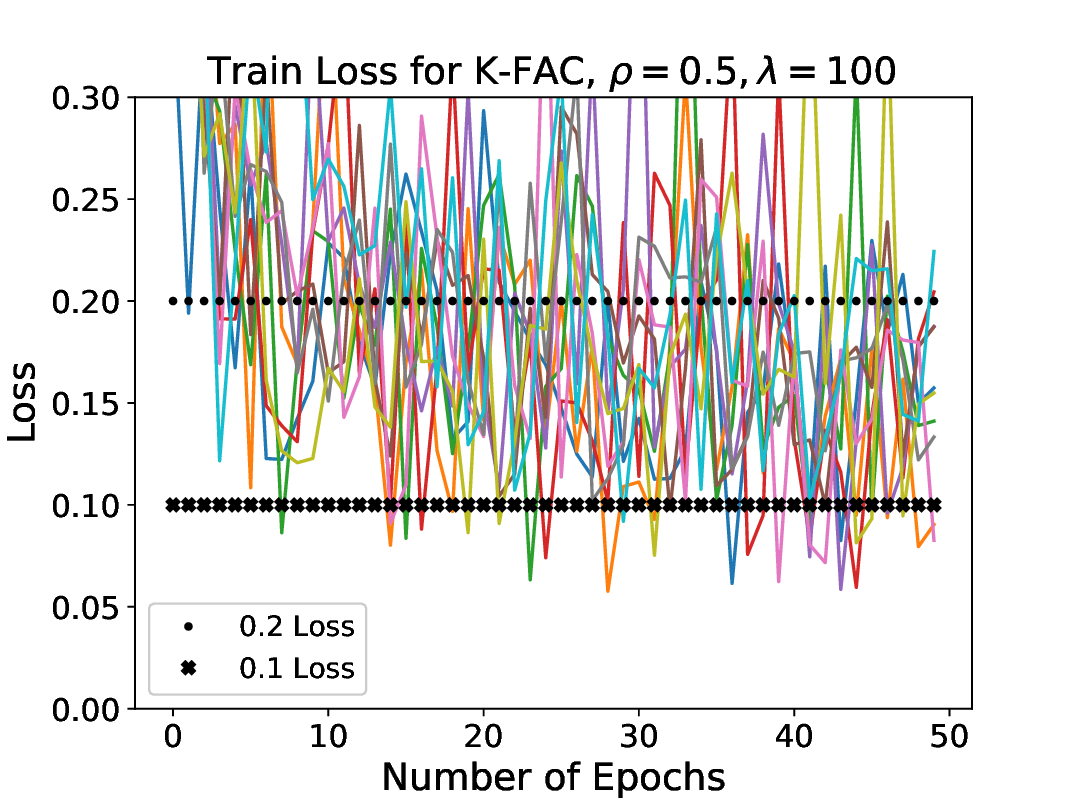}
	\includegraphics[trim={0.07cm 0.25cm 1.8cm 0.85cm},clip,width=0.24\textwidth]{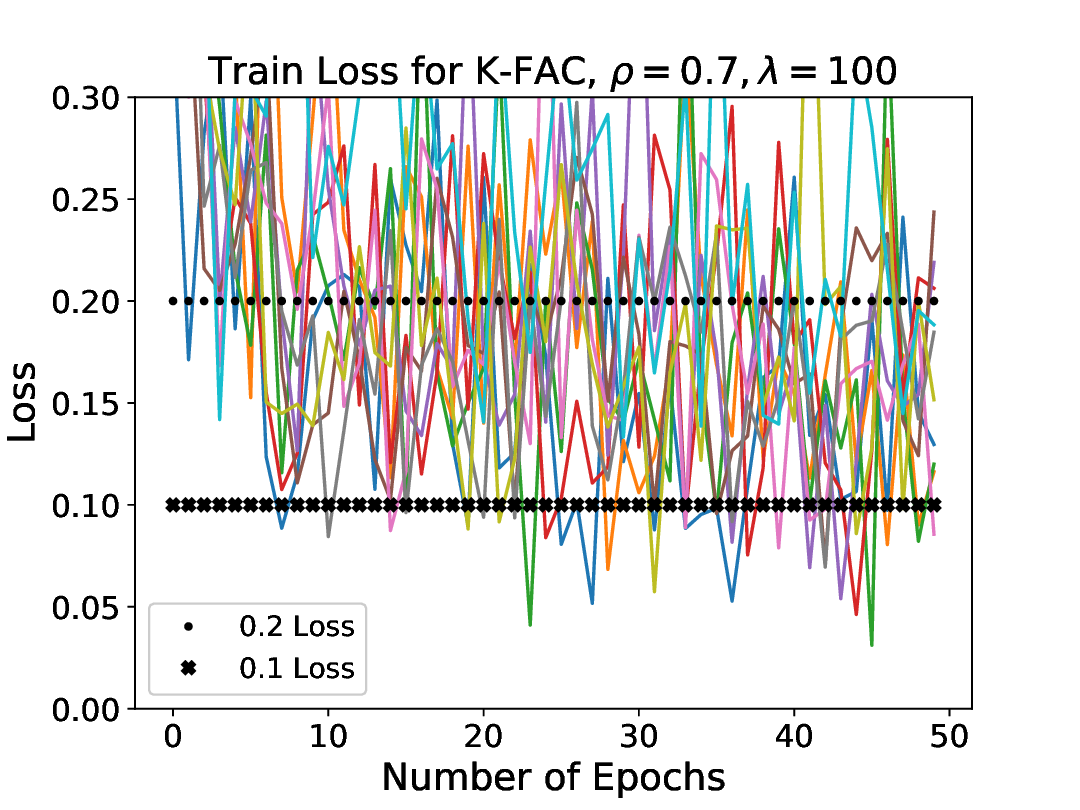}
	\includegraphics[trim={0.07cm 0.25cm 1.8cm 0.85cm},clip,width=0.24\textwidth]{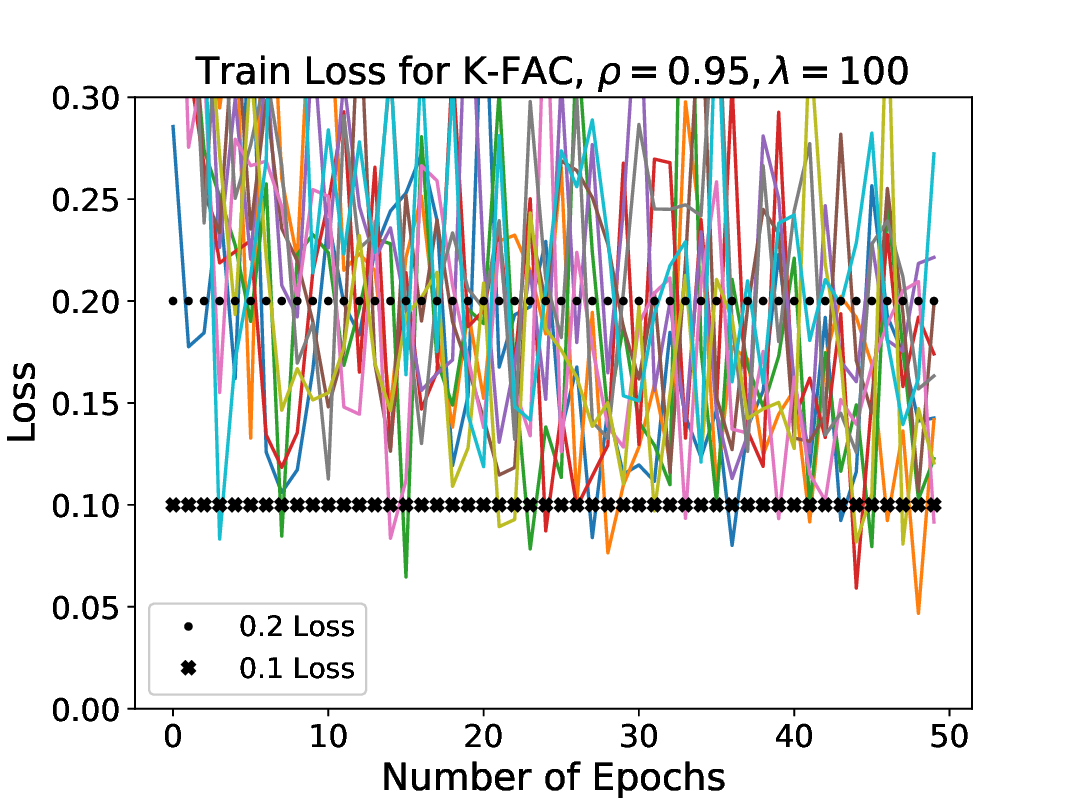}
	\vspace{-1.ex}
	\caption{\textbf{K-FAC Results (MNIST Classification)}: from top to bottom: Test Accuracy, Test Loss, Training Loss.From left to right: different $\rho$ ($0.33$, $0.5$, $0.7$, $0.95$). The unspecified hyper-parameters were the same across all runs, and stated in \textit{Section 7.1}.  }
\end{figure*}
\begin{figure*}[hp]
	\centering
	\includegraphics[trim={0.07cm 0.25cm 1.19cm 0.85cm},clip,width=0.24\textwidth]{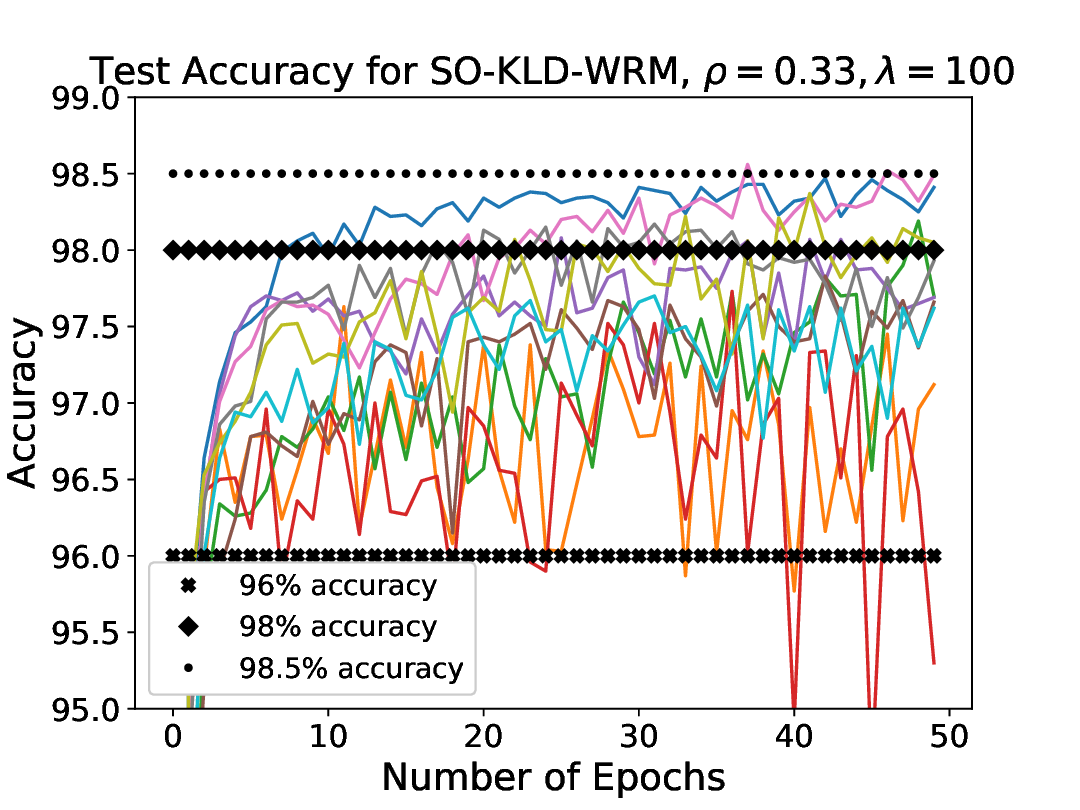}
	\includegraphics[trim={0.07cm 0.25cm 1.19cm 0.85cm},clip,width=0.24\textwidth]{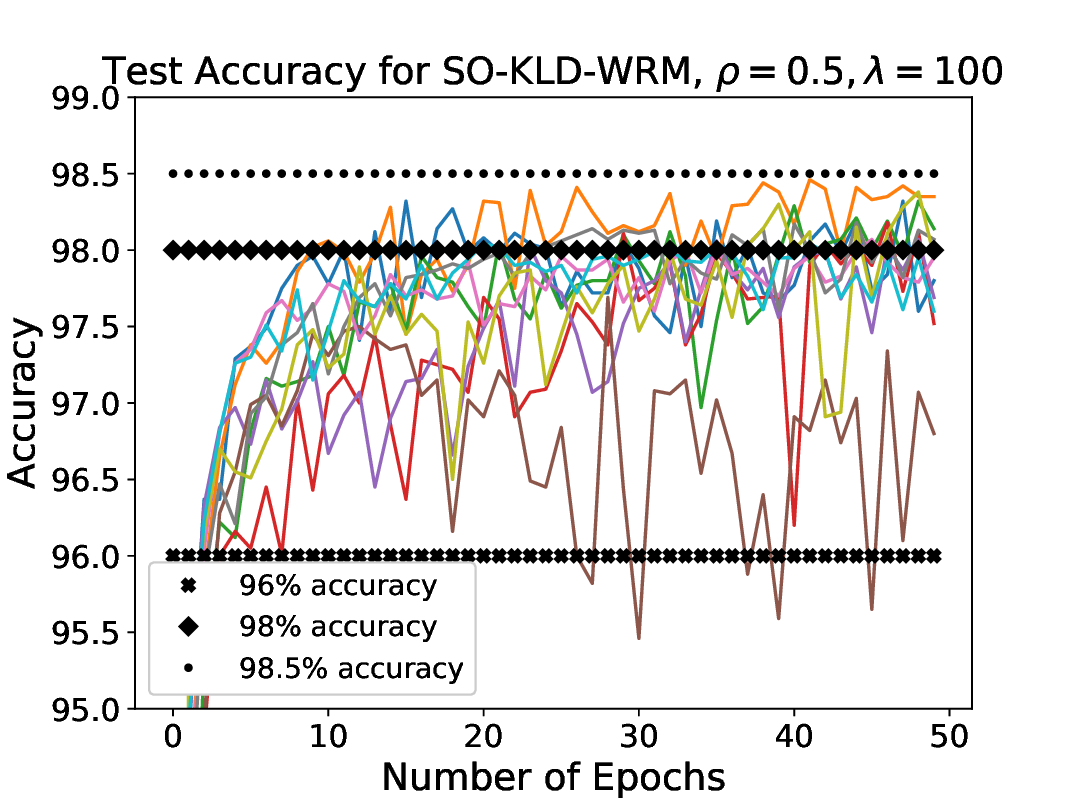}
	\includegraphics[trim={0.07cm 0.25cm 1.19cm 0.85cm},clip,width=0.24\textwidth]{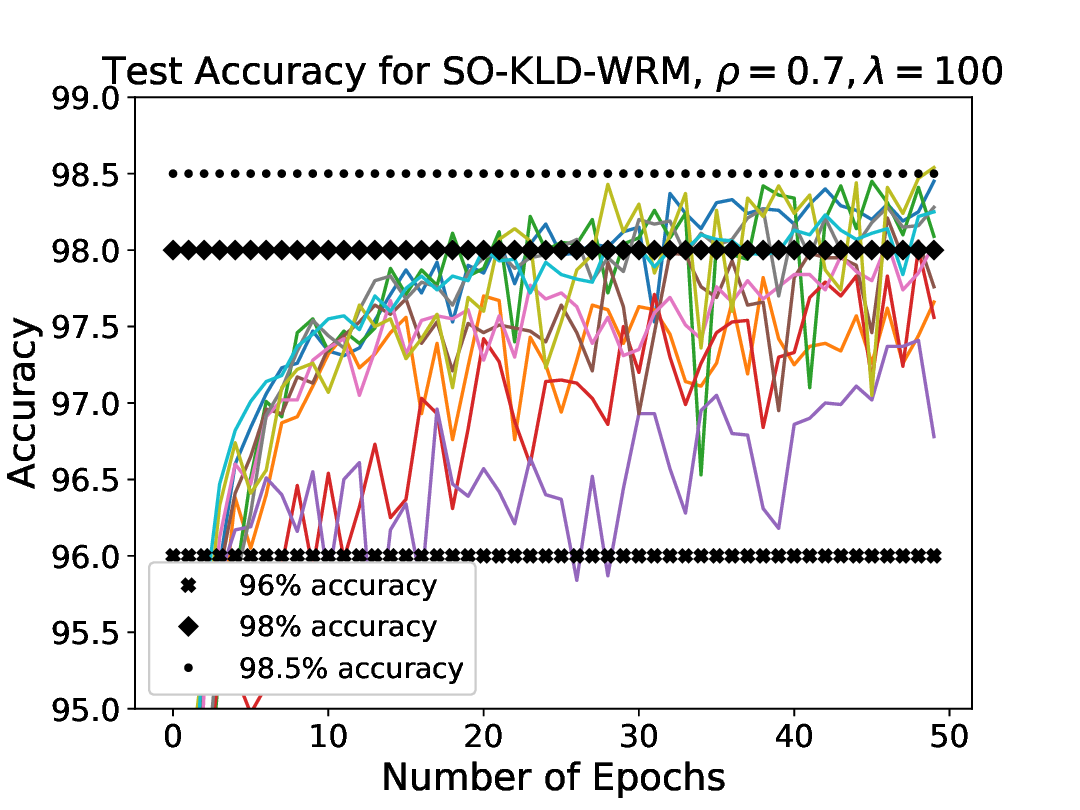}
	\includegraphics[trim={0.07cm 0.25cm 1.19cm 0.85cm},clip,width=0.24\textwidth]{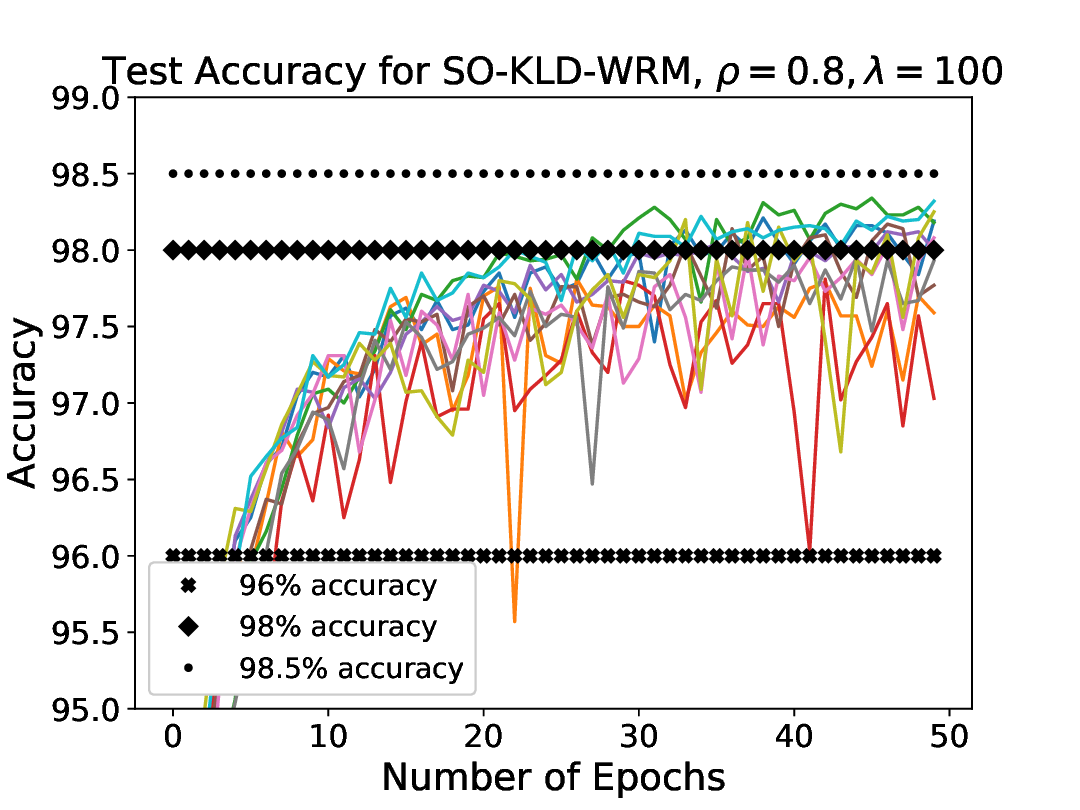}
	
	\vspace{+0.5ex}
	\includegraphics[trim={0.07cm 0.25cm 1.19cm 0.85cm},clip,width=0.24\textwidth]{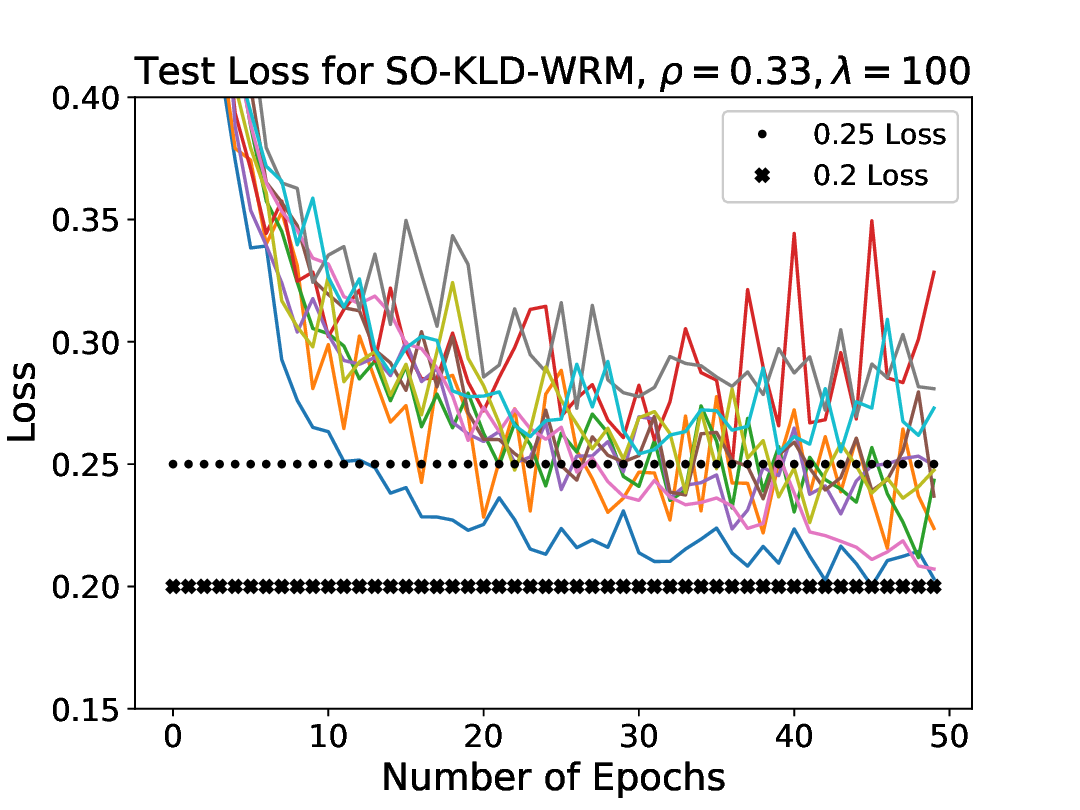}
	\includegraphics[trim={0.07cm 0.25cm 1.19cm 0.85cm},clip,width=0.24\textwidth]{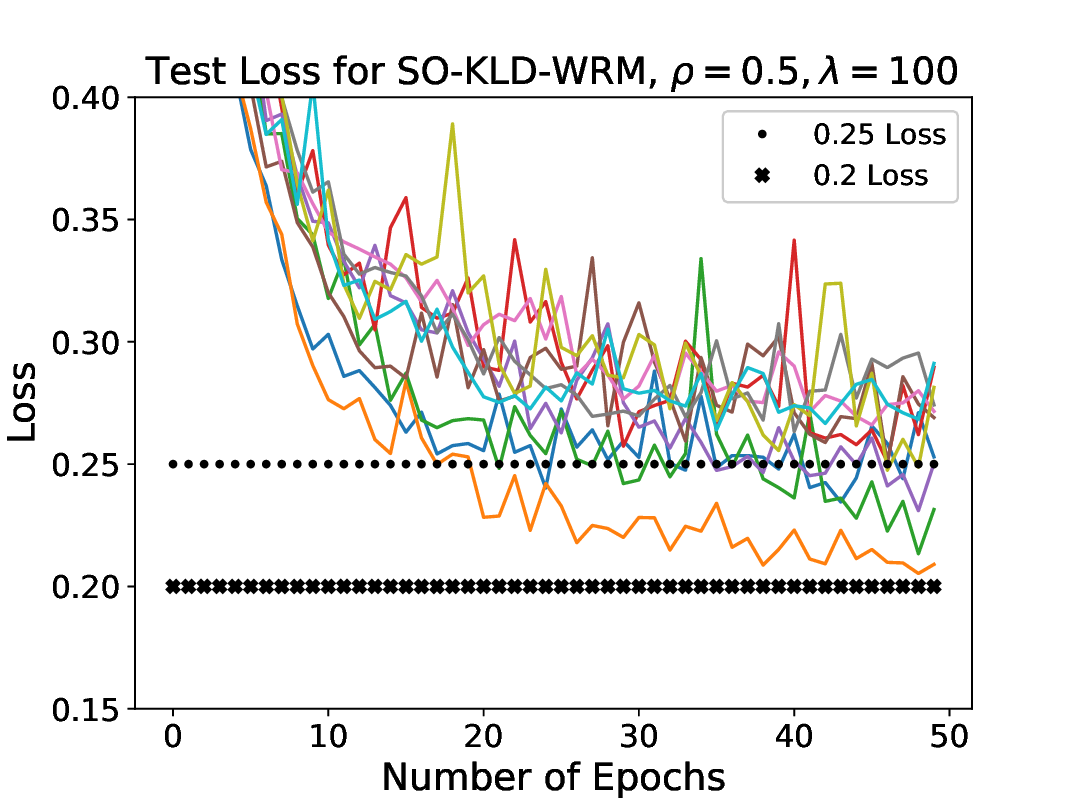}
	\includegraphics[trim={0.07cm 0.25cm 1.19cm 0.85cm},clip,width=0.24\textwidth]{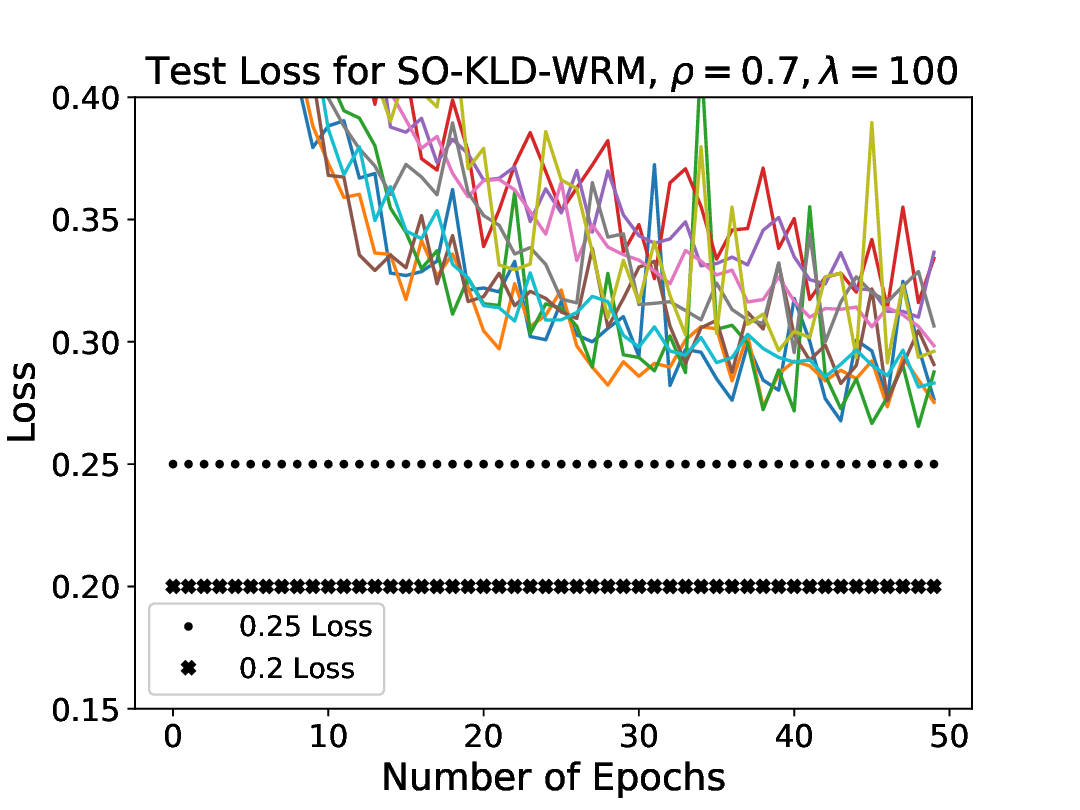}
	\includegraphics[trim={0.07cm 0.25cm 1.19cm 0.85cm},clip,width=0.24\textwidth]{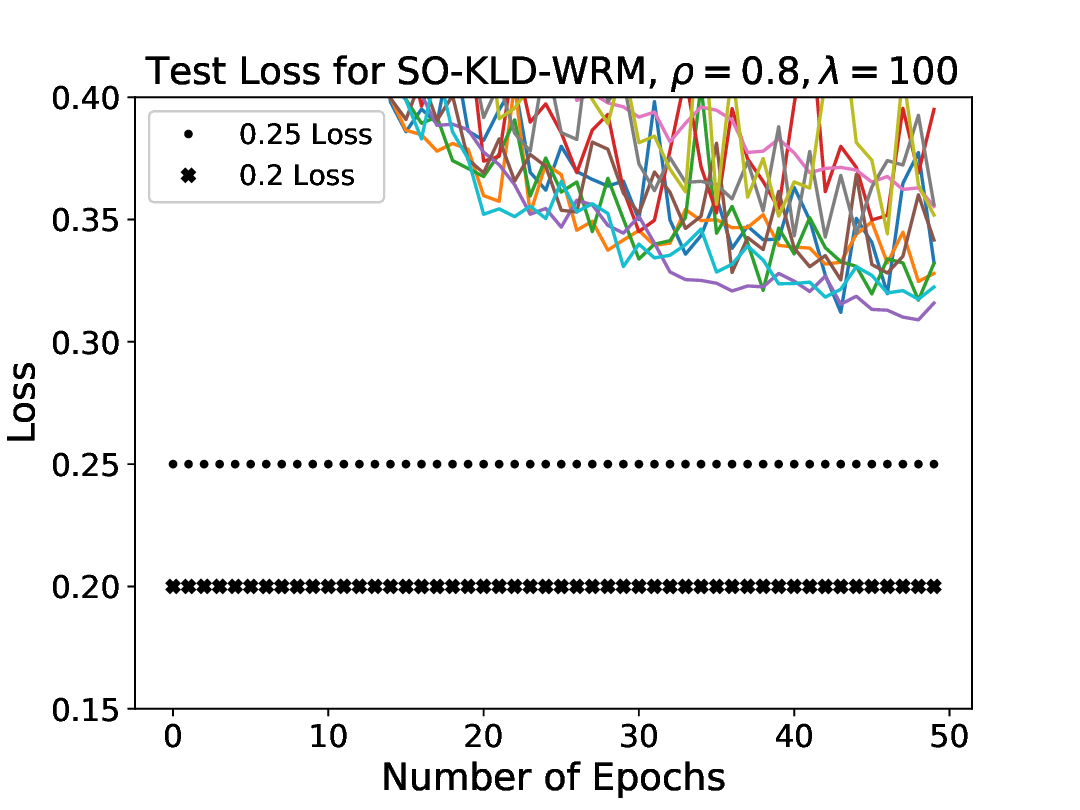}
	
	\vspace{+0.5ex}
	\includegraphics[trim={0.07cm 0.25cm 1.19cm 0.85cm},clip,width=0.24\textwidth]{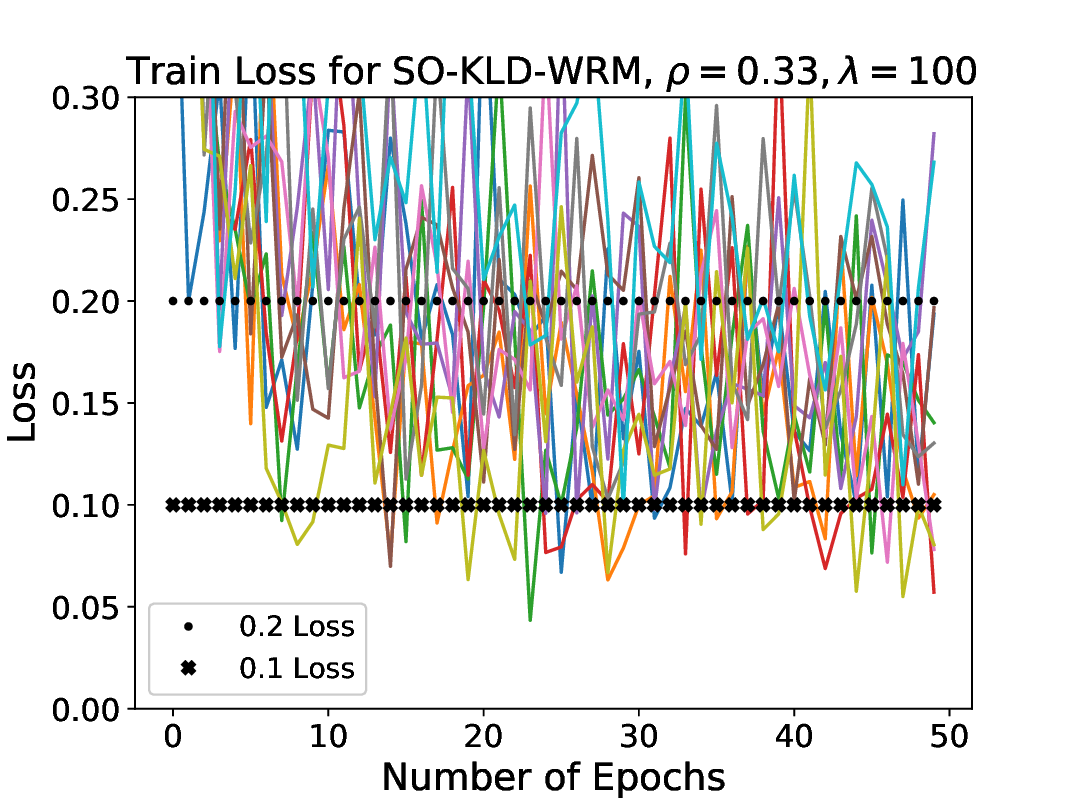}
	\includegraphics[trim={0.07cm 0.25cm 1.19cm 0.85cm},clip,width=0.24\textwidth]{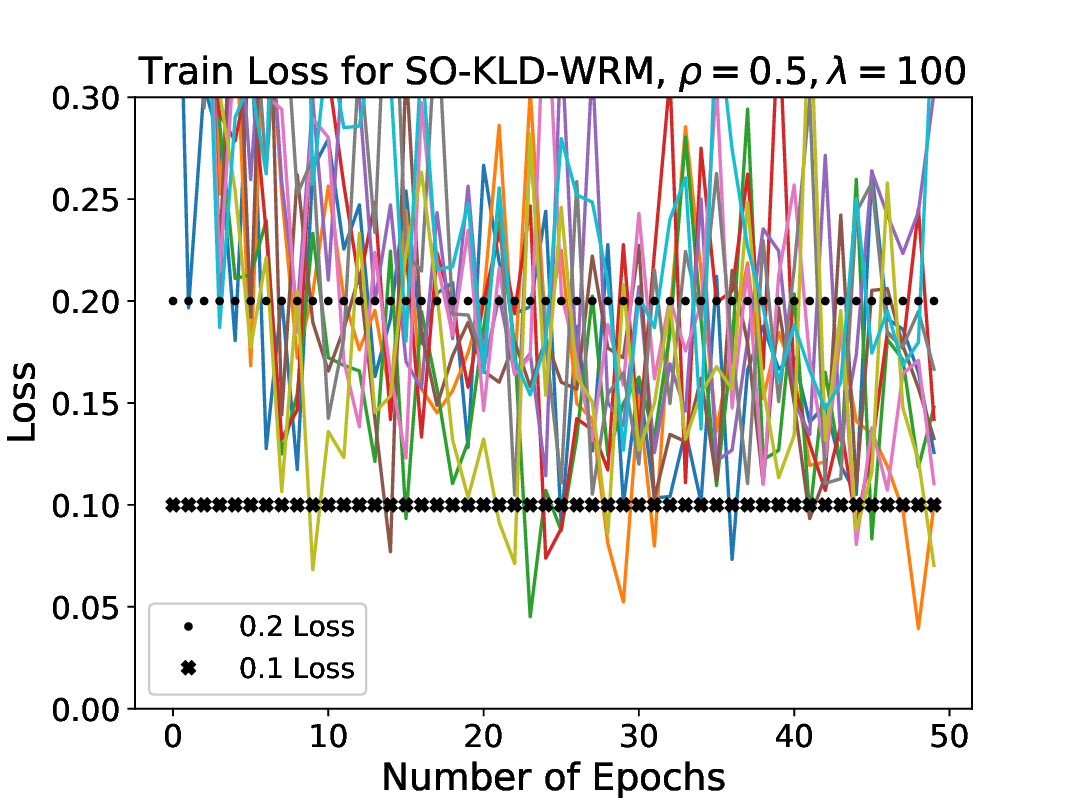}
	\includegraphics[trim={0.07cm 0.25cm 1.19cm 0.85cm},clip,width=0.24\textwidth]{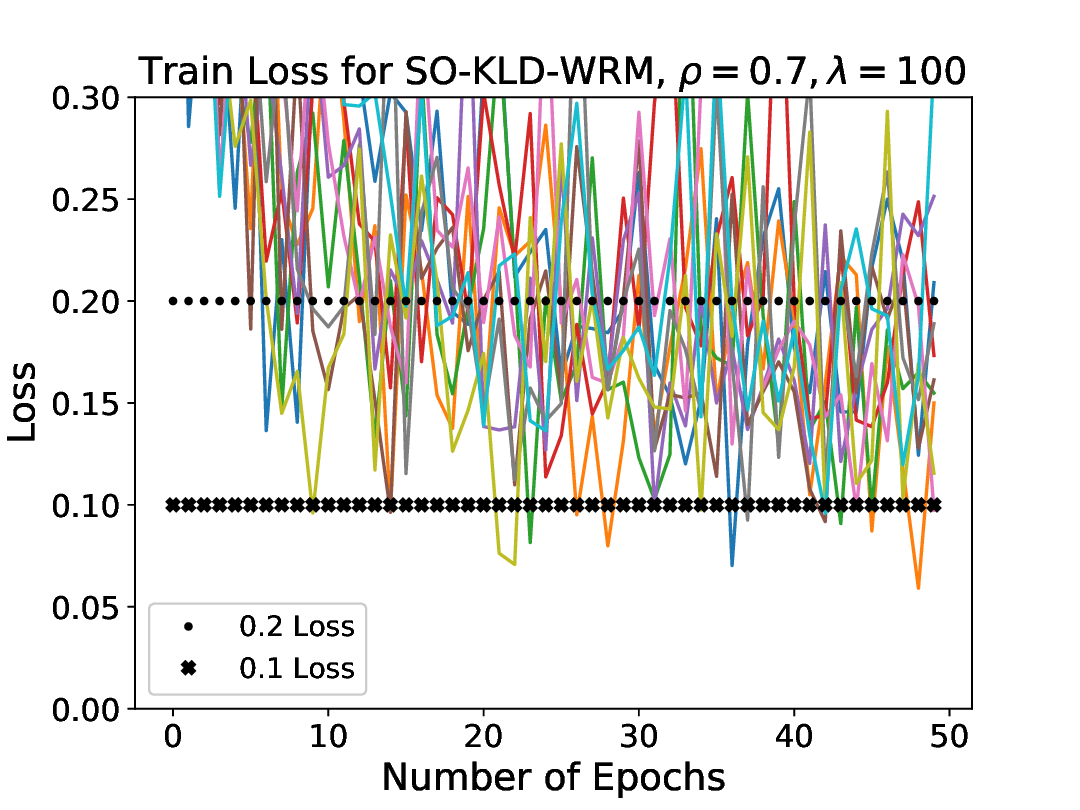}
	\includegraphics[trim={0.07cm 0.25cm 1.19cm 0.85cm},clip,width=0.24\textwidth]{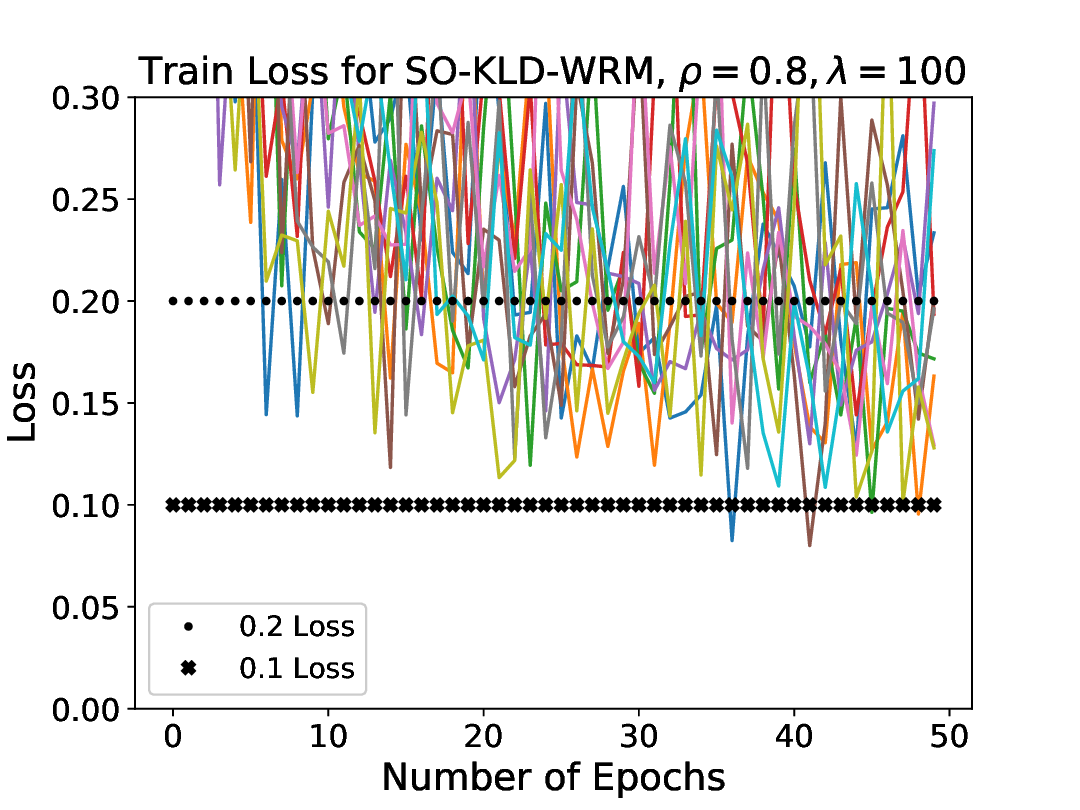}
	\vspace{-1.ex}
	\caption{\textbf{SO-KLD-WRM Results (MNIST Classification)}: from top to bottom: Test Accuracy, Test Loss, Training Loss. From left to right: different $\rho$ ($0.33$, $0.5$, $0.7$, $0.8$). The unspecified hyper-parameters were the same across all runs, and stated in \textit{Section 7.1}. }
\end{figure*}
\newpage
\begin{figure*}[hp]
	\centering
	\includegraphics[trim={0.07cm 0.25cm 1.35cm 0.85cm},clip,width=0.24\textwidth]{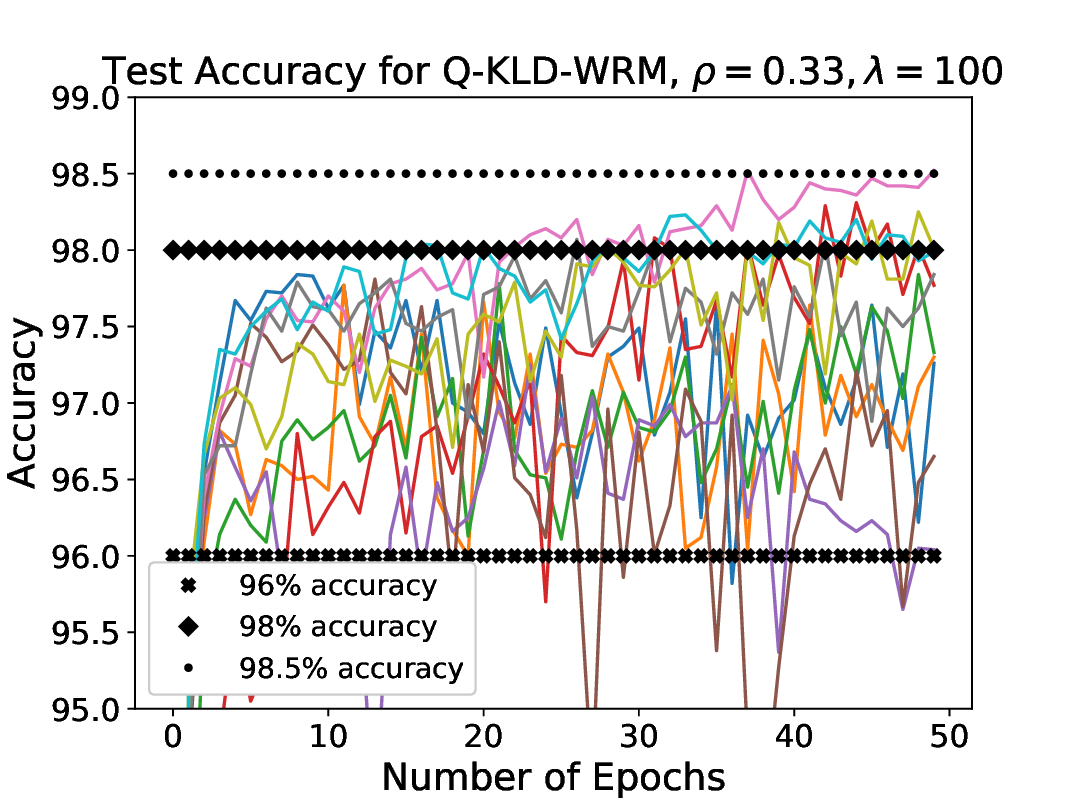}
	\includegraphics[trim={0.07cm 0.25cm 1.35cm 0.85cm},clip,width=0.24\textwidth]{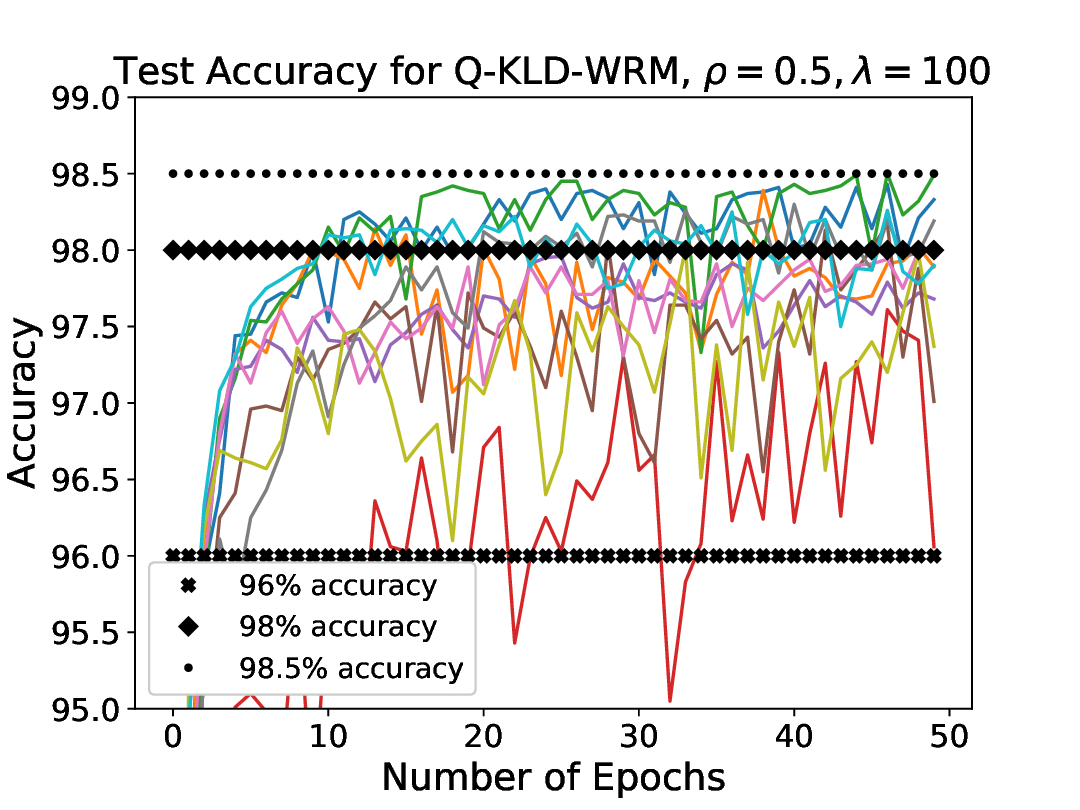}
	\includegraphics[trim={0.07cm 0.25cm 1.35cm 0.85cm},clip,width=0.24\textwidth]{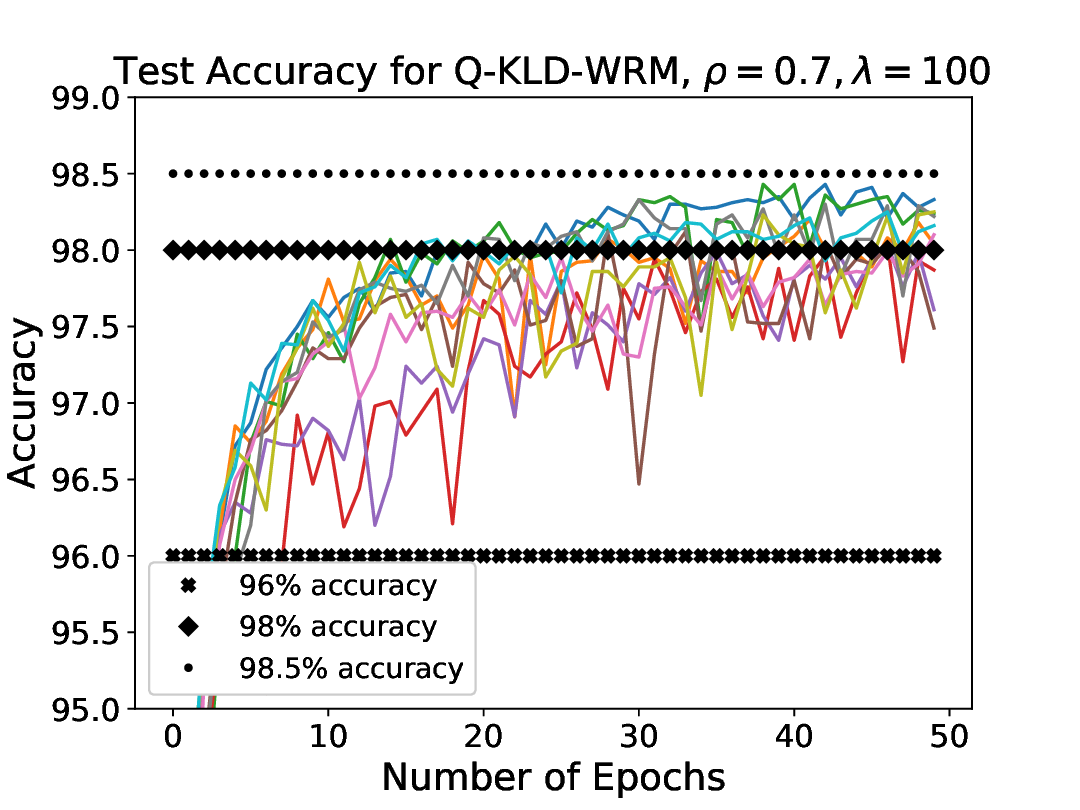}
	\includegraphics[trim={0.07cm 0.25cm 1.35cm 0.85cm},clip,width=0.24\textwidth]{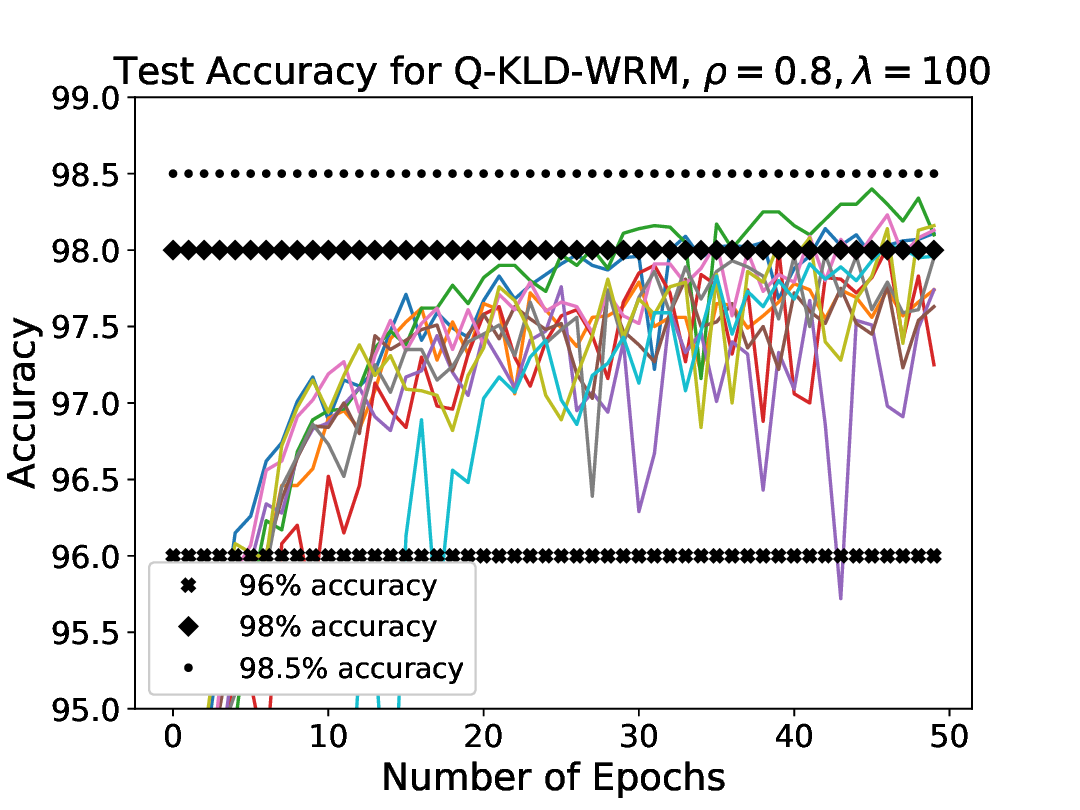}
	
	\vspace{+0.5ex}
	\includegraphics[trim={0.07cm 0.25cm 1.35cm 0.85cm},clip,width=0.24\textwidth]{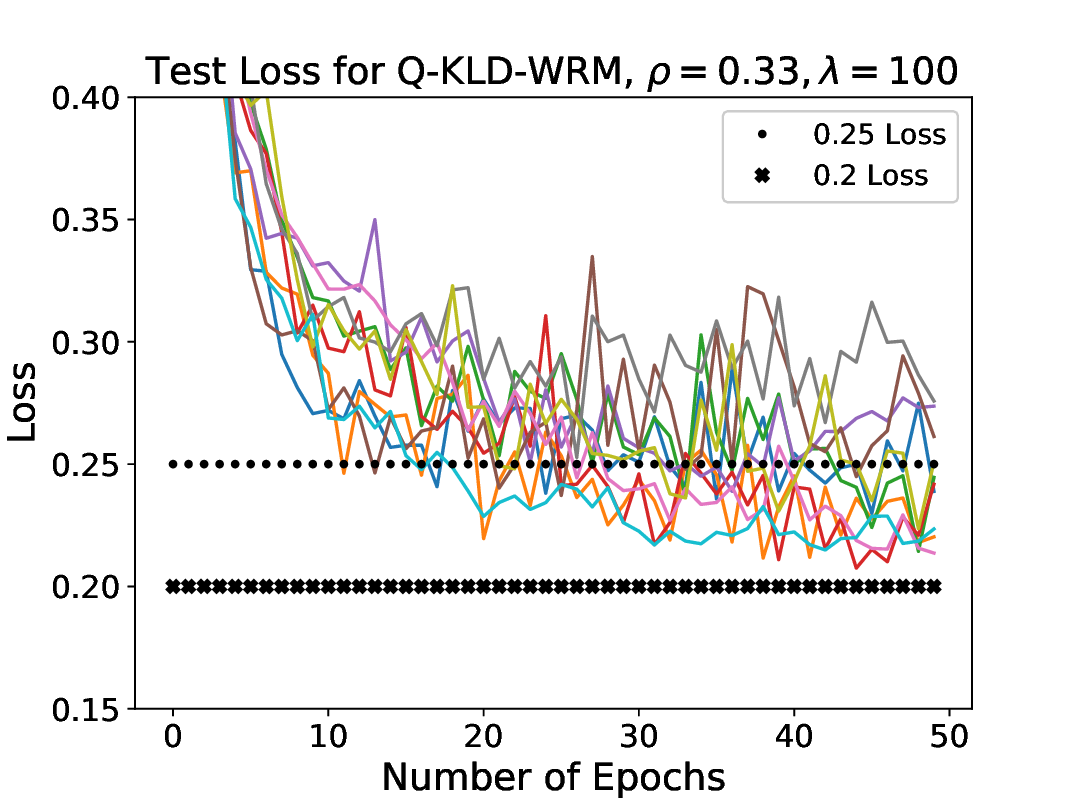}
	\includegraphics[trim={0.07cm 0.25cm 1.35cm 0.85cm},clip,width=0.24\textwidth]{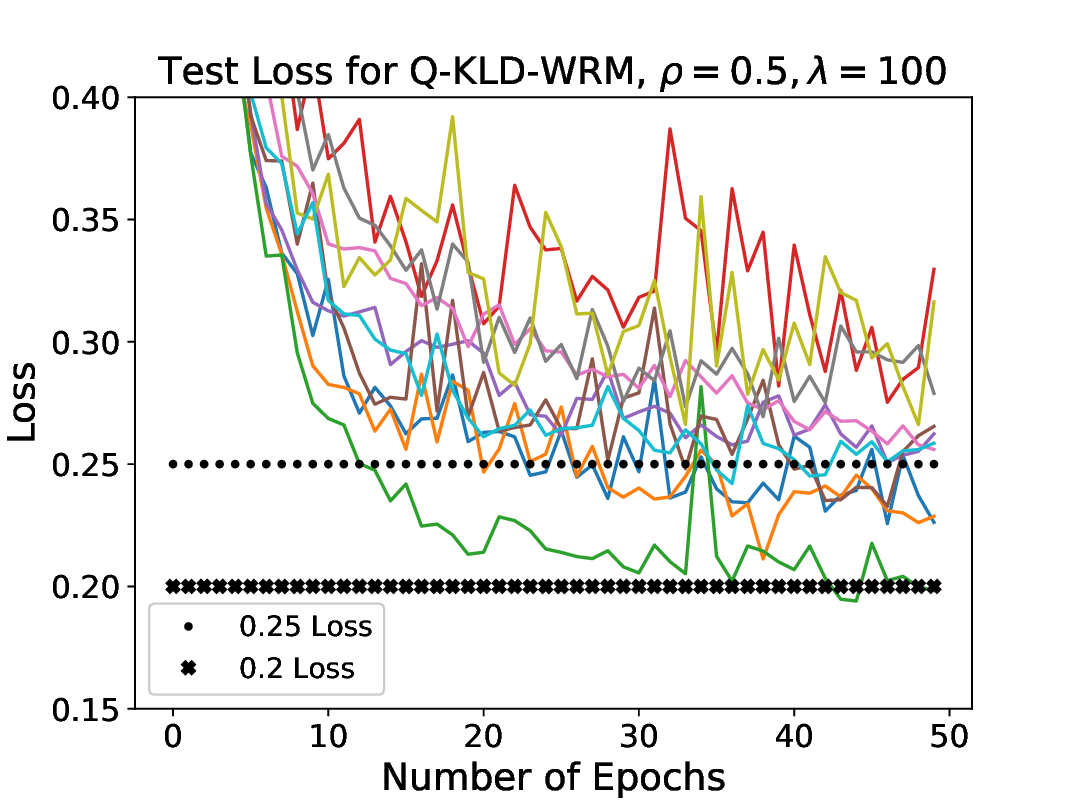}
	\includegraphics[trim={0.07cm 0.25cm 1.35cm 0.85cm},clip,width=0.24\textwidth]{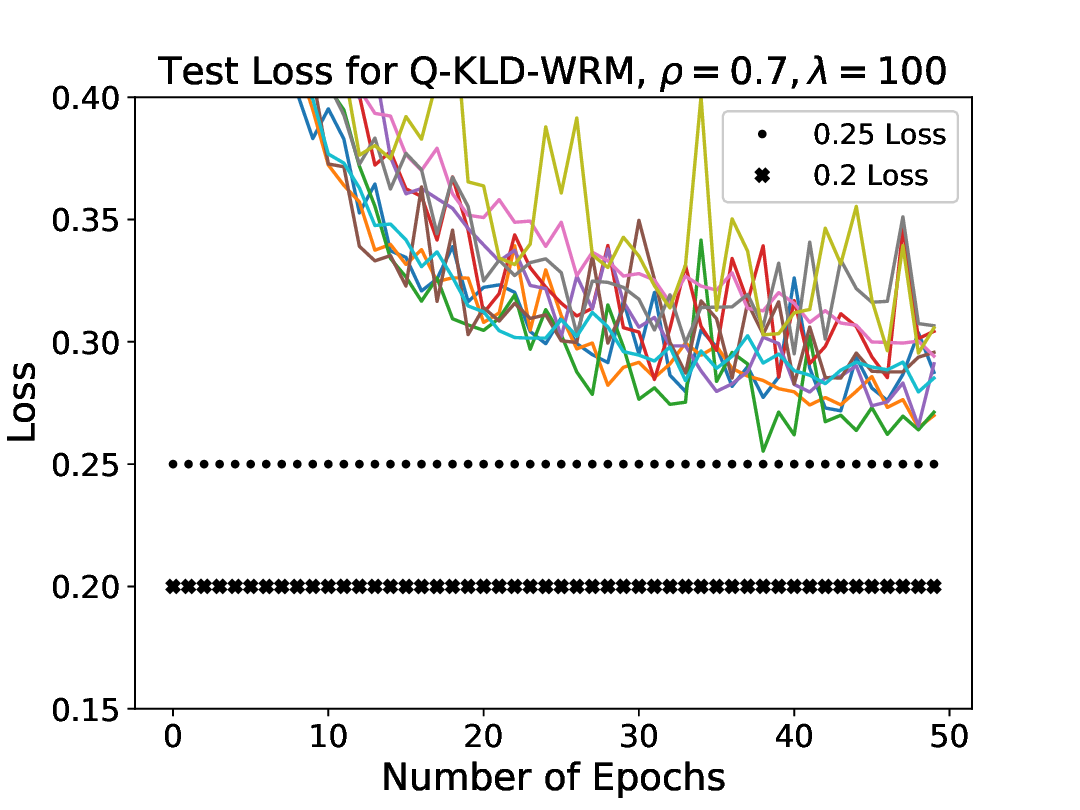}
	\includegraphics[trim={0.07cm 0.25cm 1.35cm 0.85cm},clip,width=0.24\textwidth]{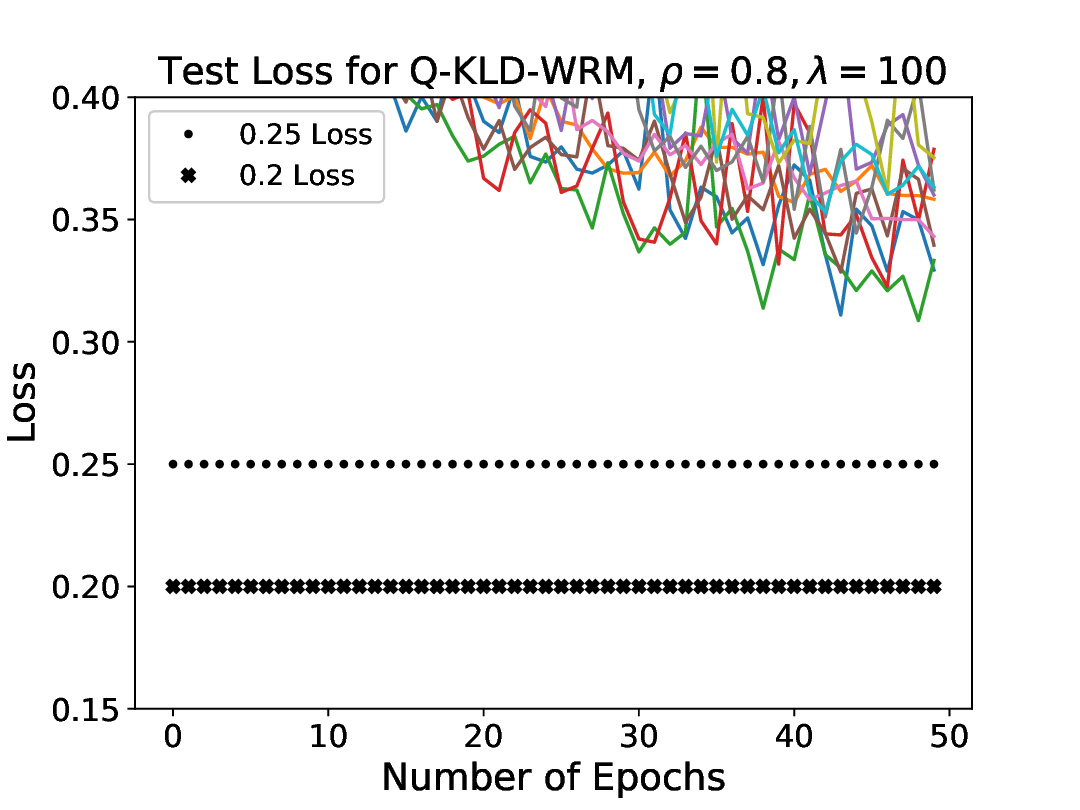}
	
	\vspace{+0.5ex}
	\includegraphics[trim={0.07cm 0.25cm 1.35cm 0.85cm},clip,width=0.24\textwidth]{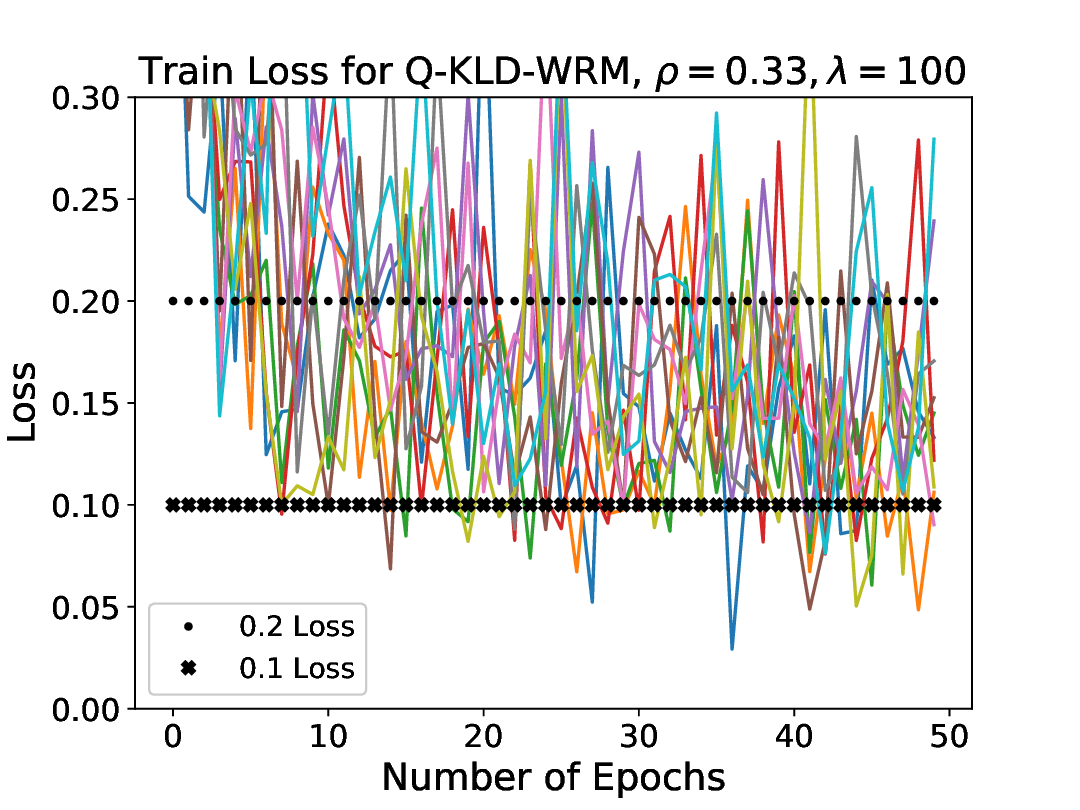}
	\includegraphics[trim={0.07cm 0.25cm 1.35cm 0.85cm},clip,width=0.24\textwidth]{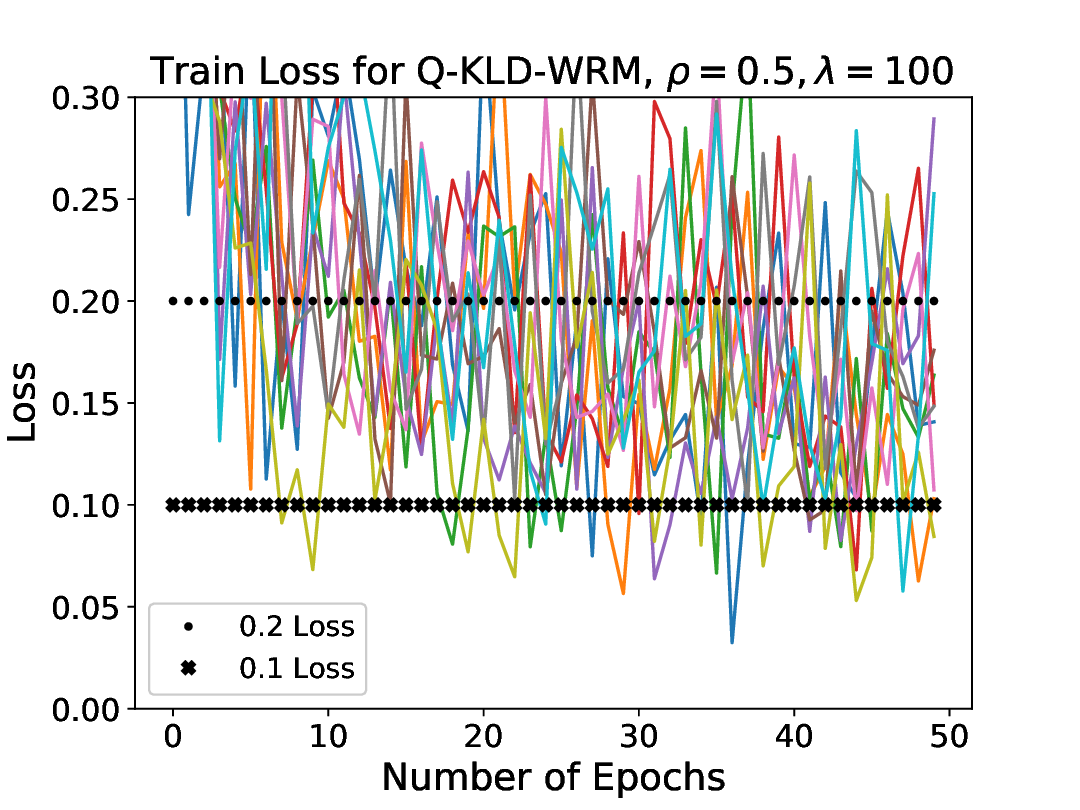}
	\includegraphics[trim={0.07cm 0.25cm 1.35cm 0.85cm},clip,width=0.24\textwidth]{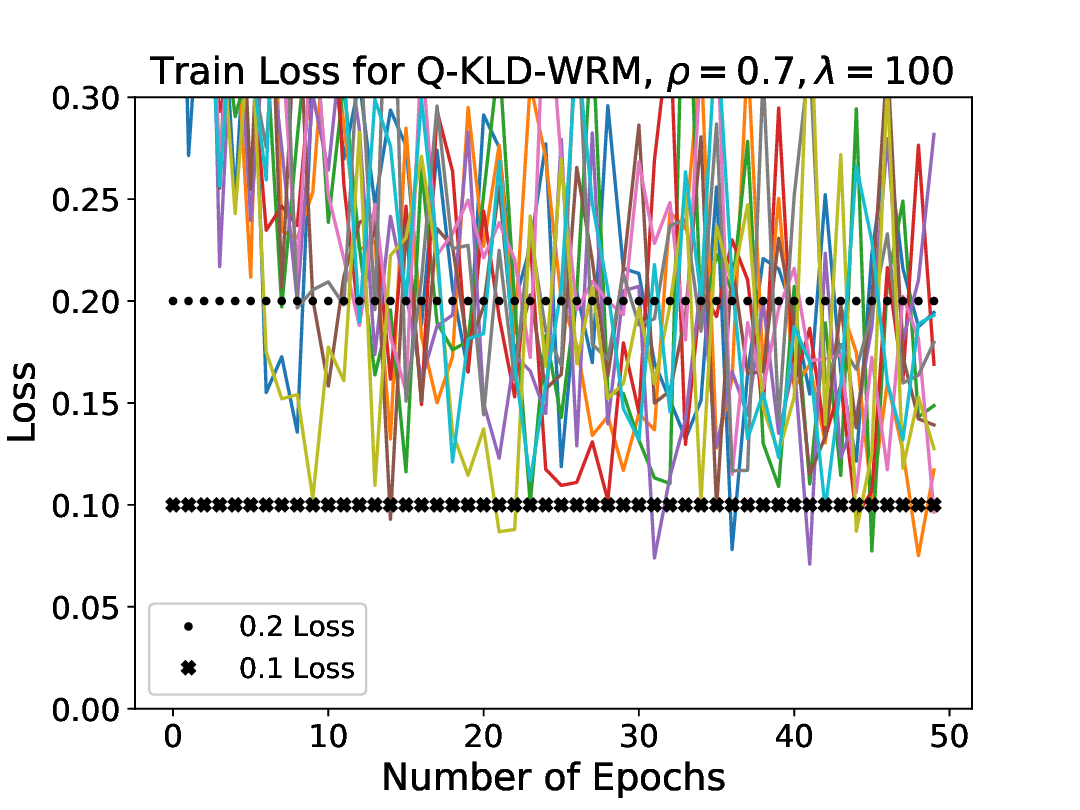}
	\includegraphics[trim={0.07cm 0.25cm 1.35cm 0.85cm},clip,width=0.24\textwidth]{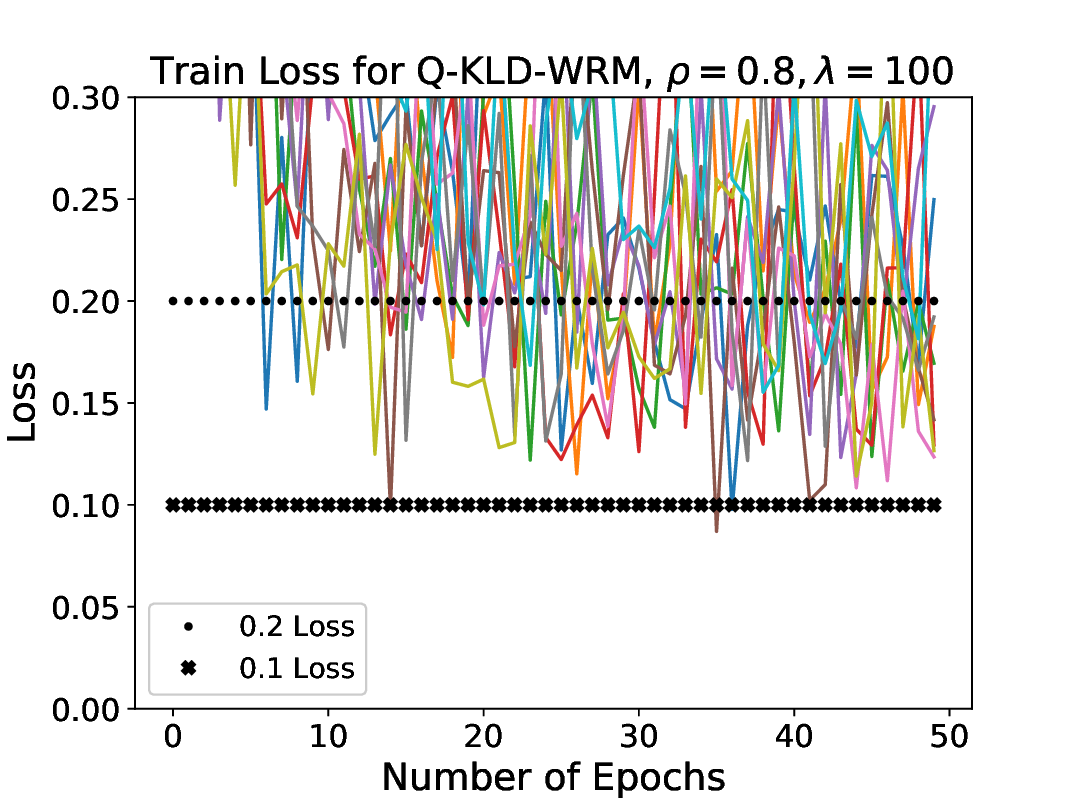}
	\vspace{-1.ex}
	\caption{\textbf{Q-KLD-WRM Results (MNIST Classification)}: from top to bottom: Test Accuracy, Test Loss, Training Loss. From left to right: different $\rho$ ($0.33$, $0.5$, $0.7$, $0.8$).The unspecified hyper-parameters were the same across all runs, and stated in \textit{Section 7.1}.  }
\end{figure*}
\begin{figure*}[hp]
	\centering
	\includegraphics[trim={0.07cm 0.25cm 1.51cm 0.85cm},clip,width=0.24\textwidth]{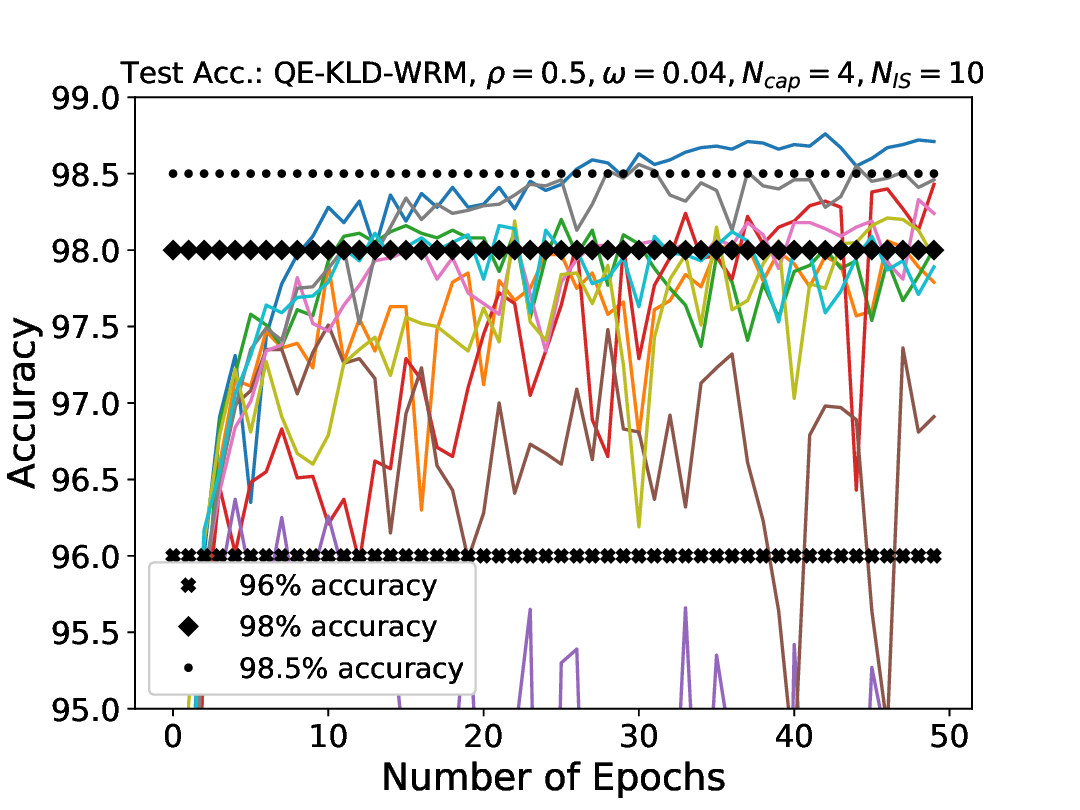}
	\includegraphics[trim={0.07cm 0.25cm 1.51cm 0.85cm},clip,width=0.24\textwidth]{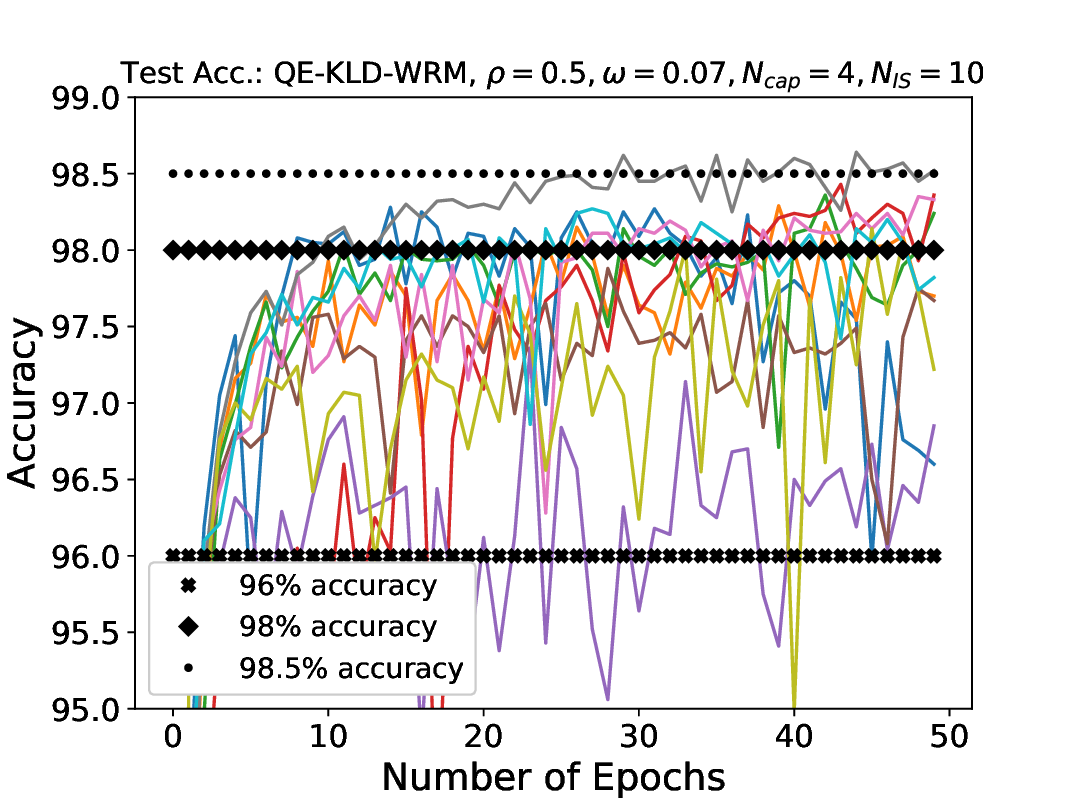}
	\includegraphics[trim={0.07cm 0.25cm 1.51cm 0.85cm},clip,width=0.24\textwidth]{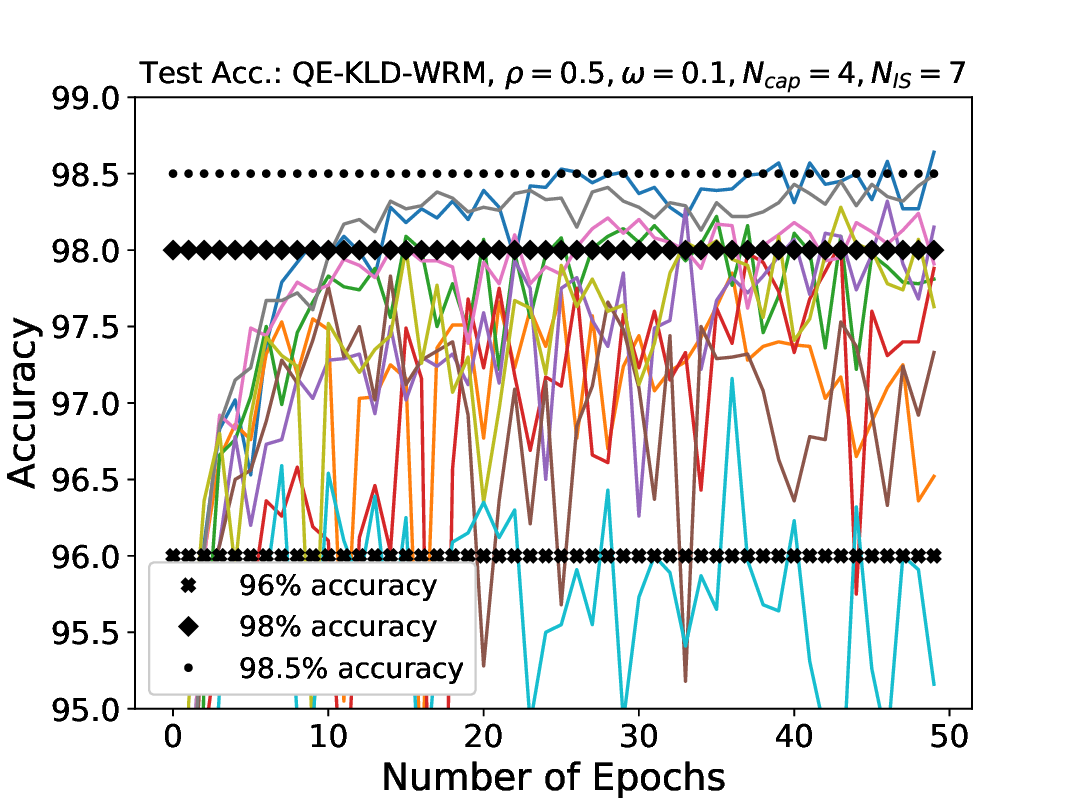}
	\includegraphics[trim={0.07cm 0.25cm 1.51cm 0.85cm},clip,width=0.24\textwidth]{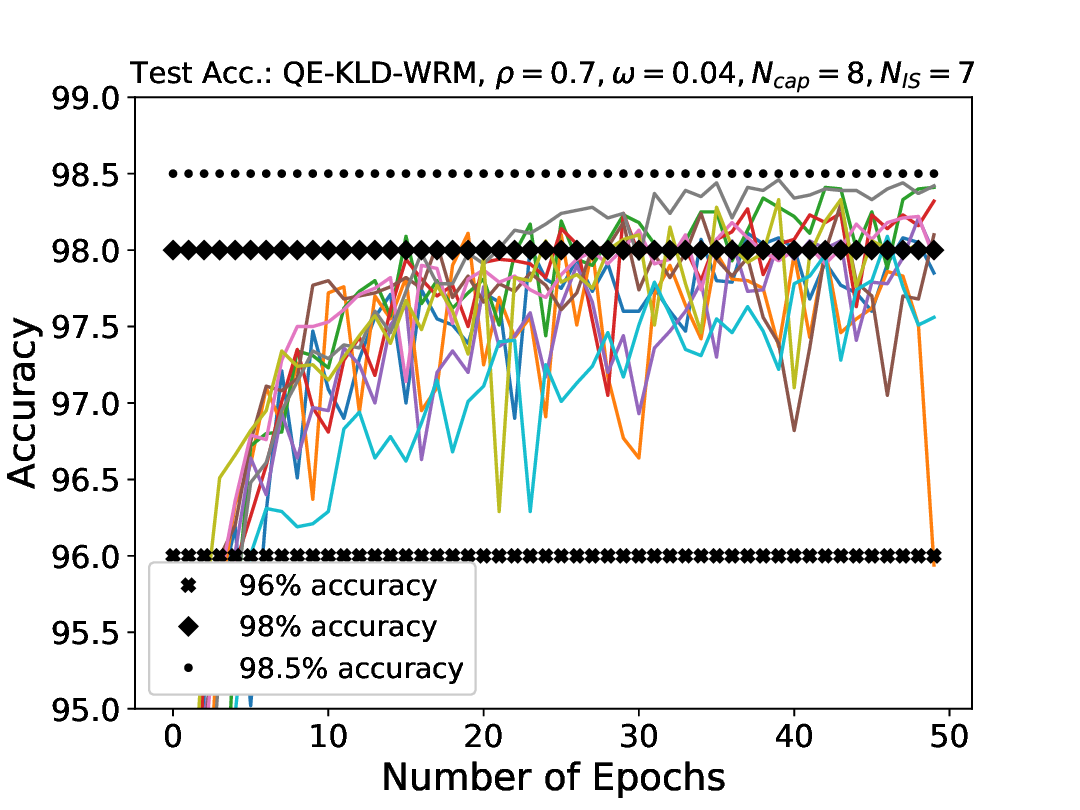}
	
	\vspace{+0.5ex}
	\includegraphics[trim={0.07cm 0.25cm 1.51cm 0.85cm},clip,width=0.24\textwidth]{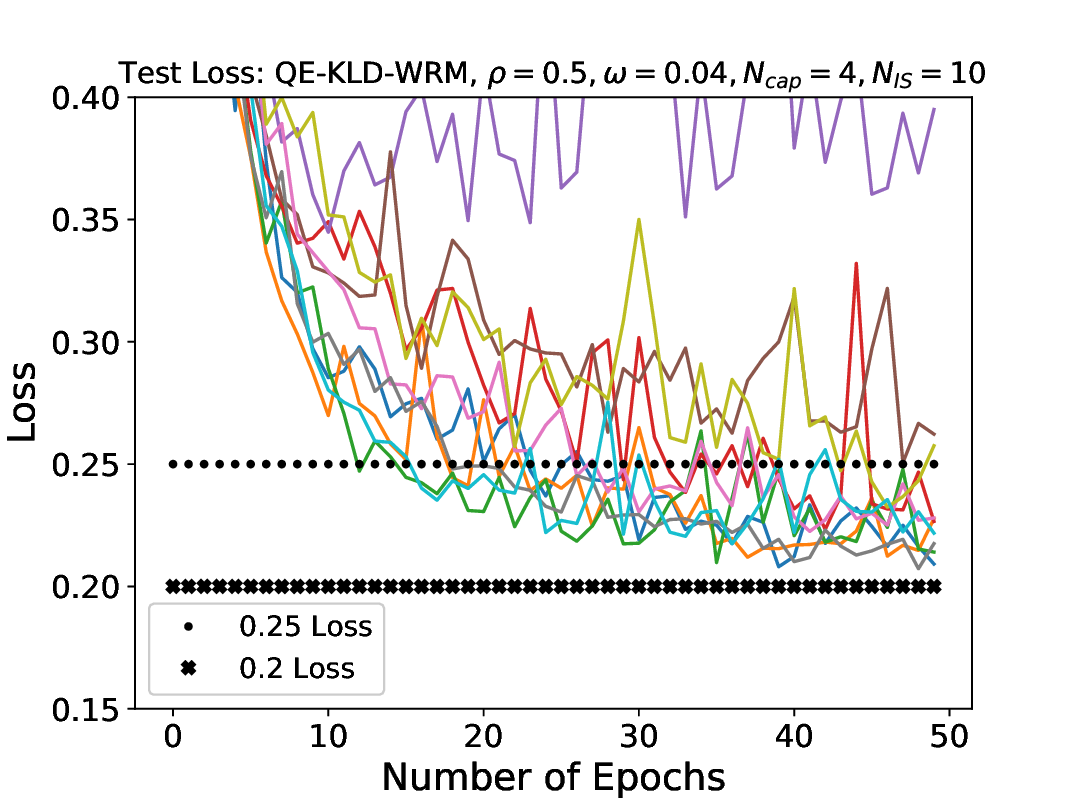}
	\includegraphics[trim={0.07cm 0.25cm 1.51cm 0.85cm},clip,width=0.24\textwidth]{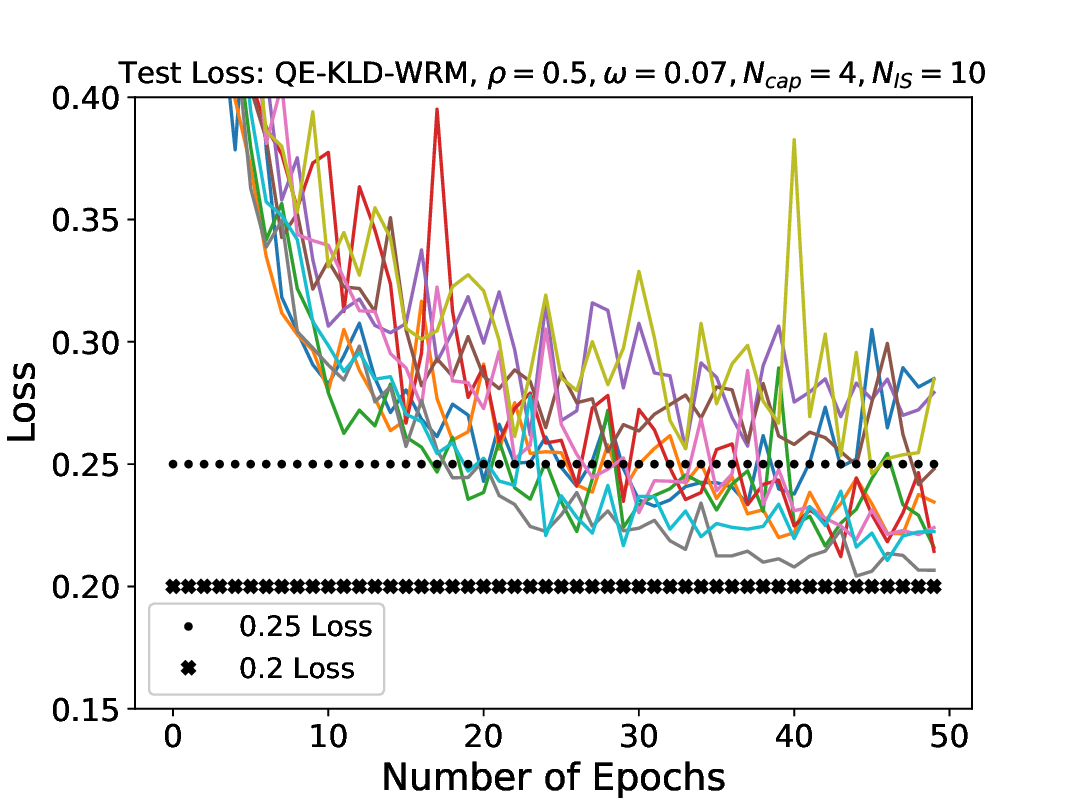}
	\includegraphics[trim={0.07cm 0.25cm 1.51cm 0.85cm},clip,width=0.24\textwidth]{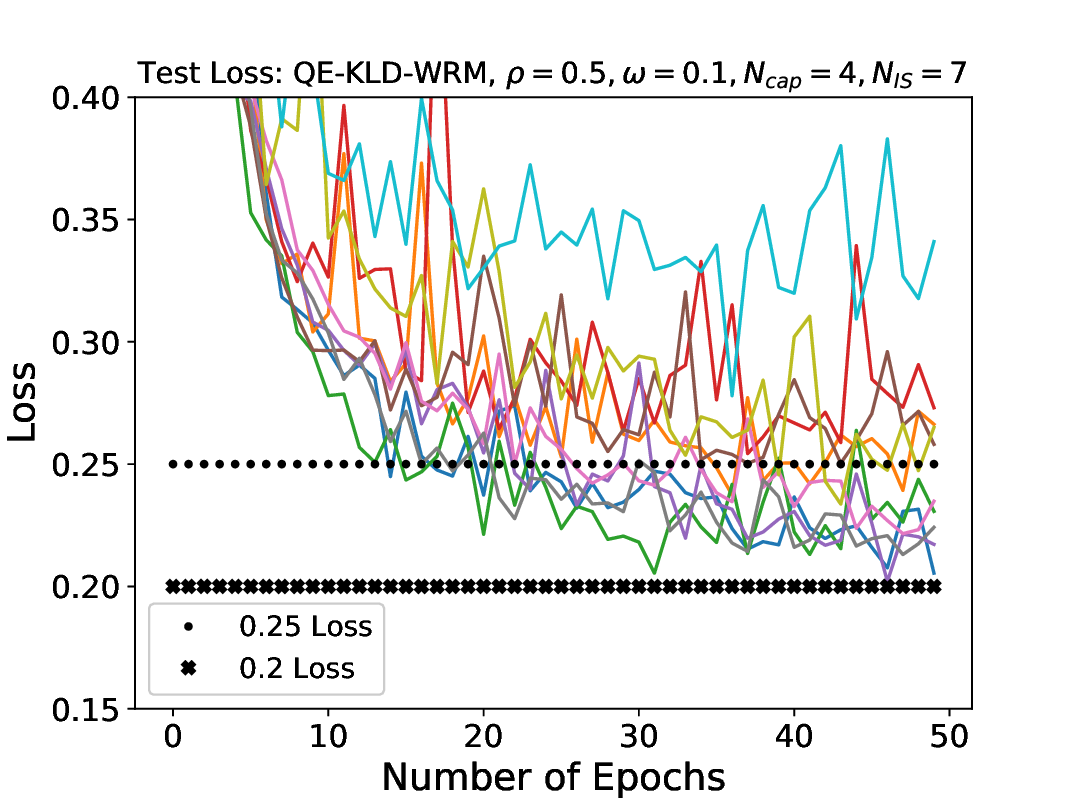}
	\includegraphics[trim={0.07cm 0.25cm 1.51cm 0.85cm},clip,width=0.24\textwidth]{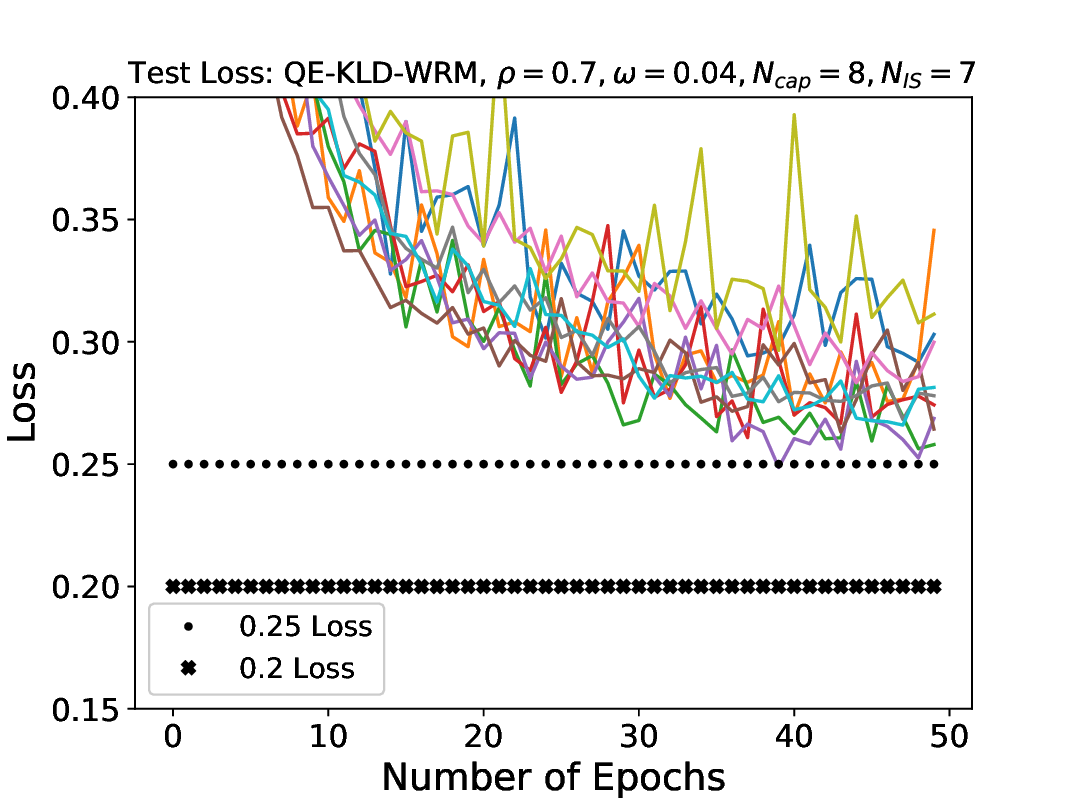}
	
	\vspace{+0.5ex}
	\includegraphics[trim={0.07cm 0.25cm 1.51cm 0.85cm},clip,width=0.24\textwidth]{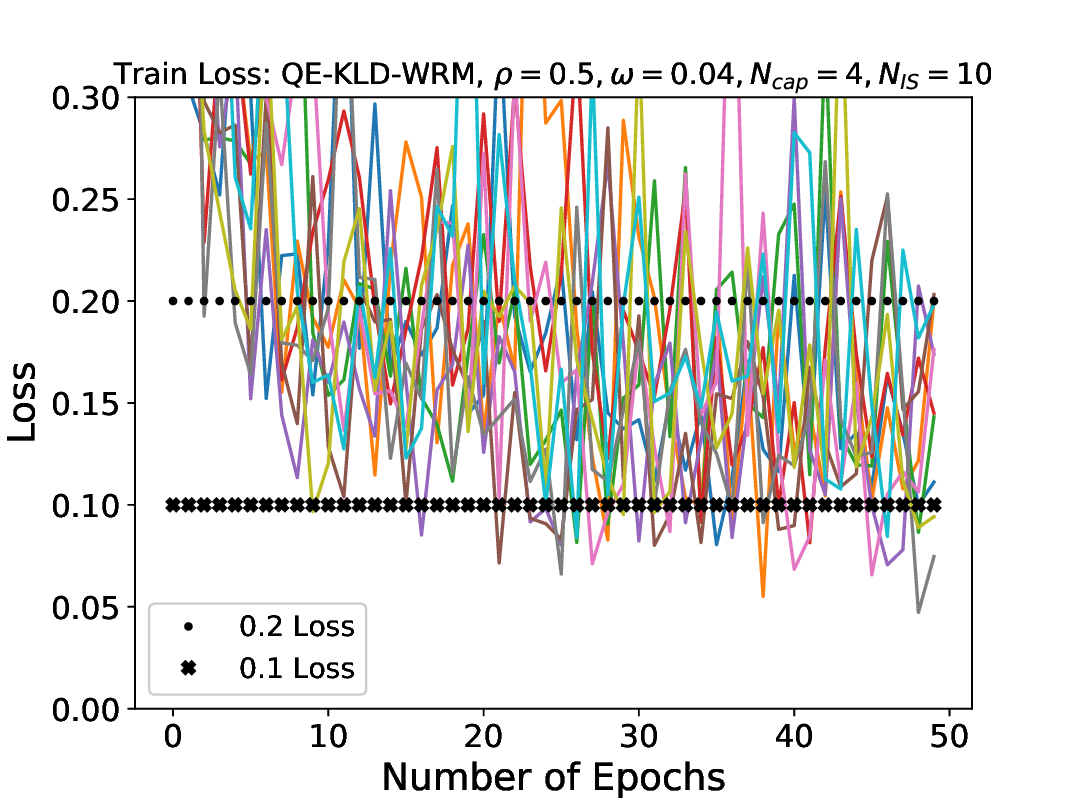}
	\includegraphics[trim={0.07cm 0.25cm 1.51cm 0.85cm},clip,width=0.24\textwidth]{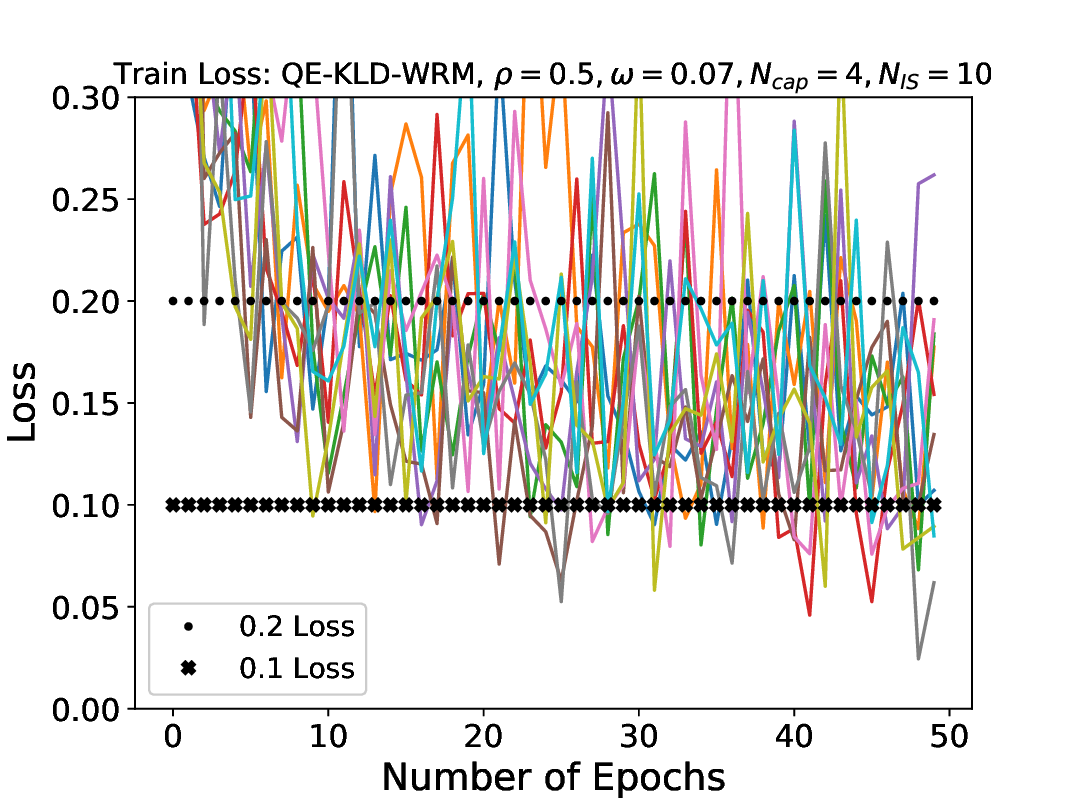}
	\includegraphics[trim={0.07cm 0.25cm 1.51cm 0.85cm},clip,width=0.24\textwidth]{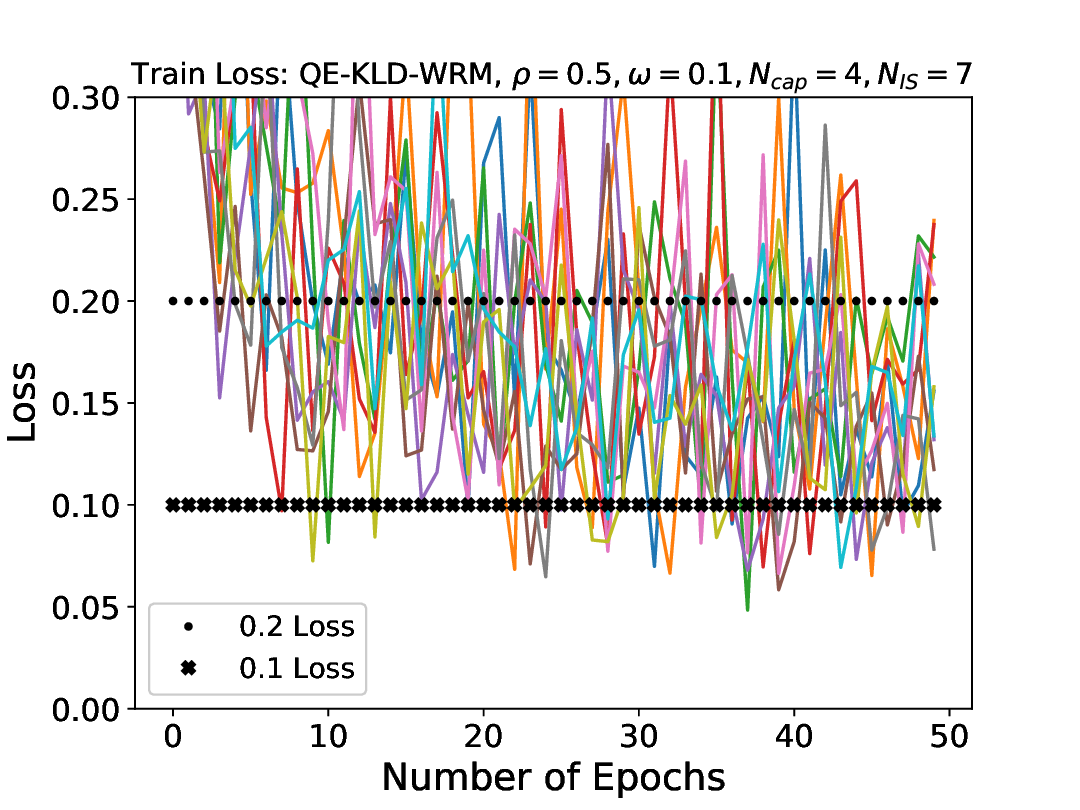}
	\includegraphics[trim={0.07cm 0.25cm 1.51cm 0.85cm},clip,width=0.24\textwidth]{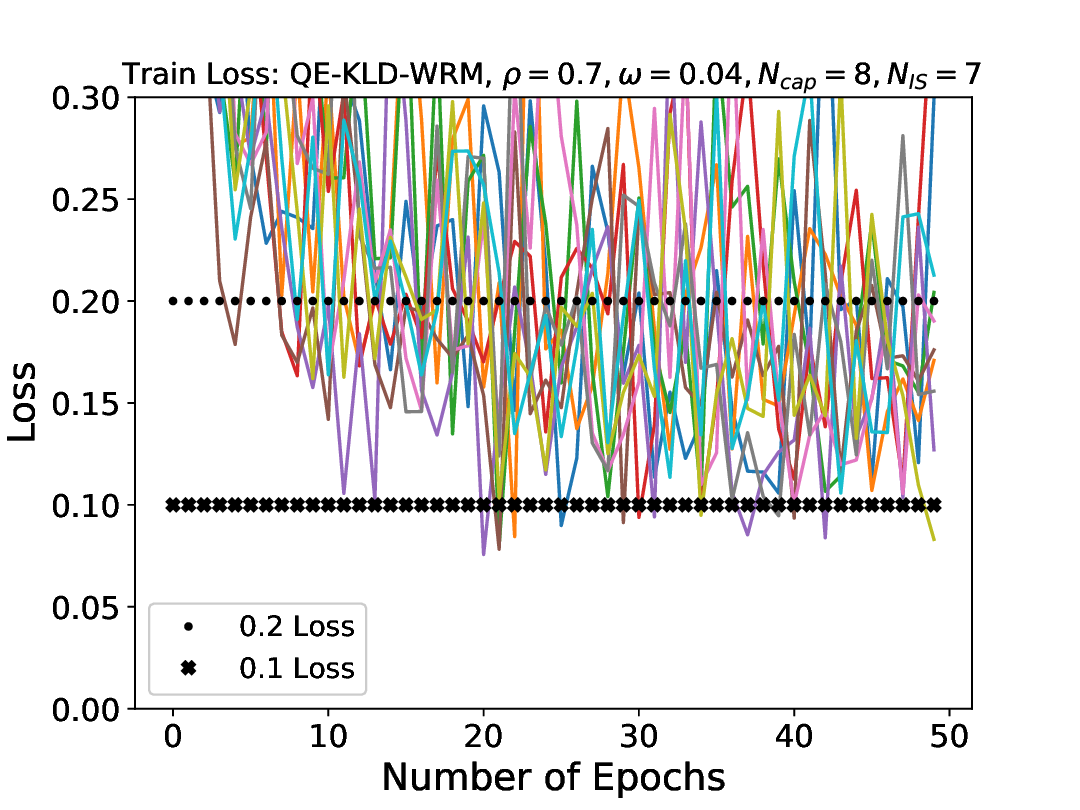}
	\vspace{-1.ex}
	\caption{\textbf{QE-KLD-WRM Results (MNIST Classification)}: from top to bottom: Test Accuracy, Test Loss, Training Loss. Left to right: different QE hyper-parameter values. The unspecified hyper-parameters were the same across all runs, and stated in \textit{Section 7.1}. }
\end{figure*}
\newpage
\begin{table*}[p]
	\caption{MNIST results summary. SO-KLD-WRM, Q-KLD-WRM and QE-KLD-WRM are our proposed KLD-WRM variants and K-FAC is the benchmark. Results are displayed for multiple hyper-parameter values of the four optimizers: K-FAC, SO-KLD-WRM, Q-KLD-WRM, QE-KLD-WRM. A slash indicates counting is somewhat debateable. Bolded entries are the best ones in their column. Top to bottom: Different optimizers and hyper-parameters. Left to right: Performance metrics (out of 10 runs).}
	\vspace{+1ex}
	\label{MNIST_results_table}
	\centering
	\scriptsize{
		\begin{tabular}{|c|c|c|c|c|c|c|c|c|c|}
			\hline
			\textbf{Optimizer / Metric} & \makecell{No.\ runs\\  test\\ acc.\\$\geq 98\%$} &\makecell{No.\ runs \\ test\\ acc.\\$> 98\%$} & \makecell{No.\ runs \\ test\\  acc.\\$\geq 98.5\%$} &  \makecell{No.\ runs\\ below\\ $0.25$ \\ test loss} & \makecell{No.\ runs\\ below \\$0.2$\\ test loss} & \makecell{Mean test \\ accuracy \\ (epoch 50)}&\makecell{STD. test \\ accuracy \\ (epoch 50)}& \makecell{Mean test\\ loss \\ (epoch 50)} & \makecell{STD.\ test\\ loss \\ (epoch 50)} \\
			\hline
			
			K-FAC: $\rho = 0.33$\ & 3 & 1 & 1 & 6 & 1 & $95.42\%$ & $3.73\%$ & 0.28 & 0.09\\
			K-FAC: $\rho = 0.5$\ & 3 & 2 & 1 & 5 & \textbf{2} & $95.69\%$ & $2.85\%$ & 0.27 & 0.08\\
			K-FAC: $\rho = 0.7$\ & 4 & 2 & 1 & 5 & \textbf{2} & $96.13\%$ & $2.58\%$ & 0.26 & 0.07\\
			K-FAC: $\rho = 0.95$\ & 3 & 3 & 1& 5 & \textbf{2} & $96.19\%$ & $3.2\%$ & 0.26 & 0.10\\
			\hline
			
			SO-KLD-WRM: $\rho = 0.33$\ & 4 & 2 & 0 & 7 & 0 & $97.60\%$ & $0.85\%$ & 0.25 & 0.04\\
			SO-KLD-WRM: $\rho = 0.5$\ & 7 / 8 & 4 /5 & 0 & 3 & 0 & $97.79\%$ & $0.41\%$ & 0.26 & 0.02\\
			SO-KLD-WRM: $\rho = 0.7$\ & 6 & 5 & 0 & 0 & 0 & $97.94\%$ & $0.5\%$ & 0.30 & 0.02\\
			SO-KLD-WRM: $\rho = 0.8$\ & 5 & 3 & 0 & 0 & 0 & $97.93\%$ & $0.37\%$ & 0.34 & 0.02\\
			\hline

			Q-KLD-WRM: $\rho = 0.33$\ & 3 & 1 & 0 & 7 & 0 & $97.47\%$ & $0.69\%$ & \textbf{0.24} & 0.02\\
			Q-KLD-WRM: $\rho = 0.5$\ & 5 & 4 & 0 & 4 & 0 & $97.69\%$ & $0.69\%$& 0.26 & 0.04\\
			Q-KLD-WRM: $\rho = 0.7$\ & 6 / 7 & 4 & 0 & 0 & 0 & $\mathbf{98.03\%}$ & $\mathbf{0.27\%}$ & 0.29 & \textbf{0.01}\\
			Q-KLD-WRM: $\rho = 0.8$\ & 4 & 1 & 0 & 0 & 0 & $97.88\%$ & $\mathbf{0.27\%}$ & 0.35 & 0.02\\
			\hline
			
			\makecell{QE-KLD-WRM: \\$\rho = 0.5,\, N_{IS} = 10$\\ $\omega = 0.04$,\, $N_{\text{cap}} = 4 $} \ & 7 / 8 & 7 & \textbf{2} & 8 & 0 & $97.85\%$ & $0.67\%$ & \textbf{0.24} & 0.03\\
			
			\makecell{QE-KLD-WRM: \\$\rho = 0.5,\, N_{IS} = 10$\\ $\omega = 0.07$,\, $N_{\text{cap}} = 4 $} ,  \ &8 & 7 & \textbf{2} &\textbf{9}& 0 & $\mathbf{98.01\%}$ & $0.64\%$ & \textbf{0.23} & 0.02\\
			
			\makecell{QE-KLD-WRM: \\$\rho = 0.5,\, N_{IS} = 7$\\ $\omega = 0.1$,\, $N_{\text{cap}} = 4 $} \ & 7 & 6 & \textbf{2} & \textbf{9} & 0 & $97.93\%$ & $0.46\%$ & \textbf{0.23} & 0.02\\
			
			\makecell{QE-KLD-WRM: \\ $\rho = 0.7,\, N_{IS} = 10$\\ $\omega = 0.04$,\, $N_{\text{cap}} = 8 $} \ & \textbf{9 / 10} & \textbf{8} & 0 & 0 & 0 & $\mathbf{98.07\%}$ & $0.44\%$ & 0.28 & 0.02\\
			\hline
		\end{tabular}	
	}
\end{table*}